\RequirePackage{fix-cm}
\documentclass[twocolumn]{svjour3}          
\smartqed  
\usepackage{natbib}
\usepackage[footnotesize]{subfigure}
\usepackage{float}
\usepackage{array}
\usepackage{hyperref}
\usepackage{amsmath,amssymb}
\usepackage{algorithm}
\usepackage[noend]{algpseudocode}
\usepackage{graphicx}
\usepackage{color}
\usepackage{xcolor}

\usepackage{marvosym}
\newcommand{\envelope}{(\raisebox{-.5pt}{\scalebox{1.45}{\Letter}}\kern-1.7pt)}

\newcommand{\nop}[1]{}

\newcommand{\eg}{{\sl e.g.}}
\newcommand{\ie}{{\sl i.e.}}

\newcommand{\G}{\mathcal{G}}
\newcommand{\R}{\mathcal{R}}
\newcommand{\E}{\mathcal{E}}

\newcommand{\x}{\mathbf{x}}

\newcommand{\cfbox}[2]{%
    \colorlet{currentcolor}{.}%
    {\color{#1}%
    \fbox{\color{currentcolor}#2}}%
}
\begin{document}

\title{Modeling Photographic Composition via Triangles
}


\author{Zihan Zhou$^*$ \and
  Siqiong He$^*$         \and
  Jia Li \and
  James Z. Wang 
}


\institute{Zihan Zhou$^*$ \envelope \and Siqiong He$^*$ \and James Z. Wang \at
	      College of Information Sciences and Technology, The Pennsylvania State University, University Park,
              PA, USA \\
              \email{zzhou@ist.psu.edu} \\
              ($^* =$ authors contributed equally)              
              \and
           Jia Li \at
              Department of Statistics, The Pennsylvania State University, University Park,
              PA, USA \\
              \email{jiali@stat.psu.edu}  
           \and
              Siqiong He \at
              \email{hesiqiong@gmail.com} 
           \and
           James Z. Wang \at
              \email{jwang@ist.psu.edu}
}

\date{Received: date / Accepted: date}

\maketitle

\begin{abstract}
The capacity of automatically modeling photographic composition is valuable for many real-world
machine vision applications such as digital photography, image retrieval, image
understanding, and image aesthetics assessment. The triangle
technique is among those indispensable composition methods on which professional photographers often rely. 
This paper proposes a system that can identify prominent triangle arrangements in two major categories of photographs: natural or urban scenes, and portraits. For the natural or urban scene pictures, the focus is on the effect of linear perspective. For portraits, we carefully examine the positioning of human subjects in a photo. We show that line analysis is highly advantageous
for modeling composition in both categories. 
Based on the detected triangles, new mathematical descriptors for composition are formulated and used to retrieve similar images. Leveraging the rich source of high aesthetics photos online, similar approaches 
can potentially be incorporated in future smart cameras to 
enhance a person's photo composition skills.

\keywords{Aesthetics \and Photographic composition \and Image
  segmentation \and Triangle detection}
\end{abstract}

\section{Introduction}
\label{sec:intro}

With the rapid advancement of digital camera and mobile imaging
technologies, we have witnessed a phenomenal increase of both
professional and amateur photographs in the past decade. Large-scale
social media companies, {\it e.g.}, Flickr, Snapchat, Instagram, and Facebook,
further empowered their users with the capability to share photos with
people all around the world. As over a billion new photos are added to
the Internet daily, there is an increasing demand for
creating machine vision application systems to manage, assess, and edit such
content. As a result, \emph{photo composition understanding} is becoming a
noteworthy area that has attracted attention of the research
community.

\begin{figure*}[ht!]
\centering
\includegraphics[height =1.05in]{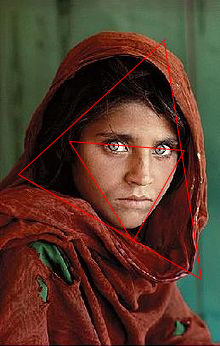}\hspace{1mm}
\includegraphics[height =1.05in]{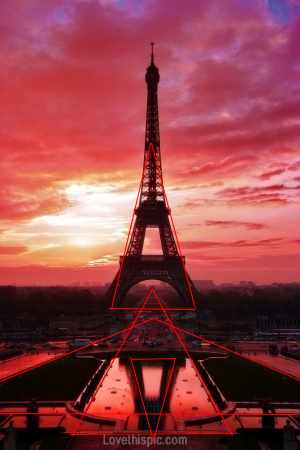}\hspace{1mm}
\includegraphics[height =1.05in]{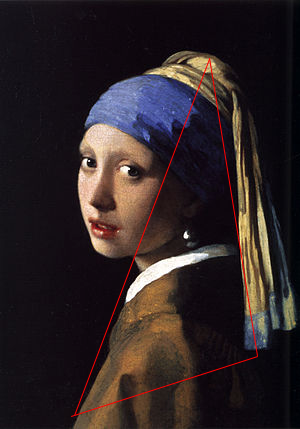}\hspace{1mm}
\includegraphics[height =1.05in]{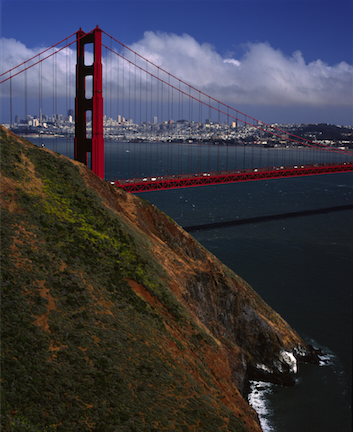}\hspace{1mm}
\includegraphics[height =1.05in]{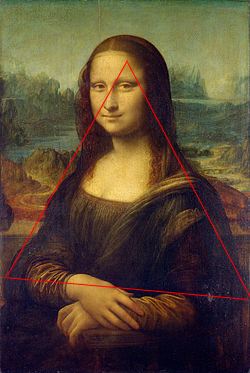}\hspace{1mm}
\includegraphics[height =1.05in]{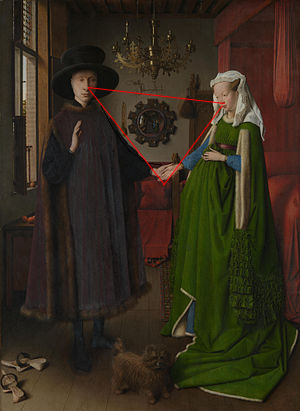}\hspace{1mm}
\includegraphics[height =1.05in]{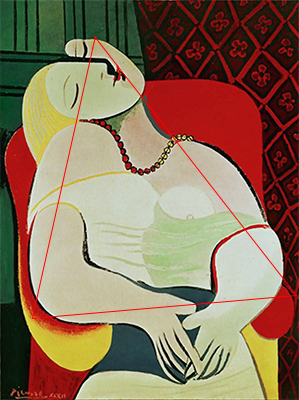}\hspace{1mm}
\includegraphics[height =1.05in]{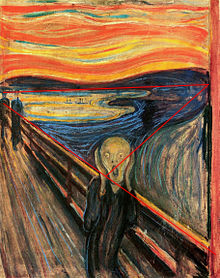} \\
\vskip 0.1in
\includegraphics[height =0.7in]{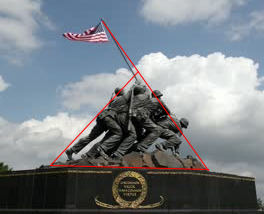}\hspace{1mm}
\includegraphics[height =0.7in]{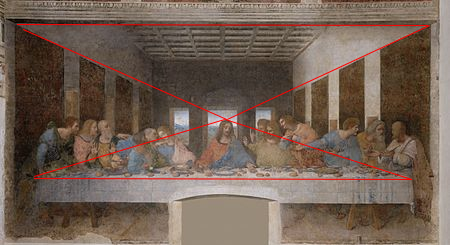}\hspace{1mm}
\includegraphics[height =0.7in]{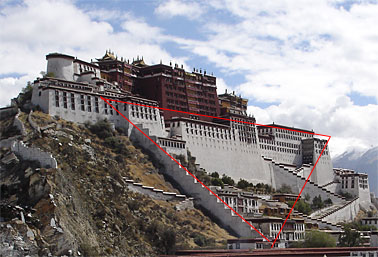}\hspace{1mm}
\includegraphics[height =0.7in]{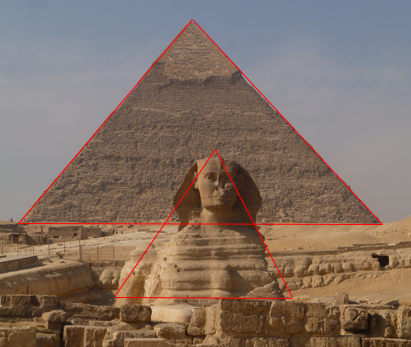}\hspace{1mm}
\includegraphics[height =0.7in]{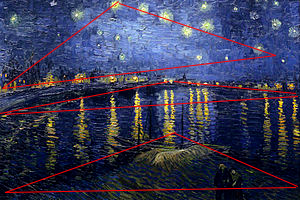}\hspace{1mm}
\includegraphics[height =0.7in]{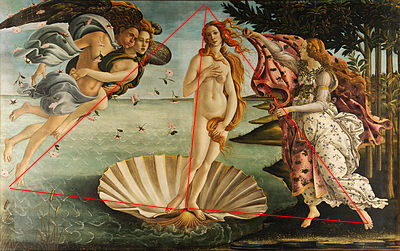} 
\caption{The use of triangles in visual art and architectural works. 
The suggested triangles are indicated in red. 
From left to right: {\bf (Row 1)} Afghan Girl, Eiffel Tower, 
Girl with a Pearl Earring, 
Golden Gate Bridge,
Mona Lisa,
The Arnolfini Portrait,
The Dream, The Scream.
{\bf (Row 2)}
Iwo Jima Memorial,
Last Supper,
Potala Palace,
Pyramid, 
Starry Night Over the Rhone,
The Birth of Venus.
}
\label{fig:triangle}
\vspace{-2mm}
\end{figure*}

Composition is the art of positioning or organization of objects and
visual elements (such as color, texture, shape, tone, and depth)
within a photograph or a visual art work. Known principles of organization
include balance, contrast, geometry, rhythm, perspective,
illumination, and viewing path. Automated understanding of photo
composition has been shown to benefit several applications such as
summarization of photo collections and assessment
of image aesthetics~\citep{ObradorSO10}. It can also be used to render
feedback to the photographer on the aesthetics of her
photos~\citep{ZhangSYQH12,YaoSQWL12}, and to suggest
improvements on the image composition through image re-targeting
\citep{LiuCWC10,BhattacharyaSS10}. In the literature, most work
on image composition understanding has focused on image-based rules
such as the simplicity of the scene, visual balance, the rule of
thirds, and the use of diagonal lines. Because of their simplicity,
these composition rules have been widely used to guide the
photographers at the moment of their creative work.

However, these rules are quite limited in capturing the wide
variations in photographic composition. As an expansion, we hereby
explore methods to identify an important composition technique,
namely, {\it the triangle technique}. In pictorial art, good composition is considered as a congruity or
agreement among the elements in a
design~\citep{LauerP11}. The design elements appear to belong
together as if there are some implicit visual connections between
them. 
One universal and interesting technique is to embed basic geometrical shapes in photographic
compositions~\citep{LauerP11,valenzuela2012picture}.  Human beings begin to
learn about basic geometrical shapes such as circles, rectangles, and
triangles at very young age. 
Moreover, because these shapes are
instantly recognized, subjects bounded within such shapes or
implicitly constructing such shapes are perceived as a unity. 
Among all basic geometric shapes, triangle is arguably the most
popular shape utilized by professional photographers to make a
composition more interesting. One can find numerous examples of the
use of triangles in visual art and photographic works
(Figure~\ref{fig:triangle}).

\subsection{Category-sensitive Composition Modeling}

We have developed category-sensitive approaches to detect the presence
of the triangle composition in an image. The proposed methods can
accurately locate a variety of triangles, even those which are
carefully designed by professional photographers but difficult to be
recognized by amateurs. 

In our work, we focus on two major categories of photographs: natural
or urban scenes, and portraits.  When we contemplate most important
photography genres, especially for consumers, arguably nature, travel,
architecture, portrait, fashion, children, street, and people scenes are
principal ones. With this work, we attempt to show that the triangle
technique can be applied to all of these main genres to improve the
quality of the photo. A useful mobile app or a smart camera based on
the triangle technique, for instance, can help the user/photographer
with any of these photographic needs. The user can indicate to the
system or device the particular type (roughly nature/urban
vs. portrait/people) before taking the photo. Alternatively, the
device can also determine the category automatically based on lens,
focal length, the presence of face/people, GPS locations, among other
available information. Other photography genres, such as animal, flower, macro, sport, and
event photography, are frequently done by more sophisticated or
professionally-trained photographers. The triangle technique is often
not as important for them as some other techniques, {\it e.g.},
depth-of-field controlling, high speed telephoto, motion blurring, and
emotion capturing. These genres are not covered in this work.

For both categories that we study here, ``lines'', as a critical
visual element in a picture, are exploited to detect triangles. From
a technical viewpoint, \emph{urban scenes} often have structures
or objects that possess relatively clean straight lines. It is
interesting to investigate\break whether a line-based technique can also be
applied to more organic images such as \emph{natural landscape photos} and \emph{portraits}. In this paper, we propose to examine ``contours'' in the images, which can be considered as a generalization of lines in both natural scenes and portraits. By incorporating the contours in the line-based analysis, our work supports the usefulness of line-based techniques even on photos where straight lines are mostly absent. Because composition analysis is in nascence,
we believe it is useful to show that the same line-based approach can
be useful for both scenes with man-made objects and natural
scenes or portraits.

While all of our techniques are based on the lines and contours, specific treatments for
these two broad categories of photos must be different. For example, with
the nature/cityscape scenes, we can leverage information about the vanishing points in determining the triangle technique used, while we cannot make use
of such information in most portraits. Despite the differences in
algorithmic treatments, we believe it is desirable to have the two
categories covered in the same publication because (1) it exemplifies the diversity of technical approaches needed to detect even simple visual elements like triangles in real-world scenarios, and (2) ultimately an
application system using the triangle technique will likely
incorporate all technologies described here under a uniform
framework.

Finally, our methods can potentially benefit many composition-based
applications. As an illustrative example, we apply the proposed methods for both categories to an image
retrieval application which aims to provide amateur users with on-site
feedback about the composition of their photos, in the same spirit as
~\citet{YaoSQWL12}. As we know, good composition
highlights the object of interest in a photo and attract the viewer's
attention immediately. However, it typically requires years of
practice and training for a photographer to master all the necessary
skills on photo composition. An effective way for an amateur or an
enthusiast to learn photography is through observing masterpieces,
ideally with guidance, and establishing comprehension about
photography. Professionally composed photographs can be valuable learning
resources for beginners. Nowadays, thanks to the increased popularity
of online photo sharing services such as Flickr and photo.net, one can
easily access millions of photos taken by people all around the
world. Such resources naturally provide us with opportunities to
develop new and more effective ways for photographers to
learn to improve composition skills.




\subsection{System Overview}



\begin{figure}[t!]
  \centering 
    \includegraphics[width=0.48\textwidth]{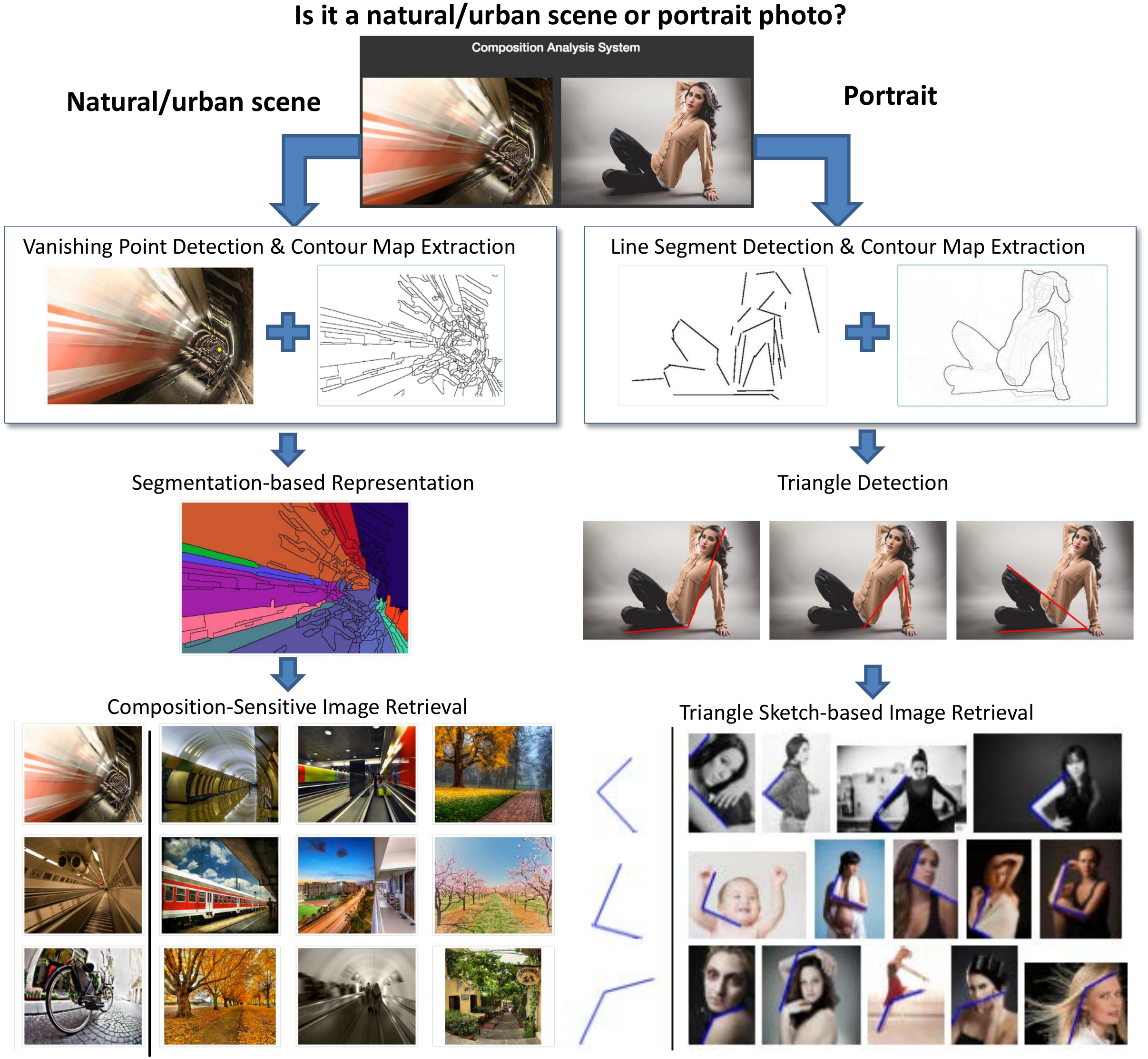}
  \caption{Overview of our composition analysis system.} 
  \label{fig:framework} 
\end{figure}

Figure~\ref{fig:framework} presents the framework of
our composition analysis system.
The user is first asked to indicate the type of photography he is
currently engaged in (\eg, using a command dial selector, often available on\break consumer-level digital
cameras). The system can provide assistance if the
selected type is natural/urban scene or portrait. Future smartphone applications can also leverage other sources of information to automatically determine this.

For natural/urban scene, the user can take a
test shot and upload it to our system as a query image. Given the
query image, our system analyzes the overall composition, particularly the use of triangles, in the image. Then, it retrieves
exemplar photos with similar compositions from a collection of photos
taken by experienced or accomplished photographers. 

For portraits, the user can provide a sketch indicating the shape and orientation of the triangle he is looking for. Given the triangle sketch, our system will retrieves exemplar photos containing the specific triangular configuration from a collection of photos taken by experienced or accomplished photographers. Here, the use of triangle sketches is motivated by the following two facts. First, a professional portrait photo typically contains multiple triangles. The triangle sketch allows users of our system to examine one such triangle at a time, and gain a deep understand on such configuration across multiple images. Second, in practice, users such as magazine editors may wish to embed certain triangular configuration in the image in order to fill in a specific page layout.

For both categories, the retrieved exemplar
photos can potentially serve as an informative guide for the users to achieve good
compositions in their photos. In the following, we discuss the algorithms we have developed for each scene type, respectively.

\subsubsection{Analyzing Triangles in Natural/Urban Scenes}

\begin{figure}[t!]
\centering
\begin{tabular}{ccc}
\includegraphics[width=0.95in]{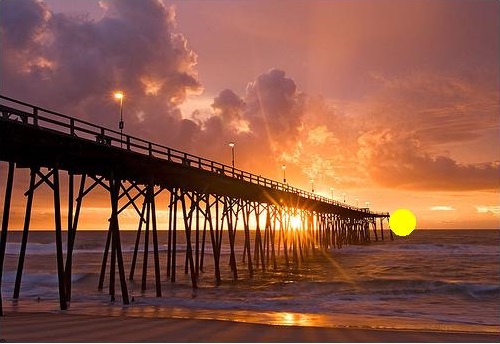}&\includegraphics[width=0.95in]{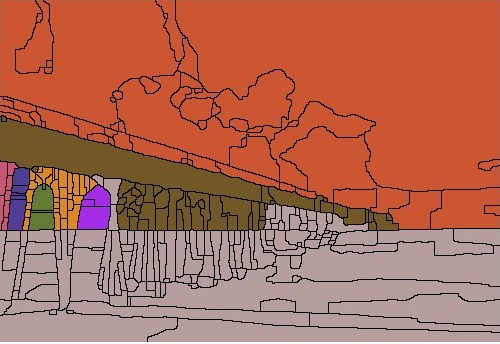}&\includegraphics[width=0.95in]{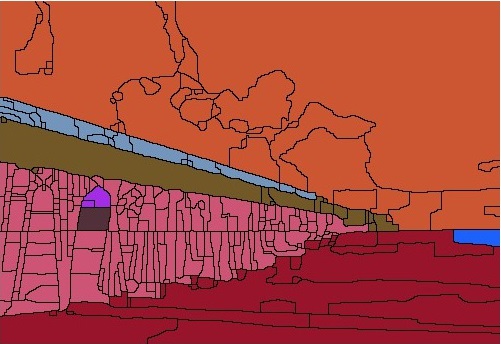}\\
\includegraphics[width=0.95in]{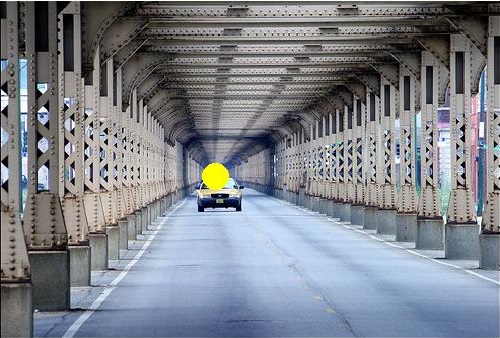}&\includegraphics[width=0.95in]{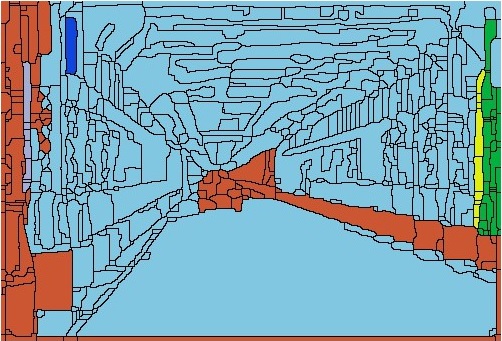}&\includegraphics[width=0.95in]{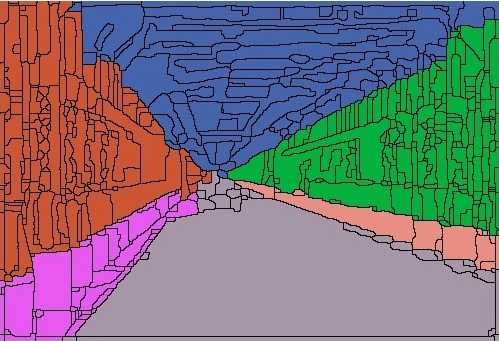}\\
\includegraphics[width=0.95in]{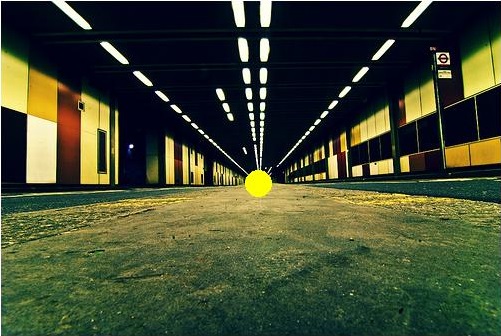}&\includegraphics[width=0.95in]{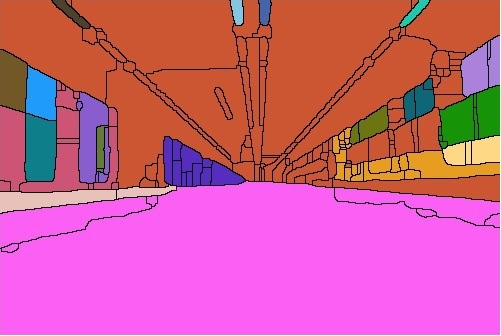}&\includegraphics[width=0.95in]{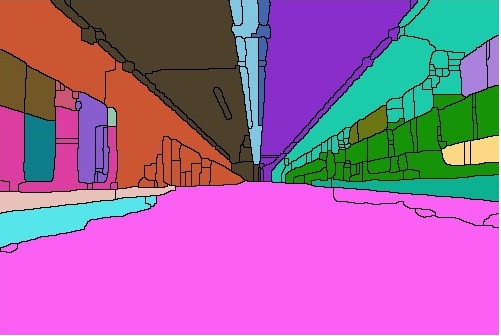}\\
\includegraphics[width=0.95in]{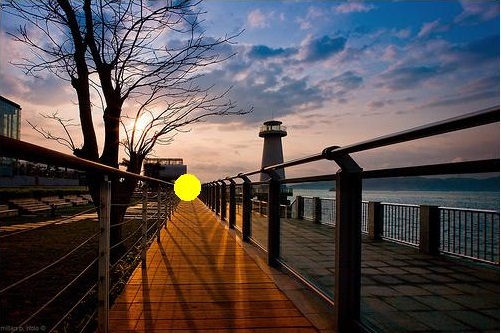}&\includegraphics[width=0.95in]{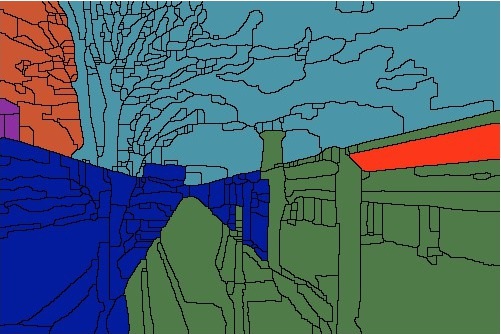}&\includegraphics[width=0.95in]{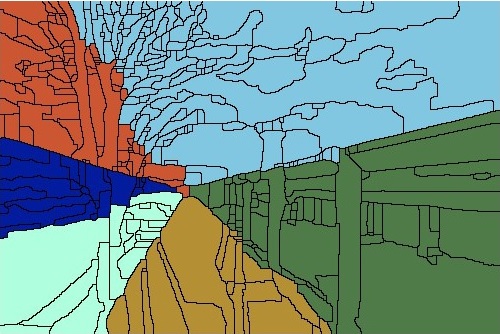}\\
{\small (a)} & {\small (b)} & {\small (c)}
\end{tabular}
\caption{Geometric image segmentation. {\bf (a)} The original image with the dominant vanishing point detected by our method (shown as a yellow round dot). {\bf (b)} Region segmentation map produced using a state-of-the-art method. {\bf (c)} Geometric image segmentation map produced by our method.}
\label{fig:1}
\end{figure}

In natural/urban scene photography, photographers often make use of the linear perspective effects in the images to emphasize the sense of 3D space in a 2D photo. 
According to the perspective camera geometry, all
parallel lines in 3D converge to a single vanishing point
in the image, generating a set of triangular regions. Figure~\ref{fig:1}
shows some examples. In order to convey a strong impression of 3D
space and depth to viewers, the vanishing point has to lie within
or near the image frame and associates with the dominant structures of
the scene ({\it e.g.}, grounds, large walls, bridges). We regard such a vanishing point as the
\emph{dominant vanishing point} of the particular image.  As one
adjusts shooting angle, both the location of the dominant vanishing
point as well as the sizes, shapes, and orientations of triangular
regions relating to the vanishing point will change. An experienced
photographer often utilizes such technique to produce various image
compositions that convey different messages or impressions to viewers.

Accordingly, to model the composition for such\break scenes, we propose to
\emph{partition an image into photometrically and geometrically
  consistent regions according to the dominant vanishing point}. In
our work, we assume that each geometric region can be roughly modeled
by a flat surface, or a plane. Due to the perspective effect, these planes are naturally projected into \emph{triangular regions} in the image. As shown in Figure~\ref{fig:1}(c), such a partition naturally provides us with a novel \emph{holistic} yet \emph{compact} representation of the 3D scene geometry that respects the perspective effects of the scene the image captured in, and allows us to derive a notion of relative depth and scale for the objects. Nevertheless, obtaining such a representation is a challenging problem for the following reasons. 

First, given any two adjacent geometric regions in an image, there may not be a distinguishable boundary in terms of photometric cues (\eg, color, texture) so that they can be separated. For example, the walls and the ceiling in the second photo of Figure~\ref{fig:1} share the same building material. Because existing segmentation algorithms primarily depend on the photometric cues to determine the distance between regions, they are often unable to separate these regions from each other (see Figure~\ref{fig:1}(b) for examples). To resolve this issue, we propose a novel hierarchical image segmentation algorithm that leverages significant geometric information about the dominant vanishing point in the image. Specifically, we compute a geometric distance between any two adjacent regions based on the similarity of the angles of the two regions in a polar coordinate system, with the dominant vanishing point being the pole. By combining the geometric cues with conventional photometric cues, our method is able to preserve essential geometric regions in the image.

Second, detecting the dominant vanishing point\break from an image itself is a nontrivial task. Typical vanishing point detection methods assume the presence of a large number of strong edges in the image. However, for many photos of outdoor scenes, such as the image of an arbitrary road, there may not be adequate clearly-delineated edges that converge to the vanishing point. In such cases, the detected vanishing points are often unreliable and sensitive to image noise. To overcome this difficulty, we observe that while it may be hard to detect the local edges in these images, it is possible to directly infer the location of the vanishing point by aggregating the aforementioned photometric and geometric cues over the entire image (Figures~\ref{fig:1} and~\ref{fig:vp}). Based on this observation, we develop a novel vanishing point detection method which does not rely on the existence of strong edges, hence works better for arbitrary images.

\subsubsection{Modeling Composition in Portraits}
Two fundamental questions are often raised when analyzing the
composition of a portrait photograph: where are the human subjects in the photo and
how do they pose? Traditional composition rules provide us with guidelines to answer the first question. For example, the rule of thirds suggests that putting the human subjects near the 1/3 point of an image is more appealing than at
the center. Based on these rules, several methods have been developed to model and assess the positioning of human subjects in a photo. 

Nevertheless, the second question remains a challenge. To address this
problem, we leverage an important observation in portrait photograph:
\textit{experienced portrait photographers often use triangle
  techniques to create interesting and good-looking poses for human
  subject.} For example, a widely-used rule for posing
is that one should try to avoid 90-degree body angles, because they
often look unnatural and strained. In addition, triangle techniques
are also frequently used to unify multiple human subjects and the surrounding environment, such as chairs and lamps, in the portrait photo.

Despite the popularity of triangle techniques in portrait photography,
it is often difficult for less experienced amateurs to recognize such
triangles, because most triangles do not have explicit edges and
sometimes are even constructed by different
objects. Moreover, triangles in portraits can be of various sizes, shapes,
orientations, and appearances. Hence, our goal 
is to automatically detect potential triangles from professional
photographers' work in order to help amateurs recognize and learn
from the usage of triangle techniques.

Our algorithm can be divided into two steps: First, a line segment
detection module is used to extract candidate
line segments from an image, which are subsequently filtered using the global contour information in the image. Second, the filtered line segments are fed into a triangle
detection module as the candidate sides of triangles. Specifically, a RANSAC
algorithm is developed to randomly pick two sides from all the candidates
and fit the triangle. Two metrics, \emph{Continuity
  Ratio} and \emph{Total Ratio}, are defined to evaluate the fitness of
these triangles. Only those triangles with high fitness scores will be shown to the
users.

\subsection{Contributions}

This paper makes the following main contributions:
\begin{itemize}
\item {\it Composition-sensitive retrieval for on-site feedback:} We
  developed triangle detection techniques for image composition
  understanding so that photographs with similar composition as the query photo can be retrieved from a collection of
  photos taken by professional photographers. User studies are conducted to verify to effectiveness of the retrieved exemplars in providing amateur users with useful information and guidance about photo composition.

\item {\it Triangle detection and geometric image segmentation for natural/urban scenes:} We model the composition of typical
  natural/urban scene images by examining the perspective effects and
  partitioning the image into photometrically and geometrically
  consistent regions using a novel hierarchical image segmentation
  algorithm.

\item {\it Dominant vanishing point detection:} By aggregating the
  photometric and geometric cues using our segmentation algorithm, we
  develop an effective method to detect the dominant vanishing point
  in an arbitrary image.

\item {\it Triangle detection in portraits:} We propose a\break
 RANSAC algorithm to
  detect triangles in portraits with a variety of sizes, shapes,
  orientations, and appearances, with the goal of helping less experienced users to
  understand and learn from the usage of
  triangles in professional photographers' work.
\end{itemize}

Admittedly, our technique for natural/urban scenes and portraits cannot yet model all
potential compositions in such photos, especially when there is a lack
of triangles. While in this paper we focus on the use of triangle techniques in photography, we also point out that there
are many works which study other important aspects of composition,
including the semantic features (\eg, buildings, trees,
roads)~\citep{HoiemEH05,HoiemEH07,GouldFK09,GuptaEH10}. It would be ideal to integrate all these features in
order to gain a deeper understanding of the image composition, but a
thorough discussion on this topic is beyond the scope of this paper.

This article is an extension of our earlier work~\citep{ZhouHLW15}. The primary new contributions are an expanded discussion on the use of triangles in photographic composition and a category-sensitive approach to study such usage (Section~\ref{sec:intro}), a novel triangle detection method for portrait photos (Section~\ref{sec:por}), and its application to providing amateur users with on-site feedback about the composition of their photos (Section~\ref{sec:retrieval-portrait}). We further demonstrate the effectiveness of our triangle detection method for portrait photos in Section~\ref{sec:exp-portrait}.

\section{Related Work}\label{sec:rel}

\subsection{Composition Modeling}
Standard composition rules such as the rule of thirds, golden ratio and low depth of field have played important roles in early works on image aesthetics assessment~\citep{DattaJLW06, LuoT08, su2011scenic}. \citet{ObradorSO10} later showed that by using only the composition features, one can achieve image aesthetic classification results that are comparable to the state-of-the-art. Recently, these rules have also been used to predict high-level attributes for image interestingness classification~\citep{DharOB11}, recommend suitable positions and poses in the scene for portrait photography~\citep{ZhangSYQH12}, and develop both automatic and interactive cropping and retargeting tools for image enhancement~\citep{LiuCWC10, BhattacharyaSS10, FangLMS14}. \citet{marchesotti2011assessing} showed that, by aggregating statistics computed from low-level generic image features, one can achieve better performance in assessing image aesthetics than using hand-crafted composition rules. In addition,~\citet{YaoSQWL12} proposed a composition-sensitive image retrieval method which classifies images into horizontal, vertical, diagonal, textured, and centered categories, and uses the classification result to retrieve exemplar images that have similar composition and visual characteristics as the query image. However, none of them study the use of triangles in photography.

\subsection{3D Modeling and Segmentation of Images}
Various methods have been proposed to extract 3D scene structures from a single image. The GIST descriptor~\citep{OlivaT01} is among the first attempts to characterize the global arrangement of geometric structures using simple image features such as color, texture and gradients. Following this seminal\break work, a large number of supervised machine learning methods have been developed to infer approximate 3D structures or depth maps from the image using carefully designed models~\citep{HoiemEH05, HoiemEH07, GouldFK09, SaxenaSN09, NedovicSRG10} or grammars~\citep{GuptaEH10, HanZ05}. In addition, models tailored for specific scenarios have been studied, such as indoor scenes~\citep{LeeHK09, HedauHF09, HedauHF10} and urban scenes~\citep{BarinovaKYLLK08}. However, these works all make strong assumptions on the structure of the scene, hence the types of scene they can handle in practice are limited. Despite the above efforts, obtaining a good estimation of perspective in an arbitrary image remains an open problem.

Typical vanishing point detection algorithms are\break based on clustering edges in the image according to their orientations. \citep{KoseckaZ02} proposed an Expectation Maximization (EM) approach to iteratively estimate the vanishing points and update the membership of all edges. Recently, a non-iterative\break method is developed to simultaneously detect multiple vanishing points in an image~\citep{Tardif09}. These methods assume that a large number of line segments are available for each cluster. To reduce the uncertainty in the detection results, a unified framework has been proposed to jointly optimize the detected line segments and vanishing points~\citep{TretiakBKL12}. For images of scenes that lack clear line segments or boundaries, specifically the unstructured roads, texture orientation cues of all the pixels are aggregated to detect the vanishing points~\citep{Rasmussen04, KongAP09}. But it is unclear how these methods can be extended to general images.

Image segmentation algorithms commonly operate on low-level image features such as color, edge, texture and the position of patches~\citep{ShiM00, FelzenszwalbH04, Li:2011, ArbelaezMFM11, MobahiRYSM11}. But it was shown in~\citep{RussellESFZ09} that given an image, images sharing the same spatial composites can help with the unsupervised segmentation task. 

\subsection{Portrait Photo Analysis}
Few studies have focused on aesthetic analysis of portrait
images. \citet{jin2010learning} studied the
critical role of lighting in portrait photography. Varying lighting patterns shift lights and shadows on the face,
change the area ratios between them, and generate 3D perception from
the 2D photograph. While the learned artistic portrait lighting
templates in that work are able to capture the
arrangement of low level features, our work aims at modeling the
usage of more holistic composition techniques in portrait
photography. ~\citet{ZhangSYQH12} developed a method to automatically recommend suitable
positions and poses of people in natural scenes. However, the
recommendation is based on matching simple 2D compositional features
and manually labeled human body poses.

\section{Natural/Urban Scene Composition Modeling}\label{sec:nat}

In this section, we present the technical details of our geometric image segmentation algorithm for triangle detection and composition modeling in natural/urban scene photos.
Since our segmentation method follows the classic hierarchical segmentation framework, we give an overview of the framework and some of the state-of-the-art results in Section \ref{sec:seg-overview}.  In Section \ref{sec:method-measure}, we introduce our geometric distance measure for hierarchical image segmentation, assuming the location of the dominant vanishing point is known. The proposed geometric cue is combined with traditional photometric cues in Section \ref{sec:method-seg} to obtain a holistic representation for composition modeling. In Section~\ref{sec:method-vp}, we further show how the proposed distance measure, when aggregated over the entire image, can be used to detect the dominant vanishing point in an image. 

\subsection{Hierarchical Image Segmentation}
\label{sec:seg-overview}

Generally speaking, the segmentation method can be considered as a greedy graph-based region merging algorithm. Given an over-segmentation of the image, we define a graph $\G = (\R, \E, W(\E))$, where each node corresponds to one region, and $\R = \{R_1, R_2, \ldots\}$ is the set of all nodes. Further, $\E = \{e_{ij}\}$ is the set of all edges connecting adjacent regions, and the weights $W(\E)$ are a measure of dissimilarity between regions. The algorithm proceeds by sorting the edges by their weights and iteratively merging the most similar regions until certain stopping criterion is met. Each iteration consists of three steps:

\begin{enumerate}
\item Select the edge with minimum weight: $$e^* = \arg\min_{e_{ij}\in \E} W(e_{ij})\;.$$
\item Let $R_1, R_2\in \R$ be the regions linked by $e^*$. Set $\R \leftarrow \R \setminus \{R_1, R_2\} \bigcup \{ R_1\bigcup R_2\}$ and update the edge set $\E$ accordingly.
\item Stop if the desired number of regions $K$ is reached, or the minimum edge weight is above a threshold $\delta$. Otherwise, update weights $W(\E)$ and repeat.  
\end{enumerate}

\begin{figure}[ht!]
\centering
\subfigure[]{
\includegraphics[height =1.02in]{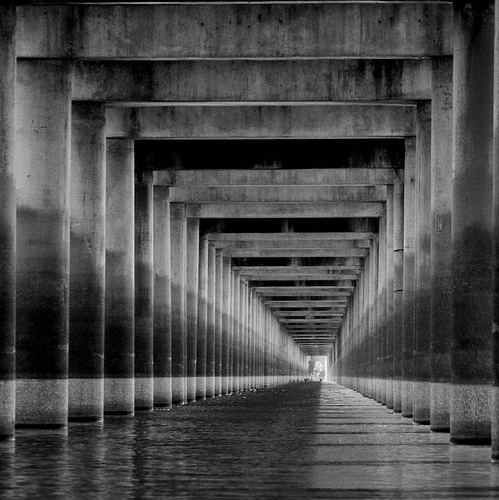}}
\subfigure[]{
\includegraphics[height =1.02in]{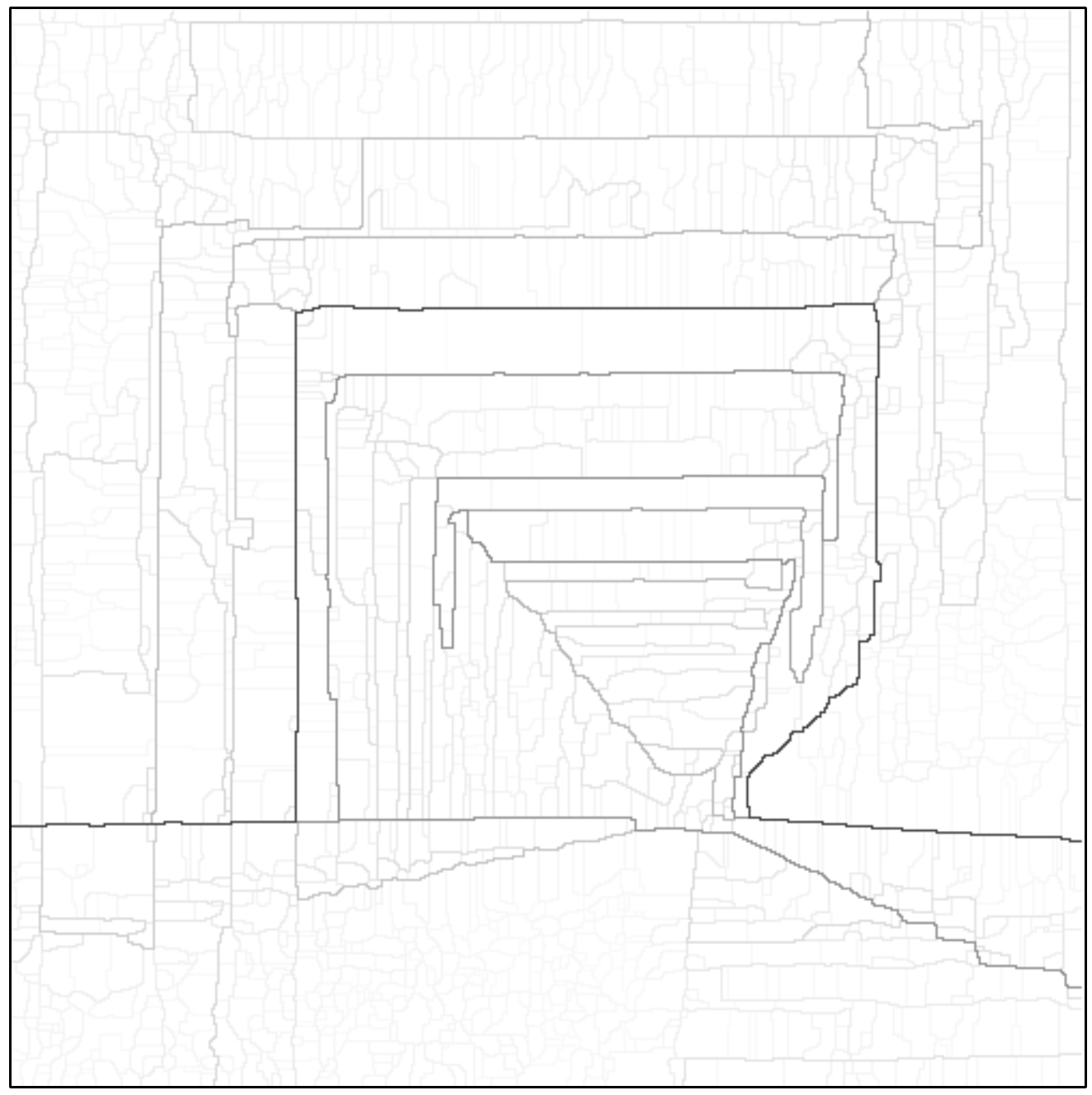}}
\subfigure[]{
\includegraphics[height =1.02in]{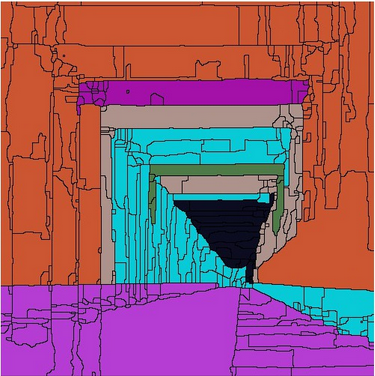}}
\caption{Hierarchical image segmentation using photometric cues only. {\bf (a)} The original image. {\bf (b)} The ultrametric contour map (UCM) generated by~\citet{ArbelaezMFM11}. {\bf (c)} The segmentation result obtained by thresholding the UCM at a fixed scale.}
\label{fig:ucm}
\end{figure}

Various measures have been proposed to determine the distance between two regions, such as the difference between the intensity variance across the boundary and the variance within each region~\citep{FelzenszwalbH04},  and the difference in coding lengths~\citep{MobahiRYSM11}. Recently, Arbelaez {\it et al.} proposed a novel scheme for contour detection which integrates \emph{global photometric information} into the grouping process via spectral clustering~\citeyear{ArbelaezMFM11}. They have shown that this globalization scheme can help identify contours which are too weak to be detected using local cues. The detected contours are then converted into a set of initial regions (\ie, an over-segmentation) for hierarchical image segmentation. We show an example of the segmentation result obtained by~\citet{ArbelaezMFM11} in Figure~\ref{fig:ucm}. In particular, in Figure~\ref{fig:ucm}(b), we visualize the entire hierarchy of regions on an real-valued image called the ultrametric contour map (UCM)~\citep{Arbelaez06}, where each boundary is weighted by the dissimilarity level at which it disappears. In Figure~\ref{fig:ucm}(c), we further show the regions obtained by thresholding the UCM at a fixed scale. It is clear that because the weights of the boundaries are computed only based on the photometric cues in~\citet{ArbelaezMFM11}, different geometric regions could be merged at early stages in the hierarchical segmentation process if they have similar appearances.

Motivated by this observation, we take the over-segmentation result generated by \citet{ArbelaezMFM11} (\ie, by thresholding the UCM at a small scale 0.05) as the input to our algorithm, and develop a new distance measure between regions which takes both photometric and geometric information into consideration.

\subsection{Geometric Distance Measure}
\label{sec:method-measure}

We assume that a major portion of the scene can be approximated by a collection of 3D planes parallel to a dominant direction in the scene. The background, \eg, the sky, can be treated as a plane at infinity. The dominant direction is characterized by a set of parallel lines in the 3D space which, when projected to the image, converge to the dominant vanishing point. Consequently, given the location of the dominant vanishing point, our goal is to segment an image so that each region can be roughly modeled by one plane in the scene. To achieve this goal, we need to formulate a dissimilarity measure which yields small values if the pair of adjacent regions belong to the same plane, and large values otherwise.

\begin{figure}[ht!]
\centering
\includegraphics[height =2in]{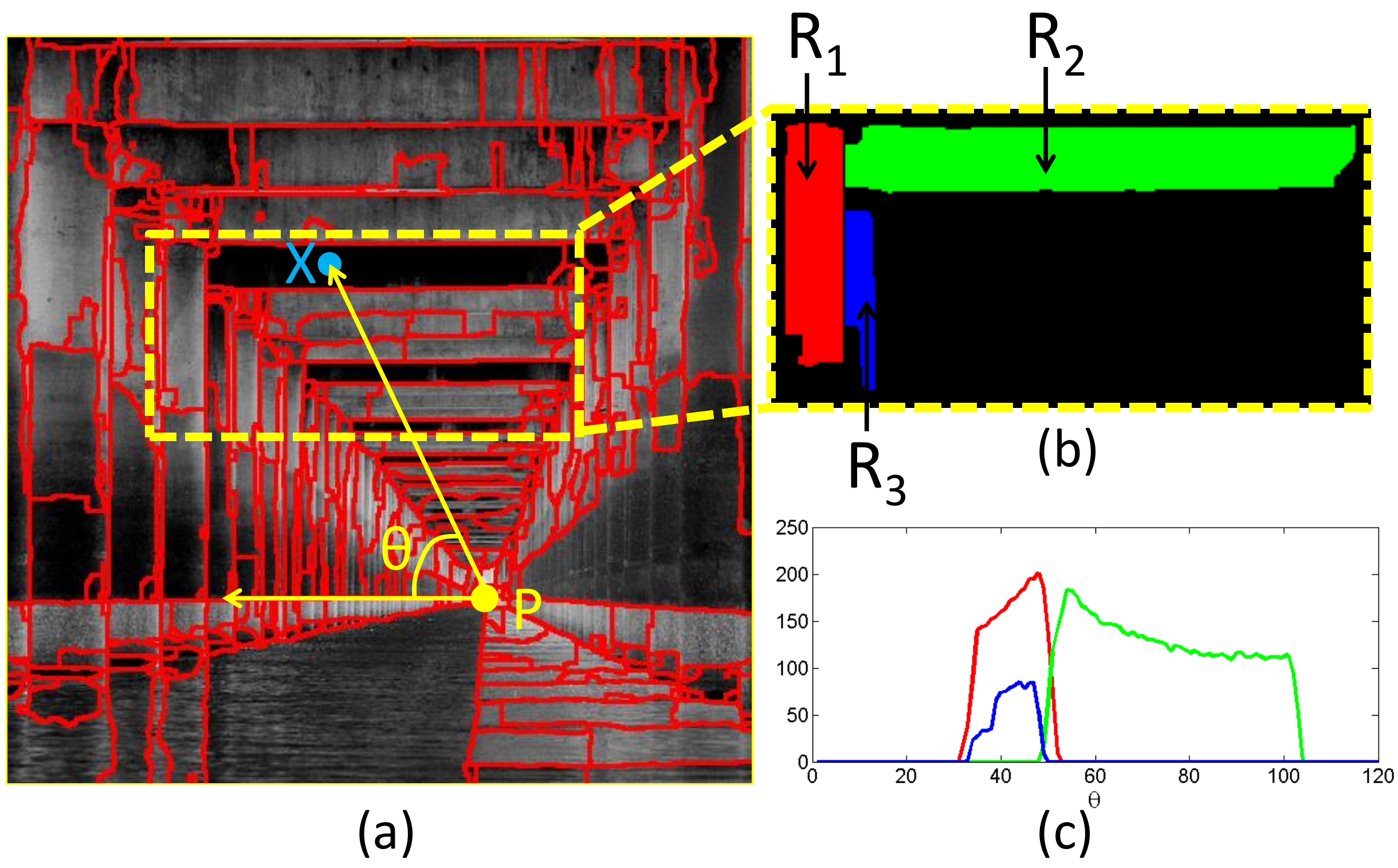}
\includegraphics[height =2in]{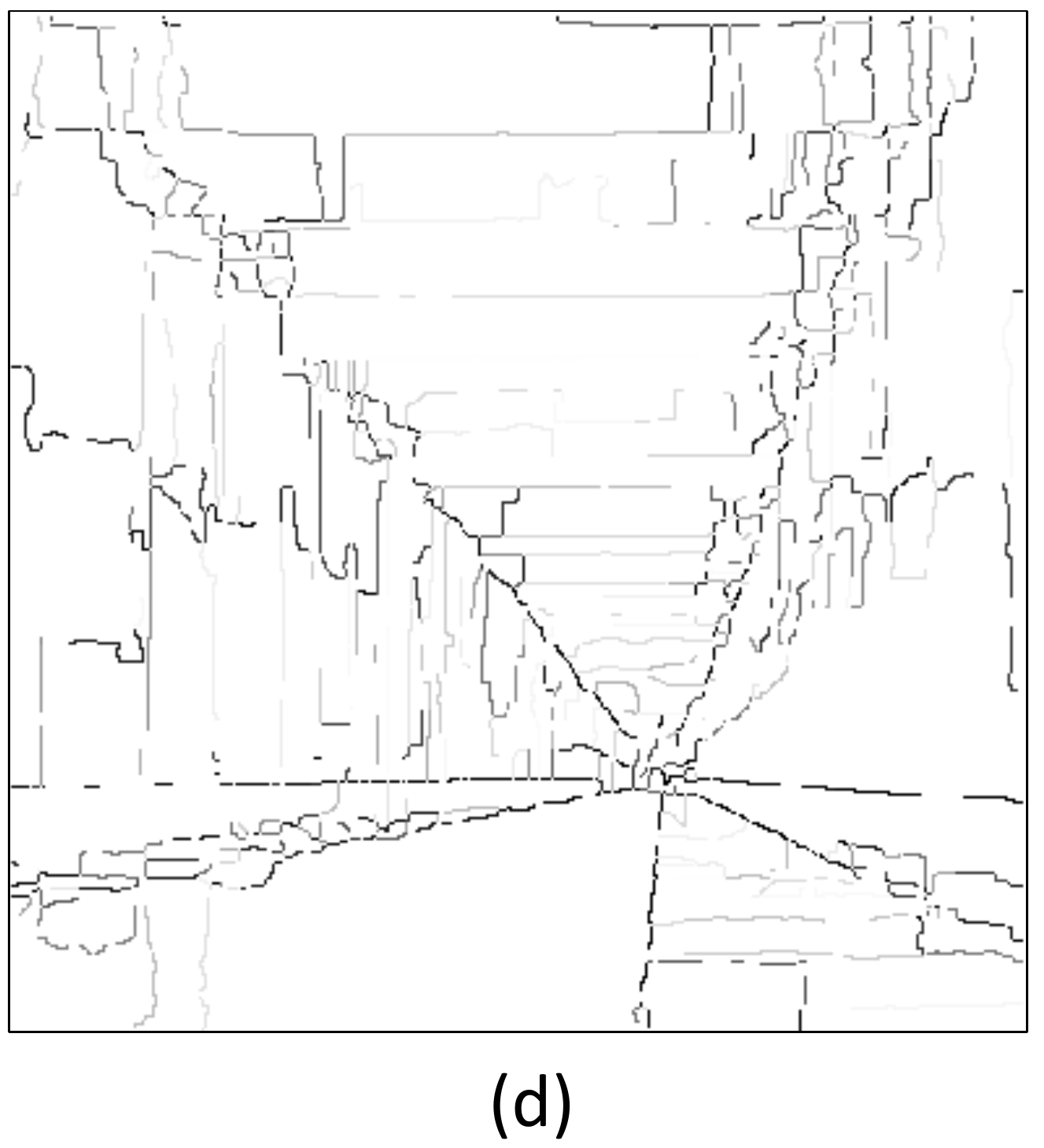}
\caption{Illustration of the the computation of the geometric distance. {\bf (a)} The over-segmentation map with the polar coordinate system. {\bf (b)} Three adjacent regions from the image. {\bf (c)} The histograms of angle values for the three regions. {\bf (d)} The boundary map weighted by the geometric distance between adjacent regions.}
\label{fig:distance}
\end{figure}

We note that any two planes that are parallel to the dominant direction must intersect at a line which passes through the dominant vanishing point in the image. Intuitively, this observation provides us with a natural way to identify adjacent regions that could potentially lie on different planes: If the boundary between two regions is parallel to the dominant direction (hence passes through the dominant vanishing point), these two regions are likely to lie on different planes. However, in the real world, many objects are not completely planar, hence there may not be a clear straight line that passes through the dominant vanishing point between them. As an example, if we focus our attention on the three adjacent regions $R_1$, $R_2$ and $R_3$ in Figure~\ref{fig:distance}, we notice that $R_1$ and $R_3$ belong to the vertical wall and $R_2$ belongs to the ceiling. However, the boundaries between the pair ($R_1$, $R_2$) and the pair ($R_1$, $R_3$) both lie on the same (vertical) line. As a result, it is impossible to differentiate these two pairs based on only the orientation of these boundaries.

To tackle this problem, we propose to look at the angle of each region from the dominant vanishing point in a polar coordinate system, instead of the orientation of each boundary pixel. Here, the angle of a region is represented by the distribution of angles of all the pixels in this region. Mathematically, let the dominant vanishing point $P$ be the pole of the polar coordinate system, for each region $R_i$, we compute the histogram of the angle value $\theta(X)$ for all the pixels $X\in R_i$, as illustrated in Figure~\ref{fig:distance}. 

Let $c_i(\theta)$ be the number of the pixels in $R_i$ that fall into the $\theta$-th bin. We use 360 bins in our experiments. We say that one region $R_i$ \emph{dominates} another region $R_j$ at angle $\theta$ if $c_i(\theta) \geq c_j(\theta)$. Our observation is that if one region $R_i$ always dominates another region $R_j$ at almost all angles, these two regions likely belong to the same plane. Meanwhile, if one region has larger number of pixels at some angles whereas the other region has larger number of pixels at some other angles, these two regions likely lie on different planes.  This observation reflects the fact a plane converging to the vanishing point often divides along the direction perpendicular to the dominant direction because of architectural or natural structures, {\it e.g.}, columns and trees.  Because perpendicular separation of regions has little effect on the polar angles, the histograms of angles tend to overlap substantially.   

Based on this observation, we define the geometric distance between any two regions $R_i$ and $R_j$ as follows:
\begin{eqnarray*}
&&W_g(e_{ij}) = 1-\\
&&\max \left( \frac{\sum_{\theta} \min(c_i(\theta), c_j(\theta))}{|R_i|}, \frac{\sum_{\theta} \min(c_i(\theta), c_j(\theta))}{|R_j|} \right)\;,
\end{eqnarray*}
where $|R_i|$ and $|R_j|$ are the total numbers of pixels in regions $R_i$ and $R_j$, respectively.
For example, as illustrated in Figure~\ref{fig:distance}(c), $R_1$ dominates $R_3$ at all angles and hence we have $W_g(e_{1,3}) = 0$. Meanwhile, $R_1$ and $R_2$ dominate each other at different angles and their distributions have very small overlap. As a result, their geometric distance is large: $W_g(e_{1,2}) = 0.95$. In Figure~\ref{fig:distance}(d), we show all the boundaries weighted by our geometric distance. As expected, the boundaries between two regions which lie on different planes tend to have higher weights than other ones. This suggests that, by comparing the angle distributions of two adjacent regions, we can obtain a more robust estimate of the boundary orientations than directly examining the orientations of boundary pixels. 

Here, a reader may wonder why we don't simply normalize the histograms and use popular metrics like KL divergence or the earth mover's distance to compare two regions. While our intuition is indeed to compare the distributions of angles of two regions, we have found in practice that computing the normalized histograms could be highly unstable for small regions, especially at the early stages of the iterative merging process. Thus, in this paper we propose an alternative geometric distance measure which avoids normalizing the histograms, and favors large regions during the process.

\subsection{Combining Photometric and Geometric Cues}
\label{sec:method-seg}

While our geometric distance measure is designed to separate different geometric structures, \ie, planes, in the scene, the traditional photometric cues often provide additional information about the composition of images. Because different geometric structures in the scene often have different colors or texture, the photometric boundaries often coincide with the geometric boundaries. On the other hand, in practice it may not always be possible to model all the structures in the scene by a set of planes parallel to the dominant direction. Recognizing the importance of such structures to the composition of the image due to their visual saliency, it is highly desirable to integrate the photometric and geometric cues in our segmentation framework to better model composition. In our work, we combine the two cues by a linear combination:
\begin{equation}
W(e_{ij}) = \lambda W_g(e_{ij}) + (1-\lambda) W_p(e_{ij})\;,
\label{eq:dist}
\end{equation}
where $W_p(e_{ij})$ is the photometric distance between adjacent regions, and can be obtained from any conventional hierarchical image segmentation method. Here we adopt the contour map generated by~\citet{ArbelaezMFM11}.

\begin{figure}[ht!]
\centering
\begin{tabular}{cccc}
\includegraphics[height =0.48in]{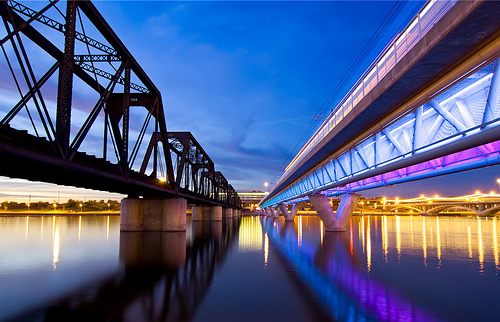}&
\hspace{-3mm}\includegraphics[height =0.48in]{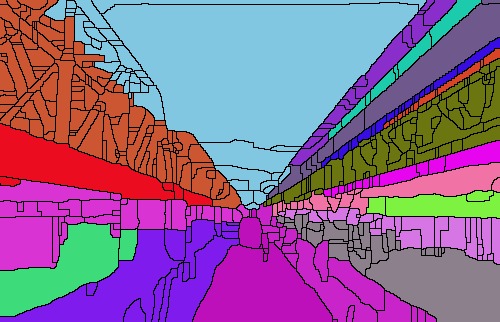}&
\hspace{-3mm}\includegraphics[height =0.48in]{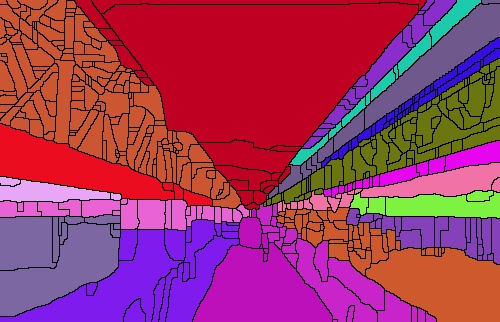}&
\hspace{-3mm}\includegraphics[height =0.48in]{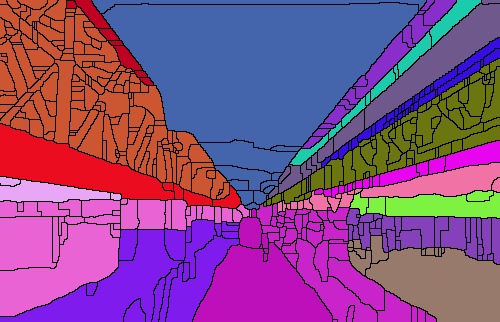}\\
&\hspace{-3mm}$\lambda = 1$ & \hspace{-3mm}$\lambda = 0.8$ & \hspace{-3mm}$\lambda = 0.6$\\
&\hspace{-3mm}\cfbox{blue}{\includegraphics[height =0.48in]{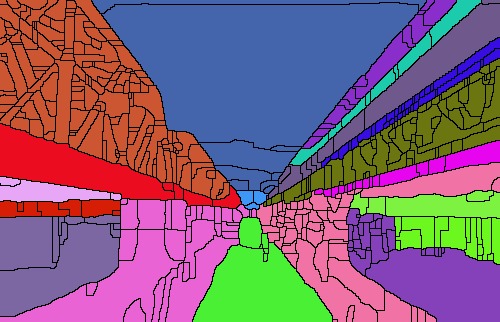}}&
\hspace{-3mm}\includegraphics[height =0.48in]{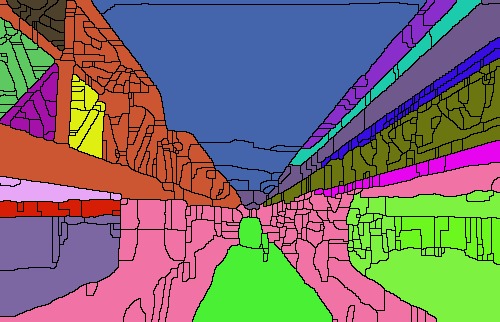}&
\hspace{-3mm}\includegraphics[height =0.48in]{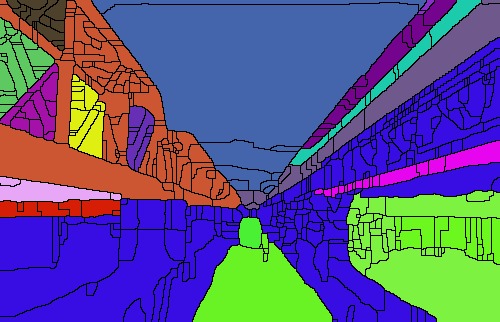} \\
& \hspace{-3mm}$\lambda = 0.4$ & \hspace{-3mm}$\lambda = 0.2$ & \hspace{-3mm}$\lambda = 0$\\
\end{tabular}
\caption{Image segmentation results by integrating the photometric and 
geometric cues. Different weighting parameter $\lambda$ have been used.}\vspace{-2mm}
\label{fig:seg-lambda}
\end{figure}

\begin{figure}[t!]
\centering
\includegraphics[height =0.4in]{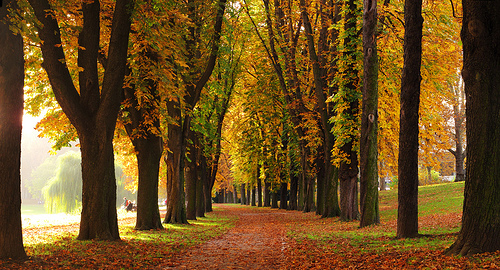}\hspace{1mm}
\includegraphics[height =0.4in]{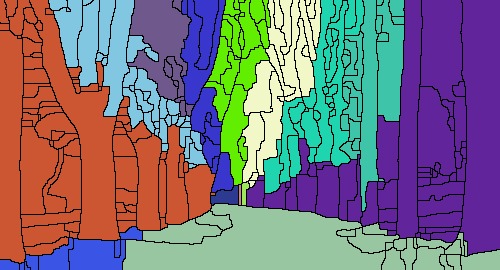}\hspace{1mm}
\includegraphics[height =0.4in]{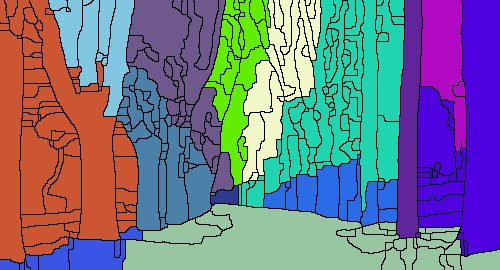}\hspace{1mm}
\includegraphics[height =0.4in]{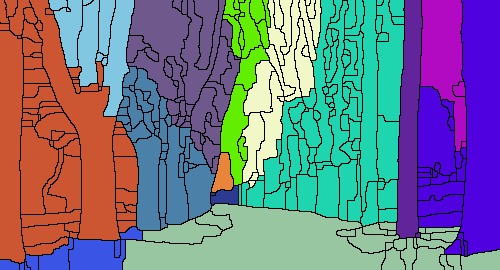}\hspace{1mm}
\\
\vskip 0.05in
\cfbox{blue}{\includegraphics[height =0.4in]{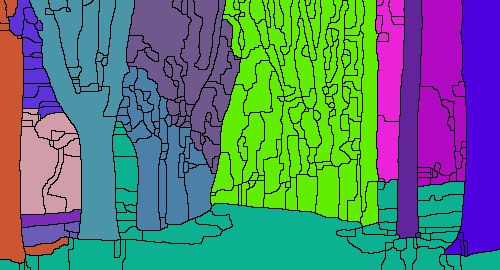}}\hspace{1mm}
\includegraphics[height =0.4in]{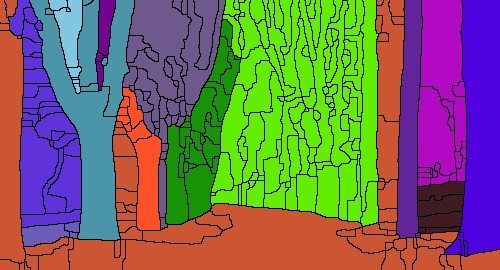}\hspace{1mm}
\includegraphics[height =0.4in]{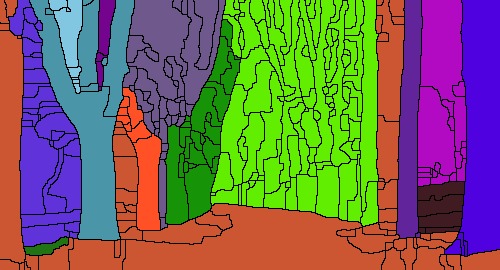} \\
\vskip 0.05in
\includegraphics[height =0.55in]{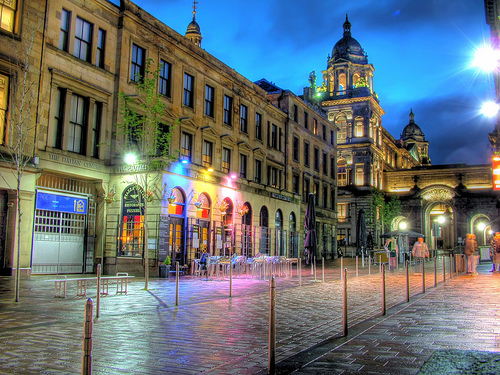}\hspace{1mm}
\includegraphics[height =0.55in]{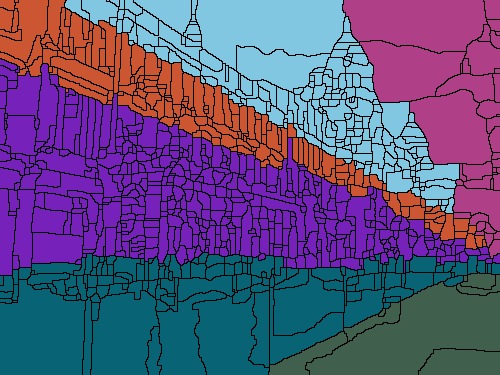}\hspace{1mm}
\includegraphics[height =0.55in]{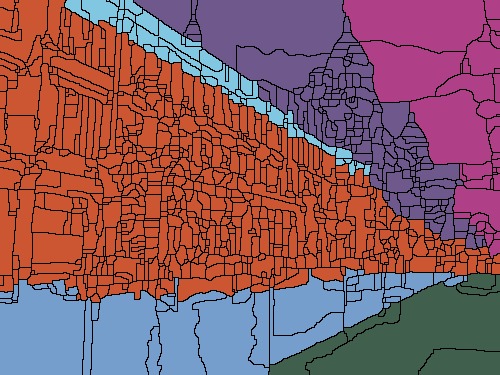}\hspace{1mm}
\includegraphics[height =0.55in]{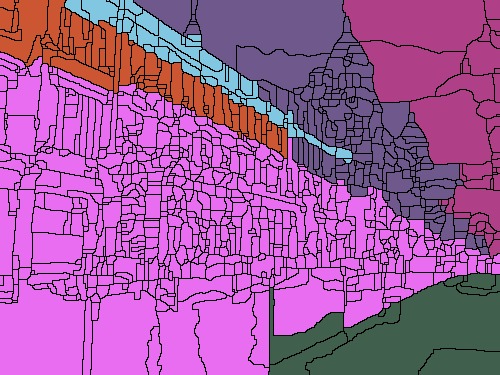}\hspace{1mm}
\\
\vskip 0.05in
\cfbox{blue}{\includegraphics[height =0.55in]{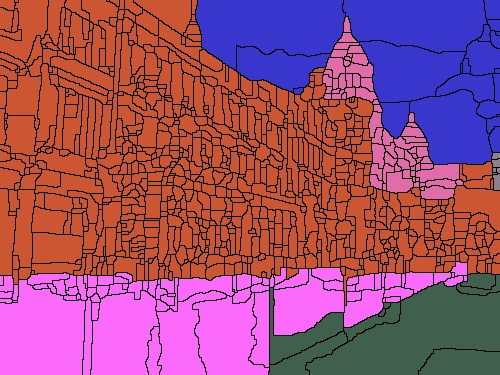}}\hspace{1mm}
\includegraphics[height =0.55in]{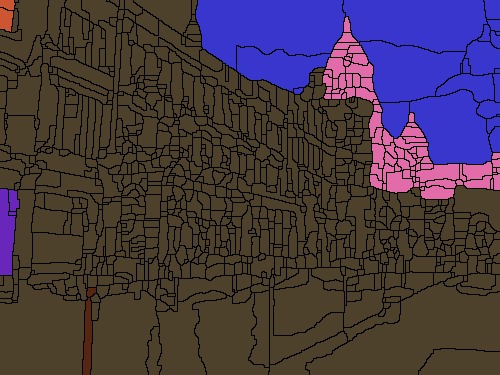}\hspace{1mm}
\includegraphics[height =0.55in]{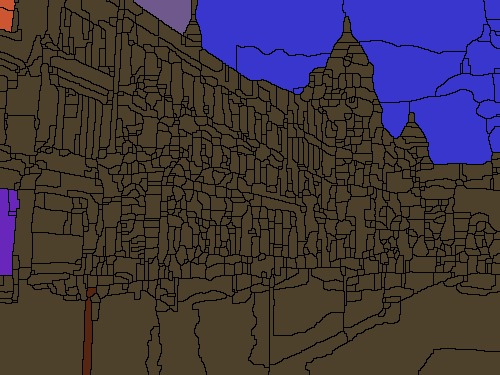} \\
\vskip 0.05in
\includegraphics[height = 0.515in]{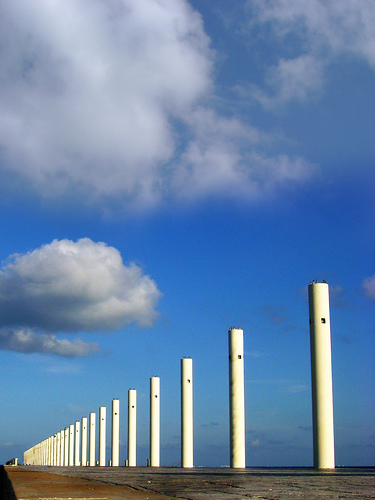}\hspace{1mm}
\includegraphics[height = 0.515in]{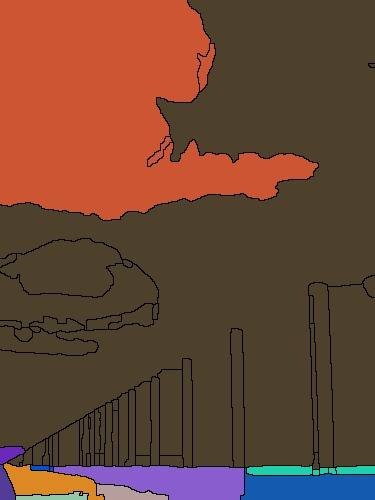}\hspace{1mm}
\includegraphics[height = 0.515in]{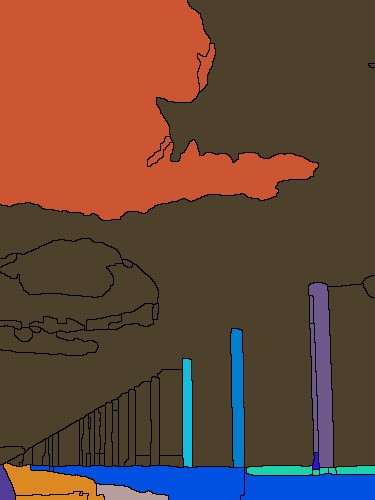}\hspace{1mm}
\cfbox{blue}{\includegraphics[height = 0.515in]{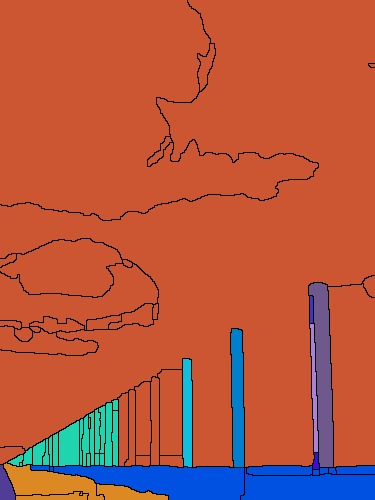}}\hspace{1mm}
\cfbox{blue}{\includegraphics[height = 0.515in]{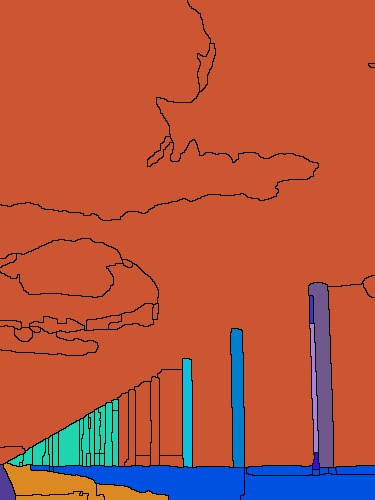}}\hspace{1mm}
\includegraphics[height = 0.515in]{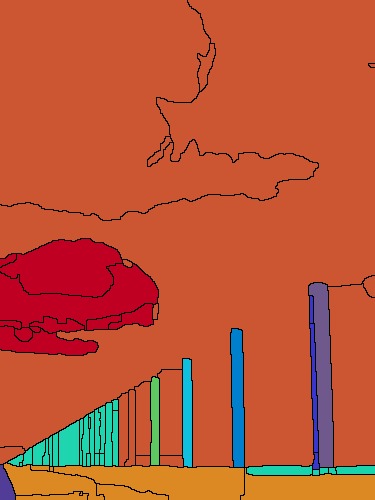}\hspace{1mm}
\includegraphics[height = 0.515in]{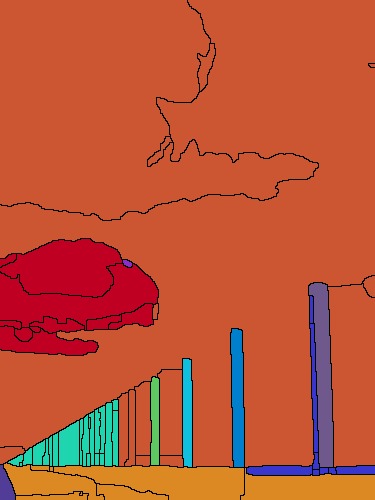} \\
\vskip 0.05in
\includegraphics[height = 0.585in]{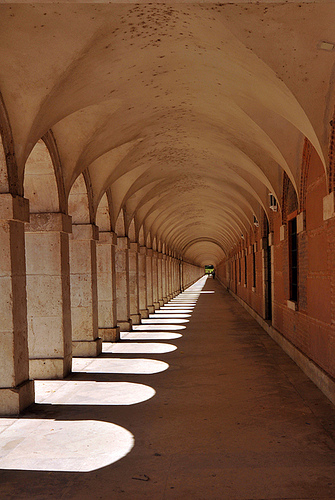}\hspace{1mm}
\includegraphics[height = 0.585in]{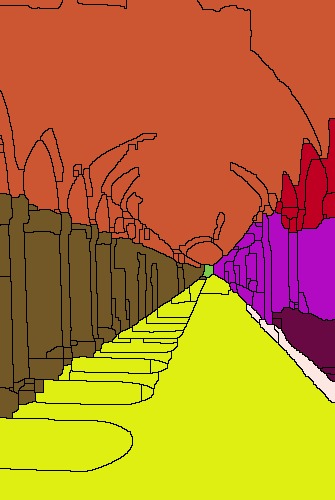}\hspace{1mm}
\includegraphics[height = 0.585in]{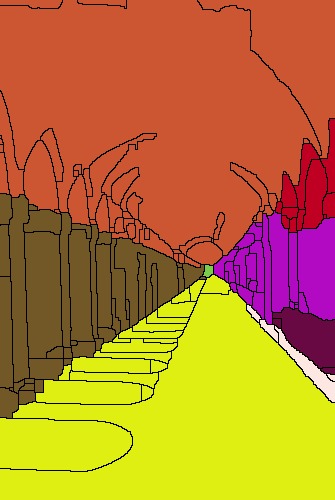}\hspace{1mm}
\includegraphics[height = 0.585in]{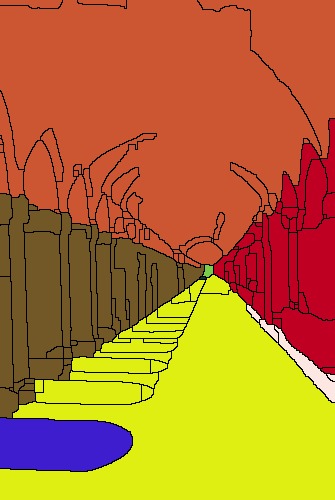}\hspace{1mm}
\cfbox{blue}{\includegraphics[height = 0.585in]{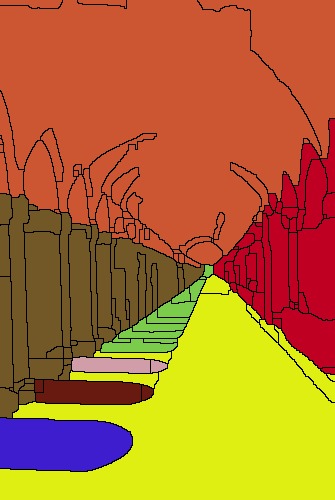}}\hspace{1mm}
\includegraphics[height = 0.585in]{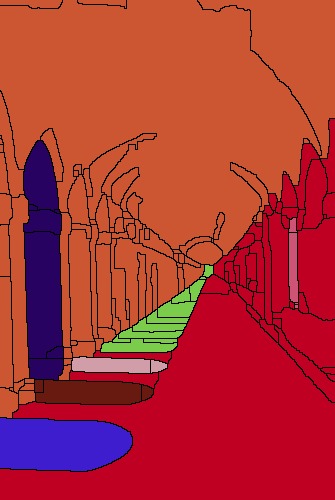}\hspace{1mm}
\includegraphics[height = 0.585in]{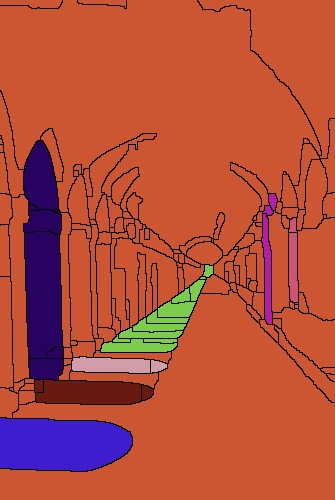} 
\caption{Additional segmentation results. For each original image, we show the results in the order of $\lambda = $1,  $0.8$, $0.6$, $0.4$, $0.2$ and $0$.}
\label{fig:seg-results-1}
\end{figure}

In Figure~\ref{fig:seg-lambda}, we show the segmentation results of an image using our method with different choices of $\lambda$ and a fixed number of regions $K$. Note that when $\lambda=1$, only the geometric cues are used for segmentation; when $\lambda=0$, the result is identical to that obtained by the conventional method~\citep{ArbelaezMFM11}. It can be seen that using the geometric cues alone ($\lambda=1$), we are able to identify most of the structures in the scene. Some of the boundaries between them may not be accurate enough (\eg, the boundary between the bridge on the left and the sky area). However, when $\lambda=0$, the algorithm tends to merge regions from different structures early in the process if they have similar colors. By combining the two cues (\eg, $\lambda=0.4$), we are able to eliminate the aforementioned problems and obtain satisfactory result. Additional results are provided in Figure~\ref{fig:seg-results-1}. Our method typically achieves the best performance when $\lambda$ is in the range of $[0.4, 0.6]$, as highlighted with blue boxes in Figures~\ref{fig:seg-lambda} and~\ref{fig:seg-results-1}. We fix $\lambda$ to 0.6 for the remaining experiments.

\subsection{Enhancing Vanishing Point Detection}

\begin{figure}[ht!]
\centering
\begin{tabular}{cc}
\includegraphics[height =0.93in]{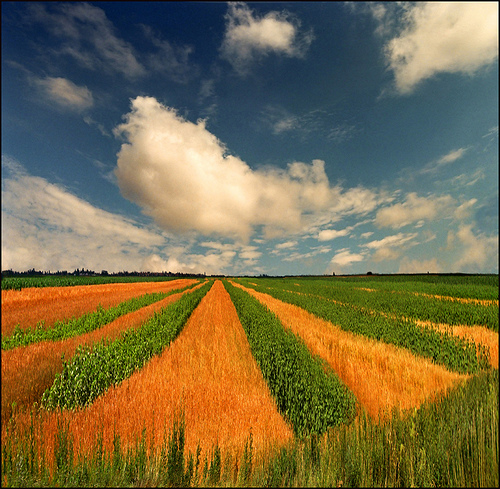} &
\includegraphics[height =0.93in]{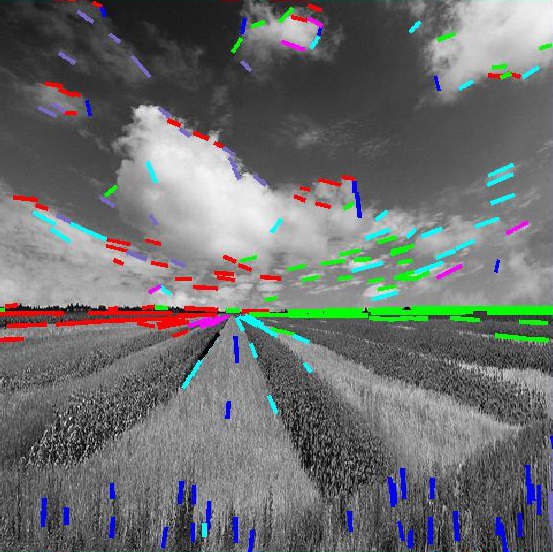} \\
(a) & (b)
\end{tabular}
\vskip 0.05in
\begin{tabular}{ccc}
\hspace{-1mm}\cfbox{gray}{\includegraphics[height =0.93in]{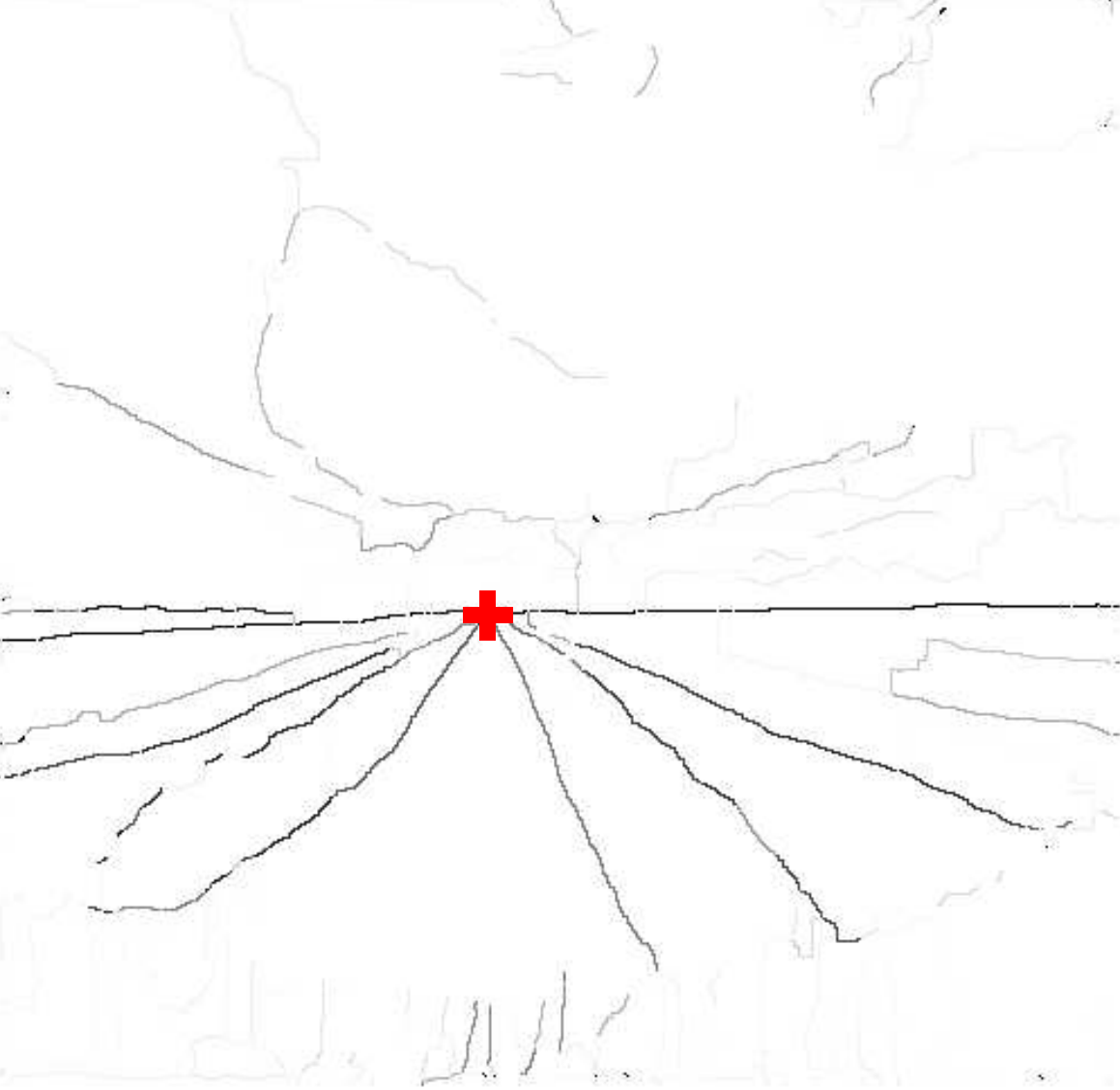}} &
\hspace{-1mm}\cfbox{gray}{\includegraphics[height =0.93in]{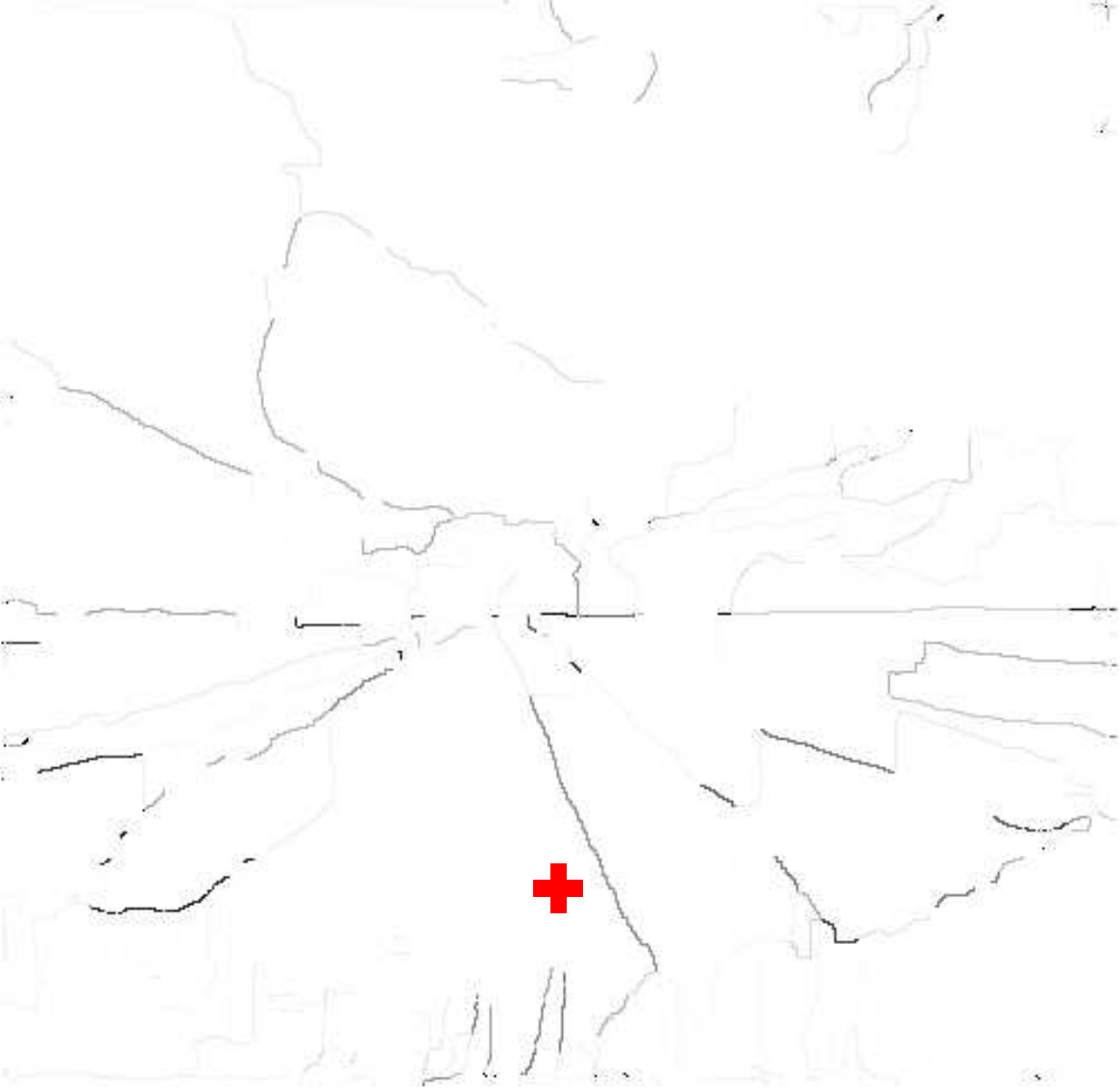}} &
\hspace{-1mm}\cfbox{gray}{\includegraphics[height =0.93in]{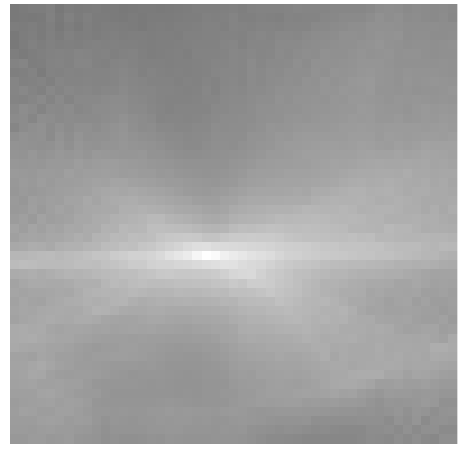}} \\
(c) & (d) & (e)
\end{tabular}
\caption{Enhancing vanishing point detection. {\bf (a)} Original image. {\bf (b)} Line segment detected. {\bf (c)} and {\bf (d)} The weighted boundary map for two different hypotheses of the dominant vanishing point location. {\bf (e)} The consensus score for all vertices on the grid.}
\label{fig:vp}
\end{figure}

In the previous subsection we demonstrated how the knowledge about the dominant vanishing point in the scene can considerably improve the segmentation results. However, detecting the vanishing point in an arbitrary image itself is a challenging problem. Most existing methods assume that (1) region boundaries in the image provide important photometric cues about the location of the dominant vanishing point, and (2) these cues can be well captured by a large number of line segments in the image. In practice, we notice that while the first assumption is generally true, the second one often fails to hold, especially for images of natural outdoor scenes. This is illustrated in Figure~\ref{fig:vp}: although human can easily infer the location of the dominant vanishing point from the orientations of the aggregated region boundaries, existing line segment detection algorithms may fail to identify these boundaries. For this reason, any vanishing point detection method relying on the detected line segments would also fail. 

\label{sec:method-vp}

To alleviate this issue, we propose to use our geometric distance measure $W_g(e_{ij})$ to obtain a more robust estimation of the orientation of each boundary and subsequently develop a simple exhaustive search scheme to detect the dominant vanishing point. In particular, given a hypothesis of the dominant vanishing point location, we can obtain a set of boundaries which align well with the converging directions in the image by computing $W_g(e_{ij})$ for each pair of adjacent regions. These boundaries then form a ``consensus set''. We compute a score for the hypothesis by summing up the strengths of the boundaries in the consensus set (Figure~\ref{fig:vp}(c) and (d)). Finally, we keep the hypothesis with the highest score as the location of the dominant vanishing point (Figure~\ref{fig:vp}(e)). Our algorithm can be summarized as follows:


\begin{enumerate}
\item Divide the image by an $m\times n$ uniform grid mesh.
\item For each vertex $P_k$ on the grid, we compute the geometric distance $W_g(e_{ij})$ for all the boundaries in an over-segmentation of the image. The consensus score for $P_k$ is defined as:
$\displaystyle f(P_k) = \sum_{e_{ij}\in \E} W_p(e_{ij}) W_g(e_{ij})\;.$
\item Select the point with the highest score as the detected dominant vanishing point:\\
 $P^* = \arg\max f(P_k)\;.$
\end{enumerate}
Here, the size of the grid may be chosen based on the desired precision for the location of the vanishing point. In practice, our algorithm can find the optimal location in about one minute on a $50\times 33$ grid on a single CPU. We also note that the time may be reduced using a coarse-to-fine procedure.

In addition, we assume that the dominant vanishing point lies in the image frame because, as we noted before, only the vanishing point which lies within or near the frame conveys a strong sense of 3D space to the viewer. But our method can be easily extended to detect vanishing points outside the frame using a larger mesh grid.

\section{Detecting and Modeling Triangles in Portrait Photography}\label{sec:por}

In this section, we describe the algorithm for handling portrait
photos.  We first introduce the line segment detection algorithm (Section~\ref{sec:por-1}),
and then discuss how triangles can be constructed from these line
segments (Section~\ref{sec:por-2}).

\subsection{Line Segment Detection}
\label{sec:por-1}
By examining high-quality portraits designed with the triangle
technique (\eg, see Figures~\ref{fig:results_scs} and~\ref{fig:results}), we observe that triangles present in portrait photographs are often
composed with parts of contours or edges, such as the contours of
arms, body parts, wearing apparels, as well as edges formed by
multiple human subjects. In general, a triangle is geometrically
defined as a polygon with three corners and three sides, where the
three sides are all straight line segments. However, contours of
natural objects like humans or hats are often slightly
curved. Therefore, to detect potential triangles in real images, our
method needs to be able to identify such curved line
segments, in addition to straight edges, in an image.

To this end, we employ the Line Segment Detector (LSD) proposed by~\citet{von2012lsd} to convert gradient map of an image to a set
of line segments. It first calculates a
level-line angle at each pixel to produce a \emph{level-line
  field}. The level line is a straight line perpendicular to the
gradient at each pixel. Then, the image is partitioned into \emph{line-support regions} by grouping connected pixels that share the same angle up to a certain tolerance. Each line-support region is treated as a candidate line segment. Next, a hypothesis testing framework is developed to test each line segment candidate. The framework approximates each line-support region with a rectangle and compare the number of ``aligned points'' in each rectangle in the original image with the \emph{expected} number of aligned points in a random image. A line segment is detected if the actual number of aligned points in a rectangle is significantly larger than the expected number.

\begin{figure}[ht!]
  \centering 
  \subfigure[original]{ 
    \label{fig:denori} 
    \includegraphics[width=0.11\textwidth]{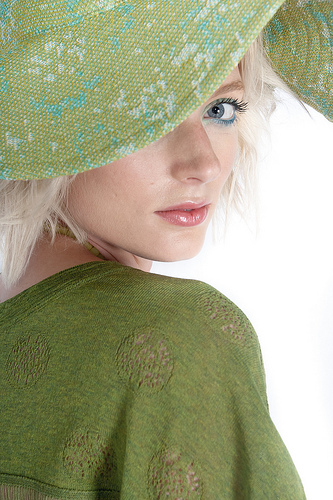}} 
  \subfigure[0.7]{ 
    \label{fig:den0.7} 
    \includegraphics[width=0.11\textwidth]{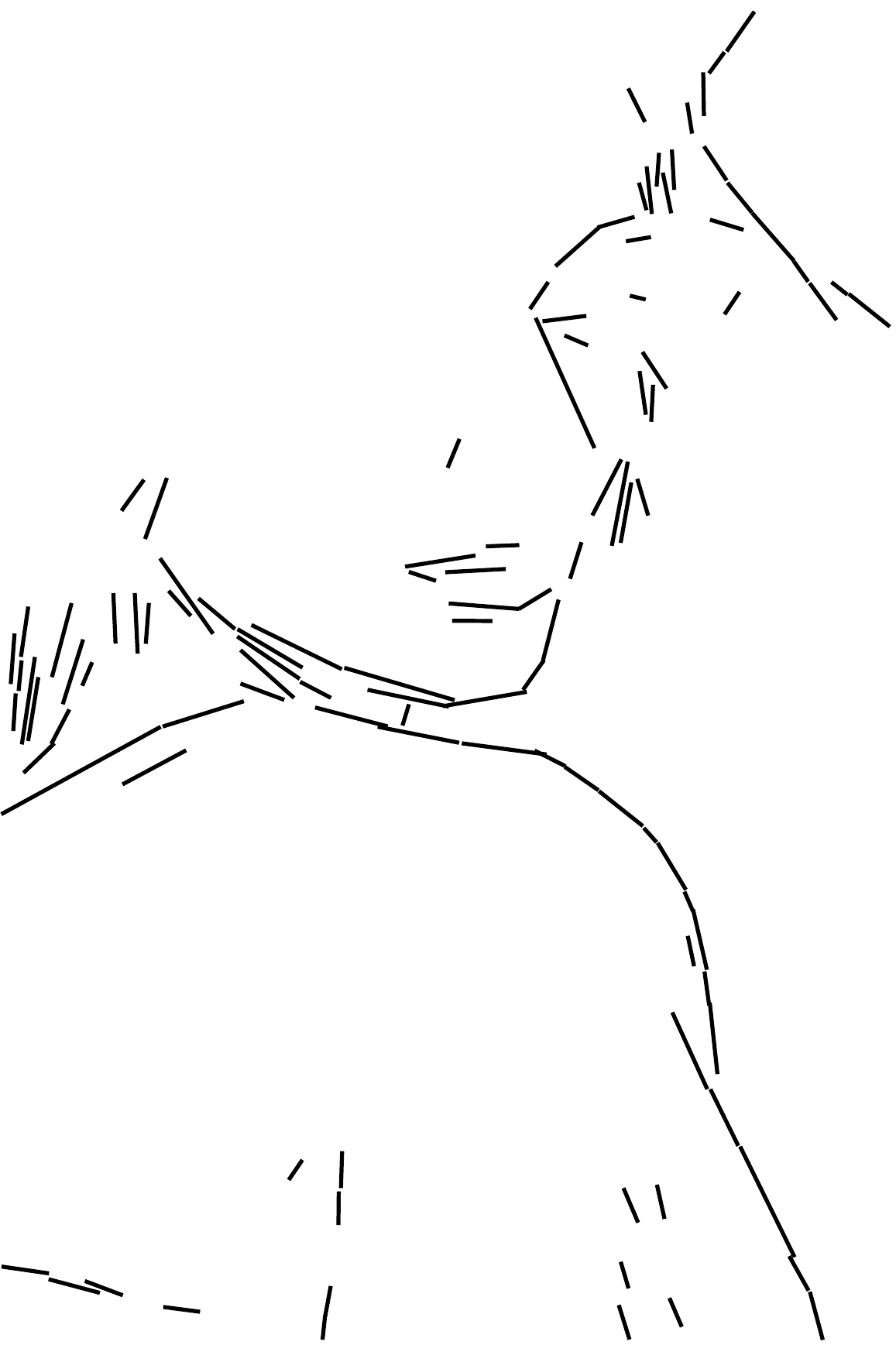}} 
  \subfigure[0.5]{ 
    \label{fig:den0.5} 
    \includegraphics[width=0.11\textwidth]{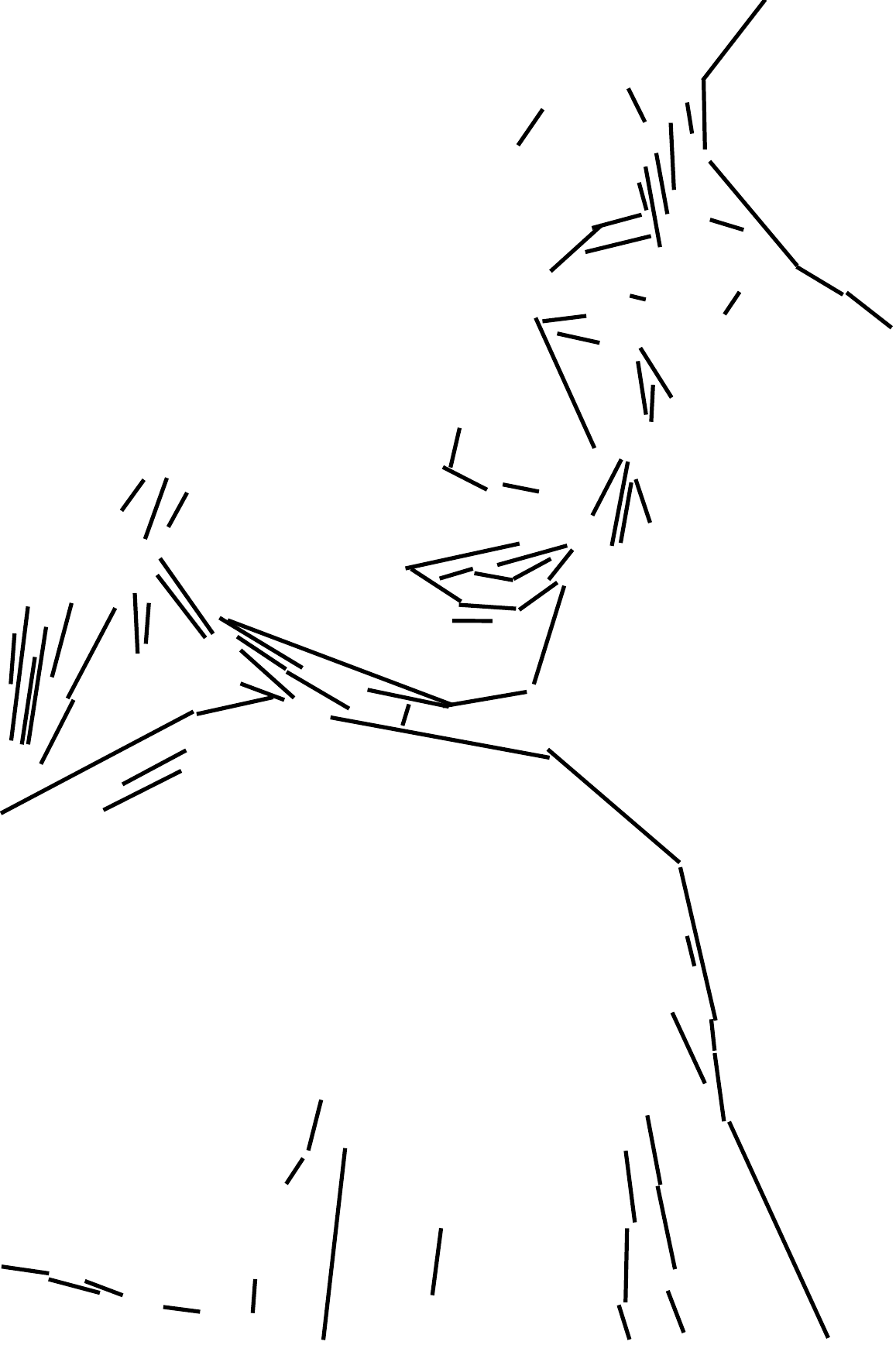}} 
  \subfigure[0.2]{ 
    \label{fig:den0.2} 
    \includegraphics[width=0.11\textwidth]{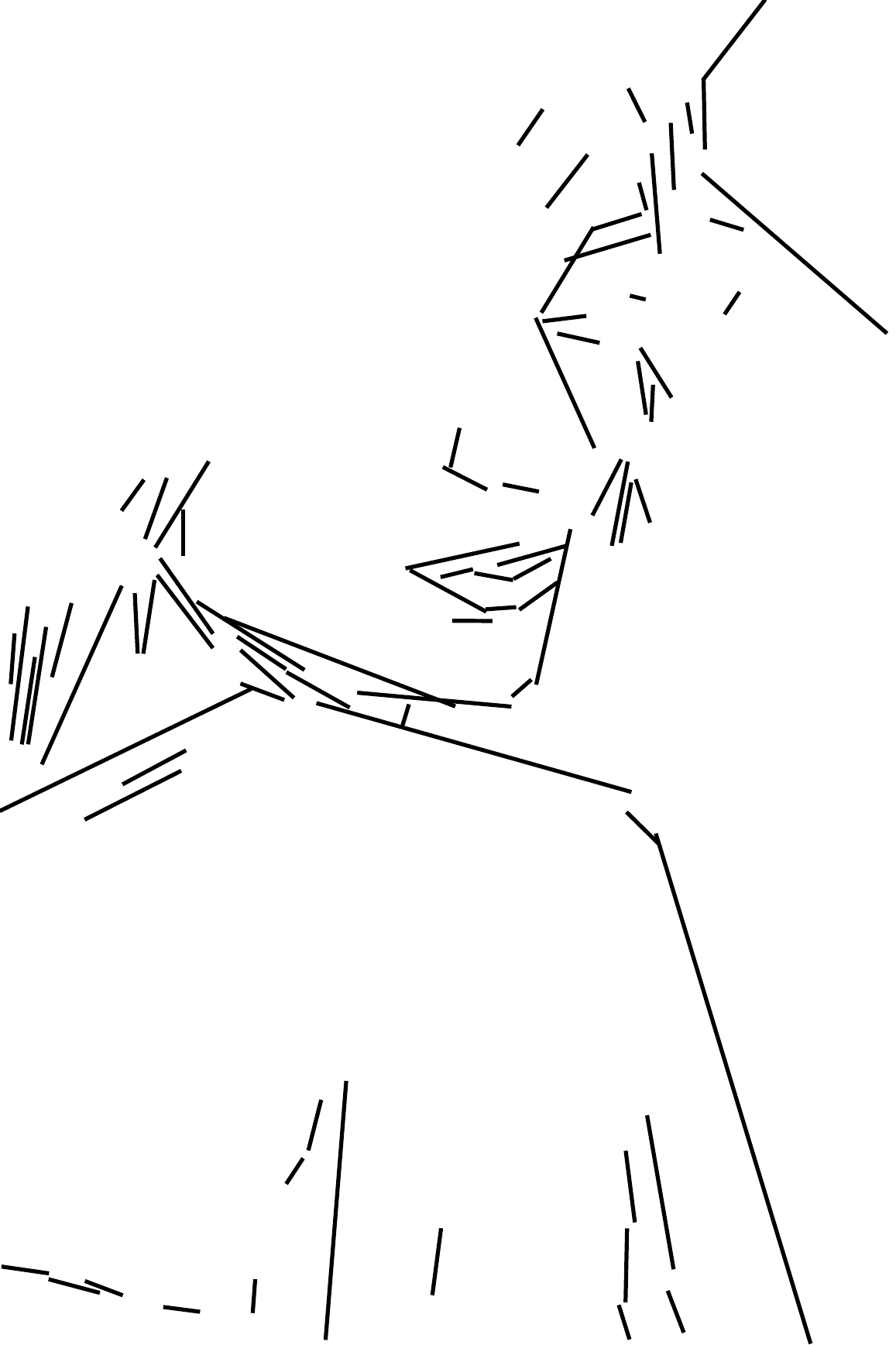}} 
  \caption{LSD results with different density (as indicated).} 
  \label{fig:lsd_diffden} 
\vskip -0.1in
\end{figure}

\begin{figure*}[ht!]
  \centering 
  \subfigure[original image]{ 
    \label{fig:flsdorig} 
    \includegraphics[width=0.18\textwidth]{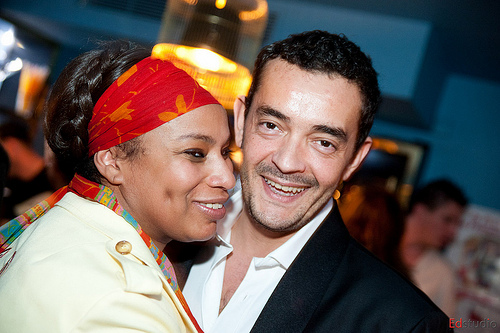}} 
  \subfigure[contour]{ 
    \label{fig:contour} 
        \cfbox{gray}{\includegraphics[width=0.18\textwidth]{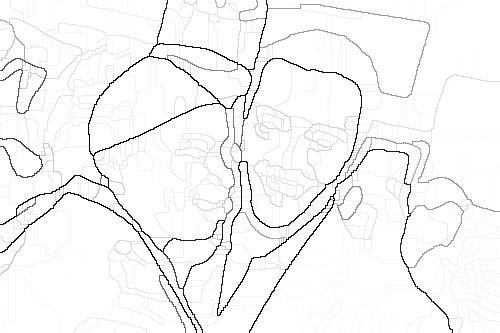}}}
  \subfigure[LSD result]{ 
    \label{fig:lsd} 
        \cfbox{gray}{\includegraphics[width=0.18\textwidth]{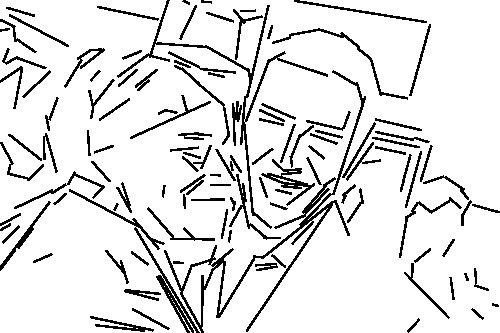}}}
  \subfigure[filtered LSD]{ 
    \label{fig:filter1} 
        \cfbox{gray}{\includegraphics[width=0.18\textwidth]{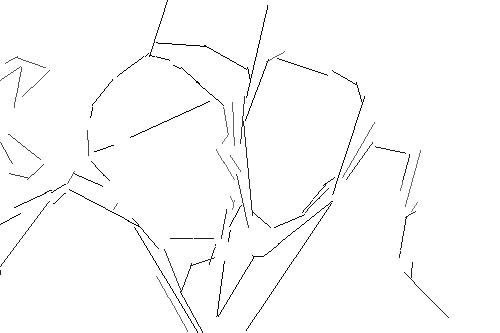}}} 
 \subfigure[filtered LSD]{ 
    \label{fig:filter2} 
        \cfbox{gray}{\includegraphics[width=0.18\textwidth]{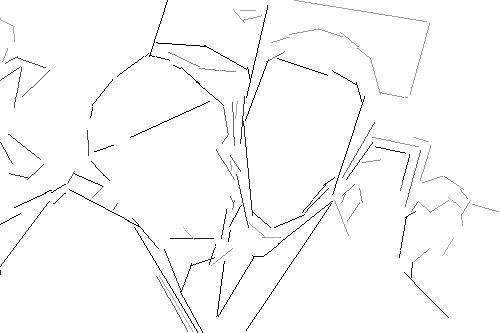}}}
  \caption{Filtering the line segments. The parameter $\alpha$ is set to $0.5$ and $0.7$ in (d) and (e), respectively.} 
  \label{fig:filteredlsd} 
\end{figure*}

A useful property of LSD is that, by approximating the line-support region using a rectangle of certain length, it is able to detect near-straight curves in the image. 
Further, it is easy to see that a larger rectangle with more unaligned points will be needed to cover a more curved line segment. Therefore, by setting a threshold on the \emph{density} of a rectangle, which is defined as the proportion of aligned points in the rectangle, we can control the degree up to which a curved line segment is considered.
Figure~\ref{fig:lsd_diffden} shows results of the line segment
detector with different density values. In our experiments, the density is empirically set
to 0.2.

While the line segment detector aims at extracting all potential line
segments from an image using \emph{local} image gradient cues, some line segments are more
\emph{globally} distinguishable, thus more visually attractive to viewers.
To further identify such line segments,
we combine the line segment detector with the ultrametric contour
map (UCM) obtained by the same image segmentation algorithm~\citep{ArbelaezMFM11} we introduced in Section~\ref{sec:seg-overview}. Note that each pixel on the contour map holds a confidence
level between 0 and 1, indicating the possibility of it being on a
boundary. Given a line segment, we identify all the pixels falling in
its support region and consider the maximum
confidence level of all pixels as the confidence level of the line
segment. Then, line segments with confidence levels under a certain
threshold are removed, where the threshold is chosen based on the maximum
confidence level present in an image. Specifically, assume the maximum
confidence level of all the line segments in an image is $C$, where $C \in
[0, 1]$, then we set the threshold as $(1-\alpha)C$ and accept line
segments whose confidence levels are within the range $[(1-\alpha)C,
C]$. The parameter $\alpha$ controls the number of accepted line
segments. Smaller $\alpha$ filters out more line segments from an
image, as shown in Figure~\ref{fig:filteredlsd}. In this paper, we empirically set $\alpha=0.5$.

\subsection{Fitting Triangles}
\label{sec:por-2}

The line segment detector described above gives us a set of candidate
triangle sides. Randomly selecting three non-parallel sides from the
set generates
a triangle.\break Thus, the problem of detecting a triangle can
be converted into finding three non-parallel sides. However, although a triangle consists of
three sides, we observe that a triangle can be uniquely determined as long as two sides are found, because the third side can be
obtained by connecting the end points of the other two sides. Moreover, the presence of the
third side is not as important in the practical usage of triangle technique
because viewers can easily ``complete'' the geometric
shapes themselves. As a result, our problem is reduced to fitting a triangle using two non-parallel line segments
selected from the candidate set. 

\begin{figure}[t!]
  \centering 
  \subfigure[]{
    \label{fig:outlier1} 
    \cfbox{gray}{\includegraphics[height=0.75in]{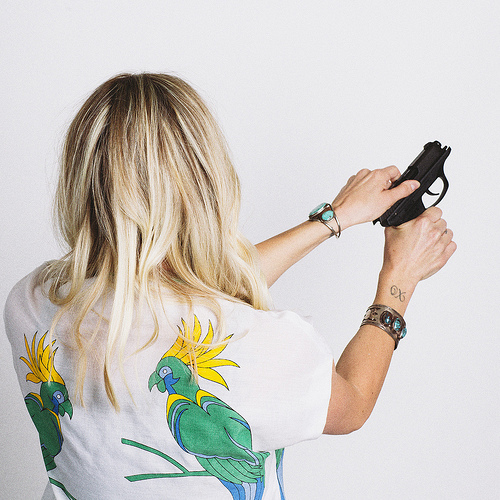}}}
  \subfigure[]{
    \label{fig:outlier2} 
    \includegraphics[height=0.75in]{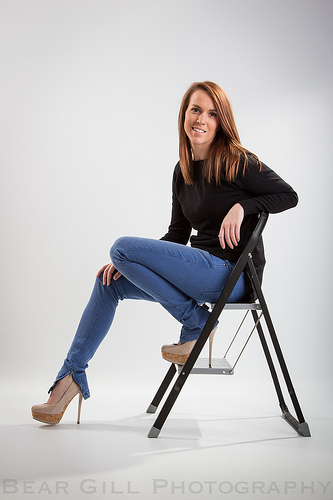}} 
  \subfigure[]{
    \label{fig:impfside1} 
    \includegraphics[height=0.75in]{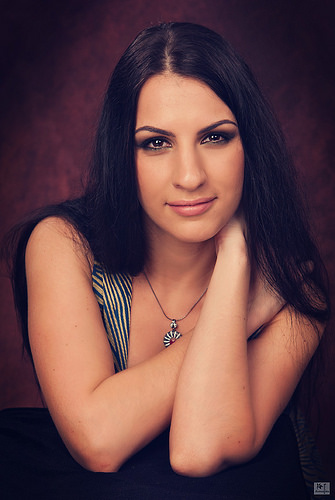}} 
  \subfigure[]{
    \label{fig:impfside2} 
    \includegraphics[height=0.75in]{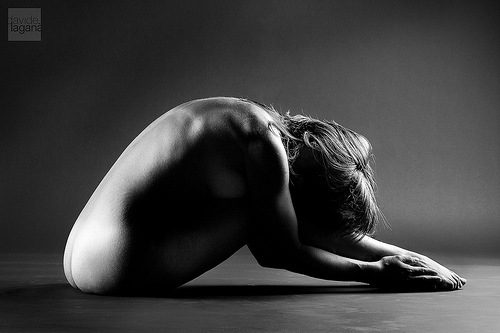}} 
  \caption{Challenges in detecting triangles: Outliers and imperfect sides.
(a) outliers: irrelevant objects, (b) outliers: multiple triangles, (c) imperfect sides: occlusion, (d) imperfect sides: bending curve.} 
  \label{fig:challenges} 
\end{figure}

Two major challenges still exist in fitting triangles using the extracted line
segments: (1) there is a large number of irrelevant line segments; and
(2) the sides of a triangle are imperfect in real images. For example, as shown in Figure~\ref{fig:outlier1}, some
objects in a photograph are irrelevant to the use of triangle technique, even
though they have high-contrast contours ({\it e.g.}, the bird pattern on the woman's
shirt). The line segments produced by such objects are all
outliers. In addition, multiple triangles often exist in
one image ({\it e.g.}, Figure~\ref{fig:outlier2}). Thus, line segments from different triangles should also be considered as outliers w.r.t. each other. 
In Figures~\ref{fig:impfside1} and~\ref{fig:impfside2}, we further show some examples of imperfect triangle sides. As shown, these sides may be
broken into several parts because of occlusion or artificial effect
introduced by the line segment detector. For instance, the girl in
Figure~\ref{fig:impfside1} has her right arm occluded by her left arm. As a result,
the contour of her right arm is broken into two line segments. In Figure~\ref{fig:impfside2}, the contour of the back of the human subject is too
curved to be approximated by a single straight line. Therefore, the
line segment detector approximates it using two straight line
segments with slightly different orientations. 

In order to tackle these two challenges, we employ a
modified RANSAC (RANdom SAmple Consensus) algorithm in favor of its
insensitivity to outliers. RANSAC is an iterative method that robustly fits a set of observed data points
(including outliers) to a pre-defined model. Our algorithm
includes three steps:
\begin{enumerate}
\item Two non-parallel line segments are randomly selected from the
  candidate set and extended to generate two lines on which the
  two triangle sides lie.
\item All the candidate line segments within neighborhoods of
  the two lines are projected onto the correspond lines, resulting in a
  number of projected pixels. Triangle sides are then constructed from all the projected pixels.
\item Once two triangle sides are constructed, two metrics \emph{Continuity Ratio} and \emph{Total Ratio} are calculated to measure the fitness
and significance of the triangle, respectively. Triangles with high scores are
accepted.
\end{enumerate}

In the remainder of this subsection, we describe each step in details.

\subsubsection{Identifying Sides From Line Segments}
By extending the two randomly selected line segments to two
lines, we obtain the shared end point of the two sides, {\it i.e.}, the
intersection of the lines. Moreover, two intersecting lines
  generate four different angles with four different opening directions:
  upwards, downwards, leftwards, and rightwards. Each angle
  corresponds to a category of triangles that contain this
  angle and two sides of varied lengths. Given one of the four angles,
  once the lengths of its two sides are determined, a
  unique triangle can be constructed.
  
  \begin{figure}[t!]
  \centering 
  \subfigure[]{
    \label{fig:rorig} 
    \cfbox{gray}{\includegraphics[height=1.8in]{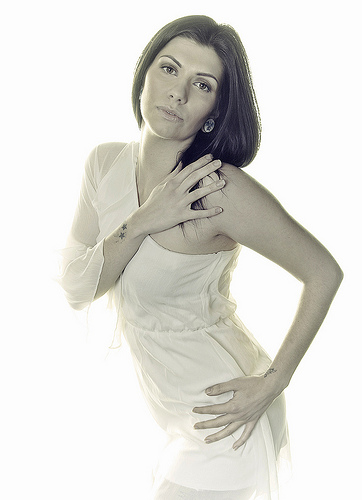}}}\hspace{-1mm}
  \subfigure[]{
    \label{fig:candside}
    \cfbox{gray}{\includegraphics[height=1.8in]{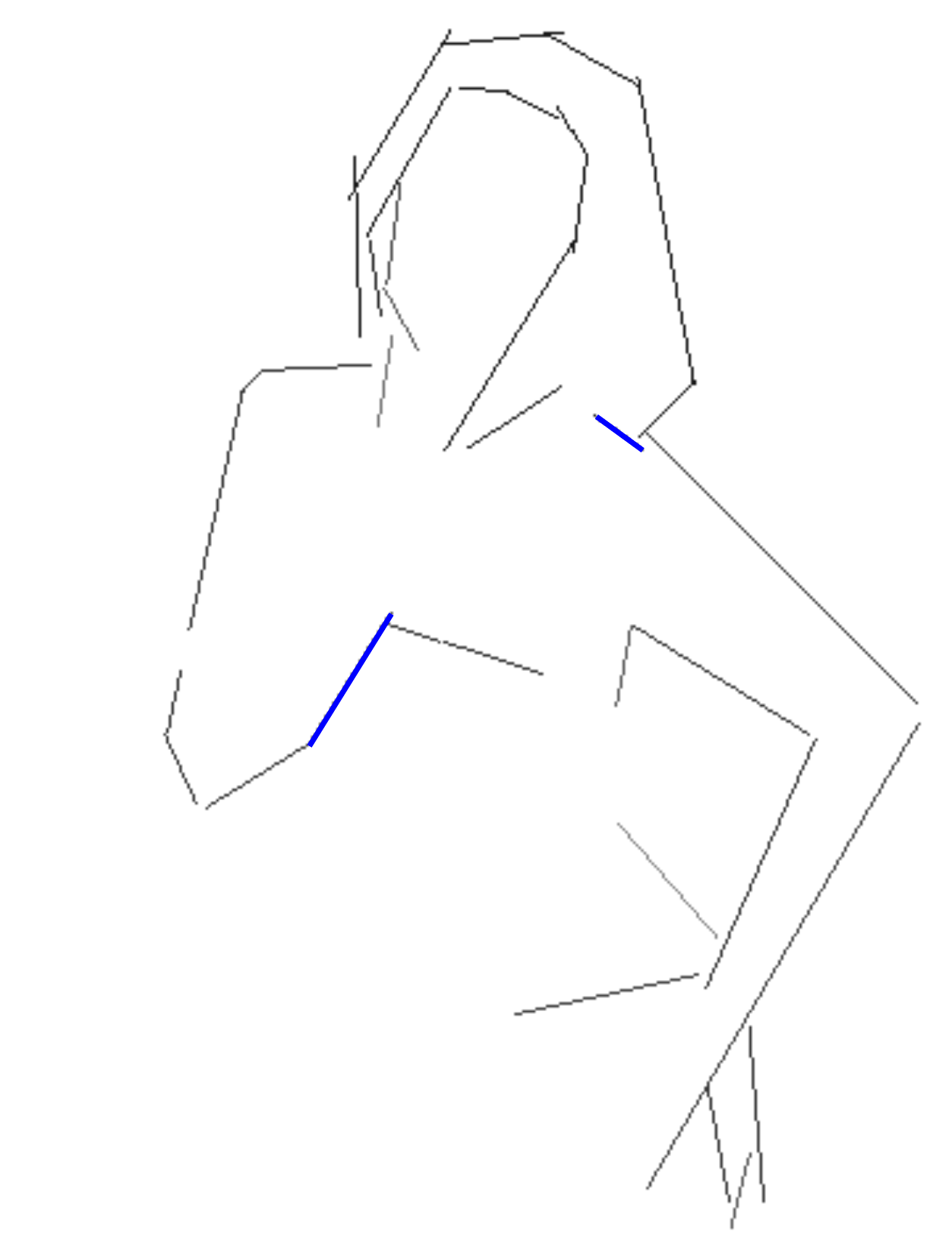}}} \hspace{-1mm}
  \subfigure[]{
    \label{fig:nbcal}
    \cfbox{gray}{\includegraphics[height=1.6in]{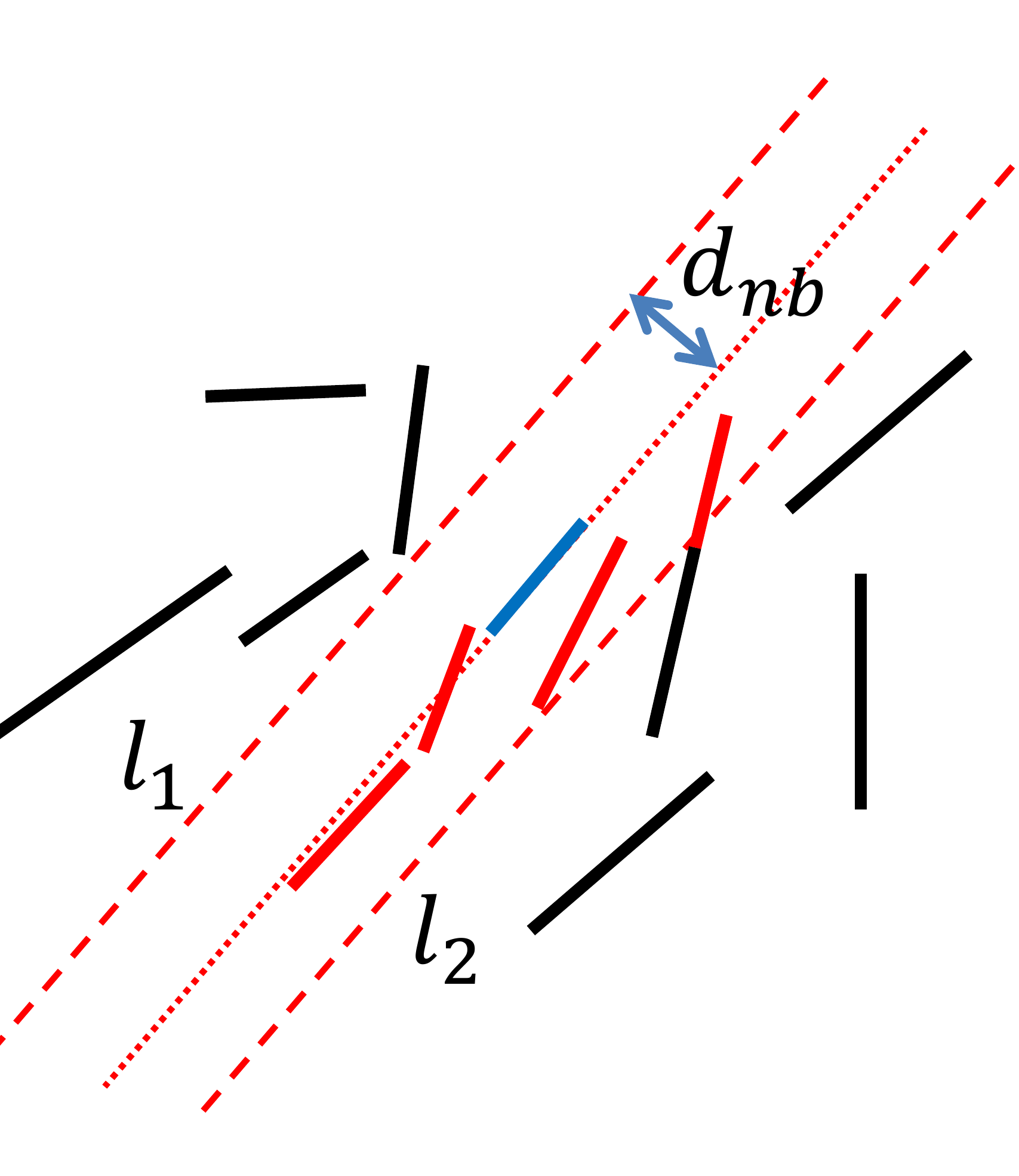}}} \hspace{-1mm}
  \subfigure[]{
    \label{fig:fittedpix}
    \cfbox{gray}{\includegraphics[height=1.6in]{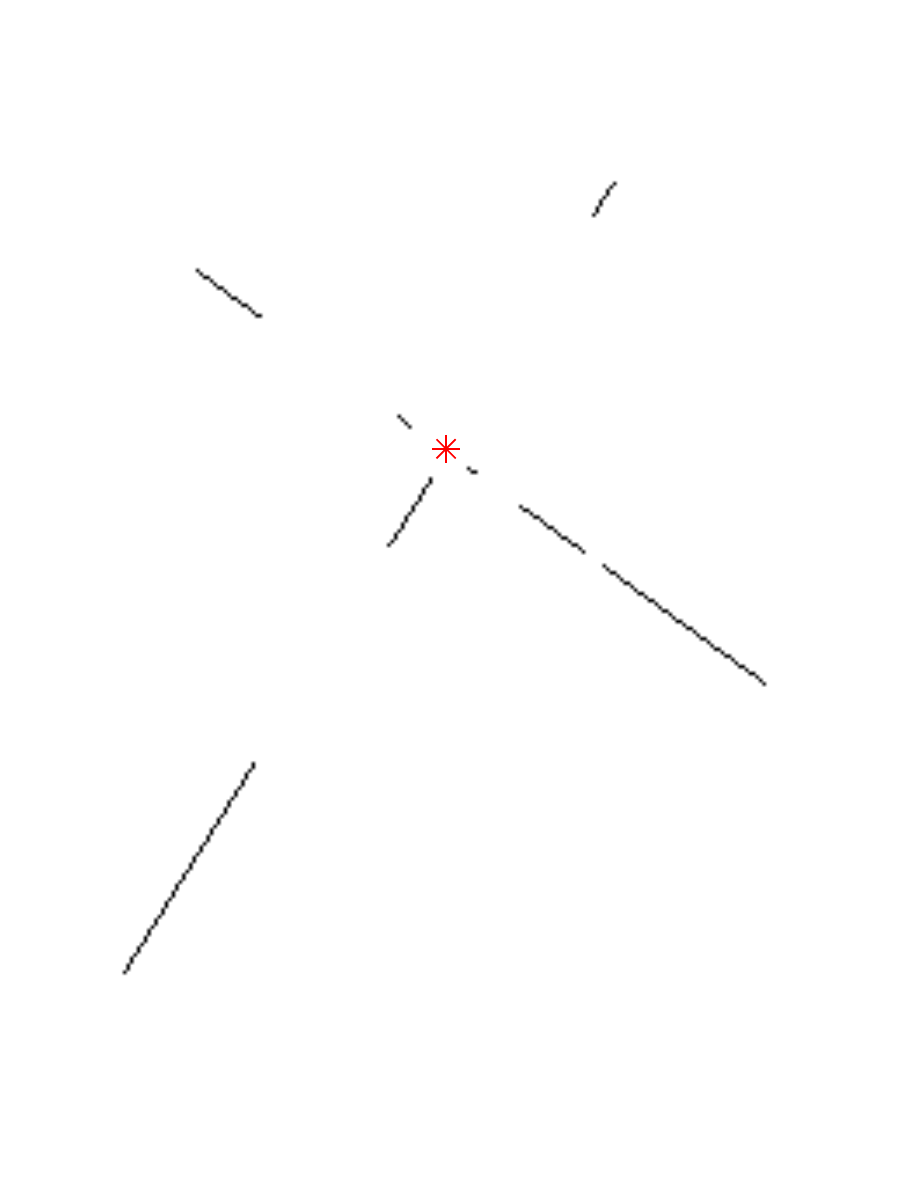}}}
  \caption{Illustration of the RANSAC algorithm. (a) original image, (b) candidate sides, (c) neighborhood region, and (d) fitted pixels.} 
  \label{fig:ransac} 
\end{figure}

\subsubsection{Fitting All Segments on Sides}
In this step, we first mark all line segments within neighborhood
of the two straight lines formed in the previous step as inliers and those falling outside 
neighborhood as outliers. The neighborhood region of a\break straight
line $l: ax+by+c=0$ is
defined to be 
\begin{equation*}
N(l)=\{(x, y) \in \mathbb{R}^2\mbox{ and } \frac{\left |
    ax+by+c \right |}{\sqrt{a^2+b^2}}\leqslant d_{nb}\}\;,
\end{equation*}
{\it i.e.}, the group of pixels whose distances to the
straight line are smaller than a certain threshold $d_{nb}$. In this paper, we fix $d_{nb}=5$ pixels. Then, all
the inlier line segments with respect to line $l$ can be calculated as
$I(l) = S \cap N(l)$ where $S$ is the set of all candidate line segments. Note that, if a line
segment is cut into two parts by the neighborhood boundary, the part of
line segment falling within the neighborhood is included as an
inlier, whereas the other part is considered as an outlier. In Figure~\ref{fig:nbcal}, we illustrate how the neighborhood of a
straight line is utilized to partition all the line segments into inliers and outliers. The blue line segment is selected from the candidate line segment set and extended to a straight line $l$. Lines $l_1$ and $l_2$ designate
boundaries of the neighborhood region. Both of them have a distance
$d_{nb}$ from $l$. The red line segments and black line segments represent the inliers and outliers, respectively.

Next, all the pixels on the inlier line
segments are projected onto the straight line. A pixel
on the straight line is called a \emph{projected pixel} if there is at
least one pixel on any inlier line segment that is projected to this
pixel. We denote the set of all projected pixels as
$$P(l)=\{(x', y')\in l \mid \exists (x, y)\in I(l)\mbox{, }(x-x',
y-y')\perp l\}.$$

Figure~\ref{fig:ransac} illustrates the fitting
process. Figure~\ref{fig:candside} shows all
the line segments extracted from the image shown in Figure~\ref{fig:rorig}. The two line segments
colored in blue are randomly selected from the candidate set. Subsequently, two straight lines, $l$ and $\tilde{l}$, are
generated by expanding the two line segments and their neighborhood
regions are identified. Finally, inliers within their neighborhoods
are projected onto the straight lines, as shown in
Figure~\ref{fig:fittedpix}. 

Here, we note that the projected pixels typically scatter along the entire
straight line. To form a triangle, we divide the line into two \emph{half lines} at the intersection point. Formally, a half line here is defined as a straight line extending from the intersection point indefinitely in one direction only. It is easy to see that, for each pair of straight lines $l$ and $\tilde{l}$, four triangles can be formed using different pairs of half lines.
Therefore, when evaluating the fit of a triangle, we only consider the
subsets of projected pixels which are on the two half lines that form the triangle, denoted
as $P(l^h)$ and $P(\tilde{l}^h)$, as opposed to the
entire sets of projected pixels $P(l)$ and $P(\tilde{l})$.

\subsubsection{Evaluating the Fitted Triangle}
In order to evaluate the quality of a fitted triangle, we first introduce a
\emph{Continuity Ratio} score to evaluate the quality of a fitted side. Here we use one half line $l^h$ as an example. The way to calculate continuity ratio for the other half line $\tilde{l}^h$ is exactly the same. Given any point $X$ lying on the half line $l^h$, we can construct a potential side $OX$ connecting the intersection point $O = (x_o, y_o)$ and $X$. We further compute the number of pixels projected onto $OX$ divided by the length of $OX$ as $\frac{P(l^h)\cap OX}{\left | OX \right |}$. Then, the \emph{Continuity Ratio} of $l^h$ is defined as the ratio of the best fitted side on $l^h$:
\begin{equation}
	C(l^h) = \max_{X\in  P(l^h)} \frac{P(l^h)\cap OX}{\left | OX \right |}\;.
\end{equation}
Finally, the continuity ratio for the entire triangle constructed by the two half lines $l^h$ and $\tilde{l}^h$ is 
\begin{equation}
C(l^h, \tilde{l}^h) = C(l^h) \times C(\tilde{l}^h)\;.
\end{equation}

In addition to the continuity ratio, we define another \emph{Total Ratio} score which is calculated as the area of the triangle divided by the area of the entire picture. Intuitively, the \emph{Continuity Ratio}  describes how well the extracted line segments fit the given side. Meanwhile, the \emph{Total Ratio} represents the significance
of a fitted triangle in terms of sizes. As a bigger triangle can be more easily recognized and has more impact on composition of the entire image, we only keep the triangles whose \emph{Continuity Ratio} and \emph{Total Ratio} scores are both above certain thresholds.

\section{Experiments}\label{sec:exp}
\subsection{Image Segmentation for Natural/Urban Scenes}

In this section, we compare the performance of our method with the state-of-the-art image segmentation method, $gPb$-owt-ucm~\citep{ArbelaezMFM11}. For this experiment, we assume known dominant vanishing point locations. We emphasize that our goal here is not to compete with that work as a generic image segmentation algorithm, but to demonstrate that information about the vanishing point (\ie, the geometric cue), if properly harnessed, can empower us to get better segmentation results.

To quantitatively evaluate the methods, we use three popular metrics to compare the result obtained by each algorithm with the manually-labeled segmentation:\break Rand index (RI), variation of information (VOI) and segmentation covering (SC). First, the RI metric measures the probability that an arbitrary pair of pixels have the same label in both partitions or have different labels in both partitions. The range of RI metric is $[0,1]$, higher values indicating greater similarity between two partitions. Second, the VOI metric measures the average condition entropy of two clustering results, which essentially measures the extent to which one clustering can explain the other. The VOI metric is non-negative, with lower values indicating greater similarity. Finally, the SC metric measures the overlap between the region pairs in two partitions. The range of SC metric is $[0,1]$, higher values indicating greater similarity. We refer interested readers to~\citep{ArbelaezMFM11} for more details about these metrics.

\begin{figure}[ht!]
\centering
\includegraphics[height =0.65in]{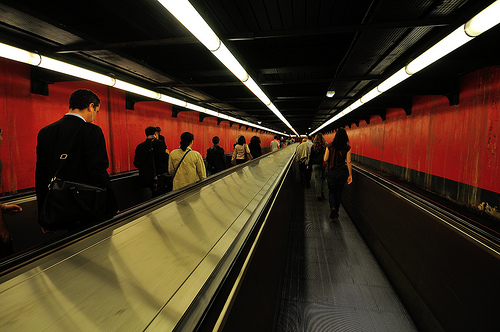} 
\includegraphics[height =0.65in]{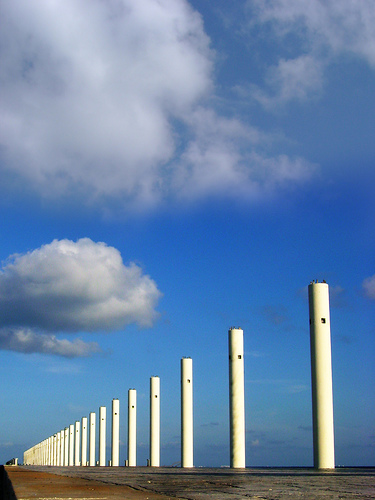} 
\includegraphics[height =0.65in]{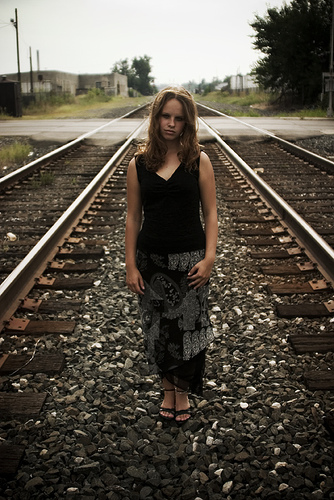} 
\includegraphics[height =0.65in]{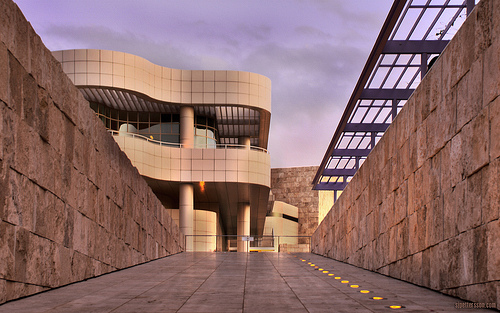} 
\\
\includegraphics[height =0.65in]{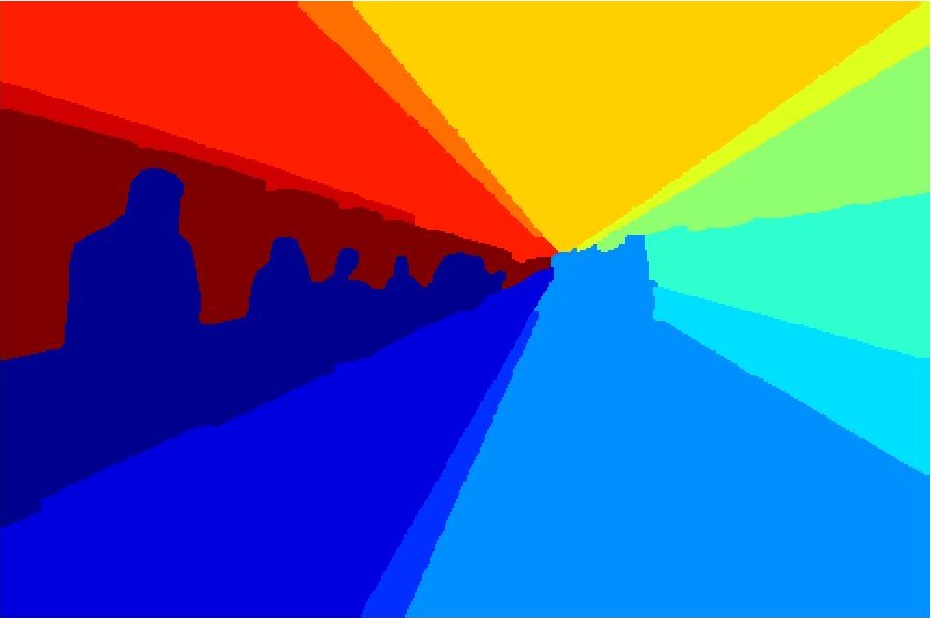} 
\includegraphics[height =0.65in]{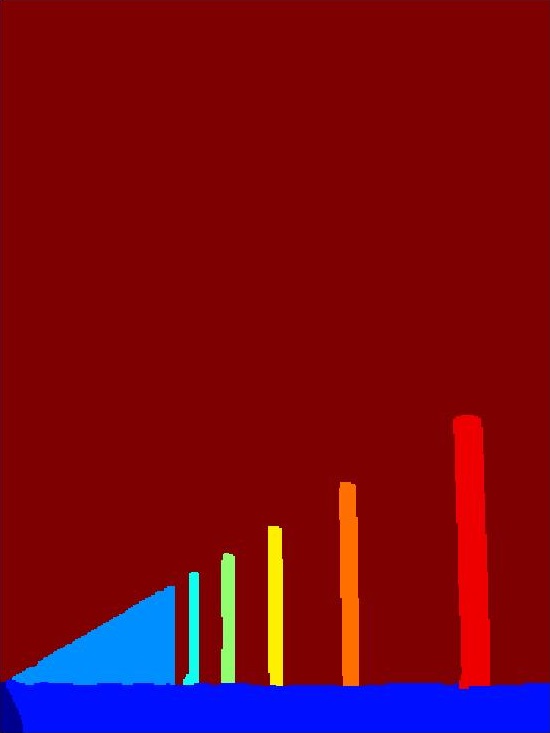} 
\includegraphics[height =0.65in]{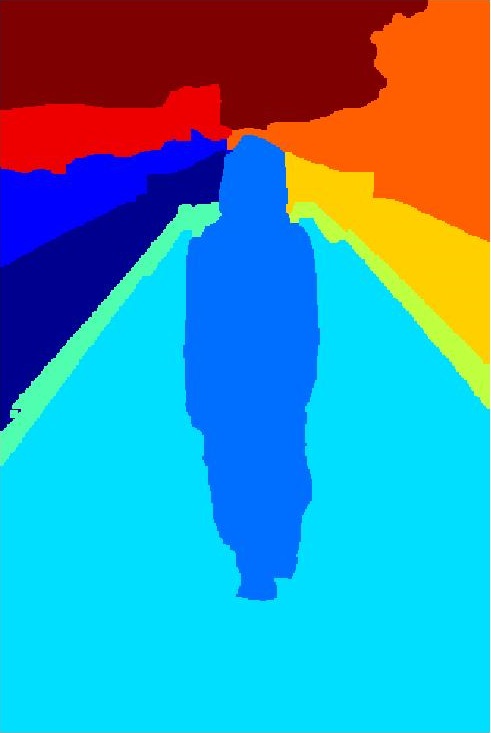} 
\includegraphics[height =0.65in]{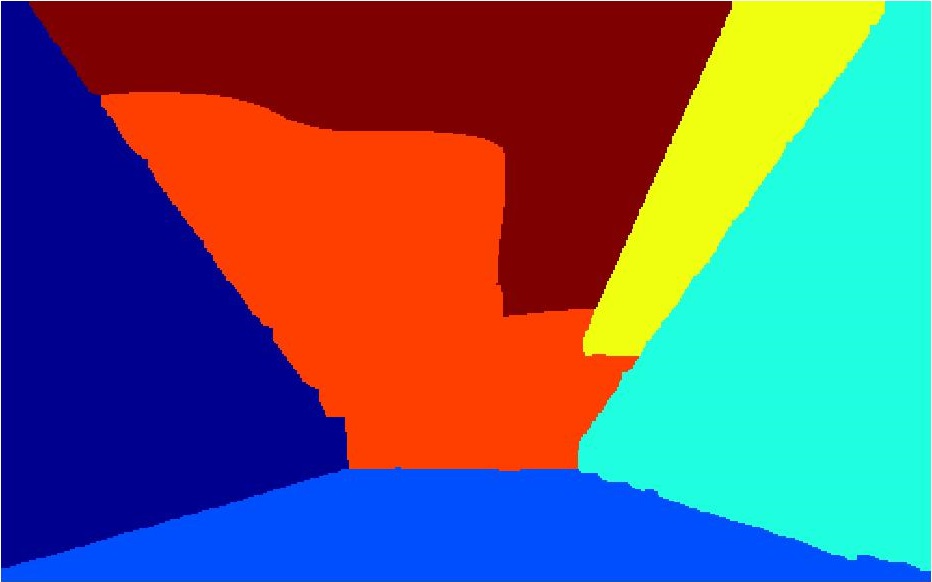} 
\caption{Example test images with the manually labeled segmentation maps.}
\vspace{-5mm}
\label{fig:curve1}
\end{figure}

For this experiment, we manually labeled 200 images downloaded from
{\tt flickr.com}. These images cover a variety of indoor and outdoor
scenes and each has a dominant vanishing point. During the labeling
process, our focus is on identifying all the regions that differ from
their neighbors either in their geometric structures or photometric
properties. We show some images with the hand-labeled segmentation
maps in Figure~\ref{fig:curve1}. Note that, ideally this process should be done by someone unrelated to
the work and even by multiple people. But since this particular
segmentation task is relatively well-defined, the level of subjectivity or
inter-rater variation is not expected to be high. Also, the process is labor
intensive as it traces region boundaries. Through doing the task
ourselves, we ensure quality of the segmentation maps.  The
manually-labelled segmentation maps will be made available so researchers can
examine for correctness and experiment with them.

\begin{figure*}[t!]
\centering
\subfigure[Rand Index]{\includegraphics[height =1.2in]{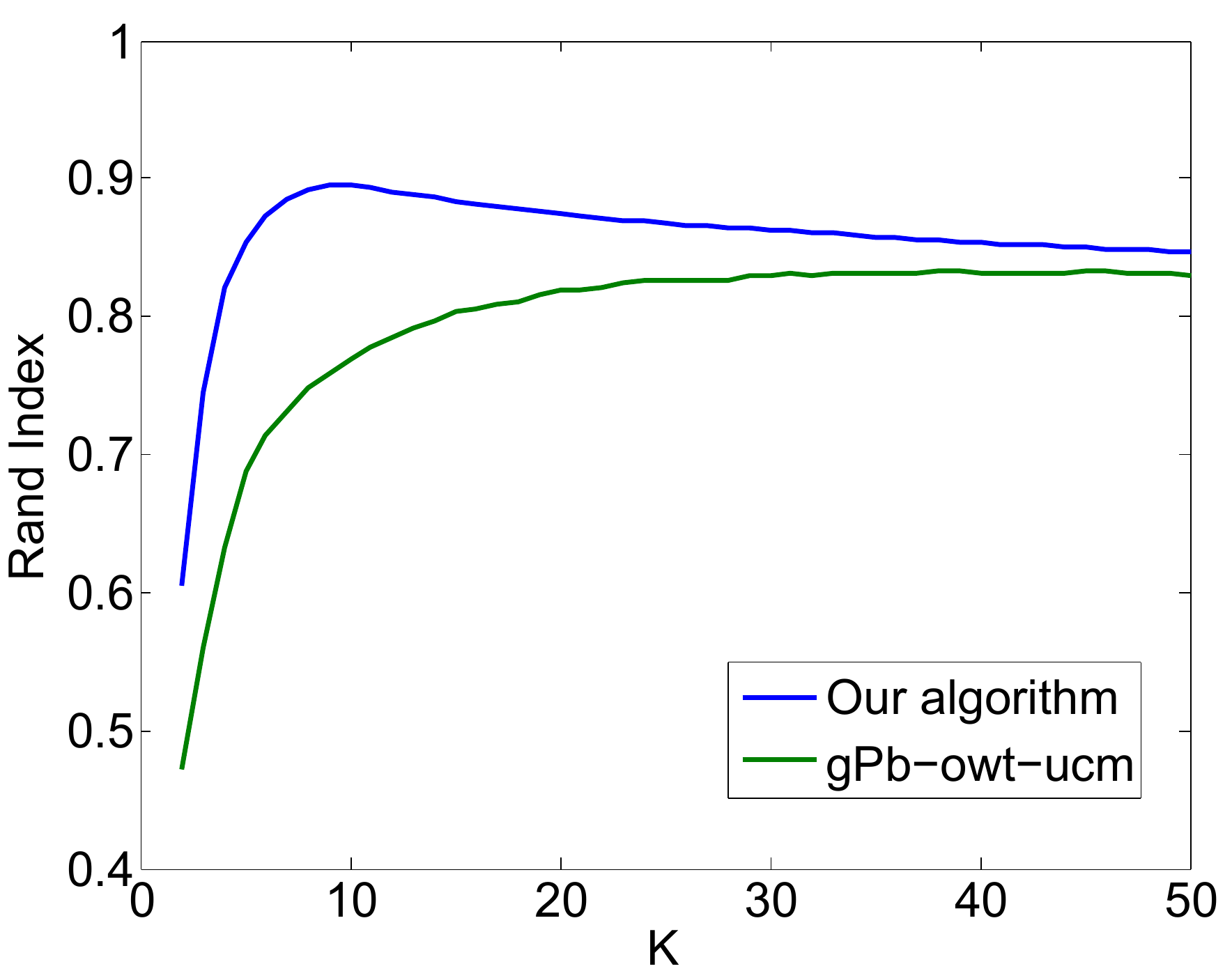}} \hspace{3mm}
\subfigure[Segmentation Covering]{\includegraphics[height =1.2in]{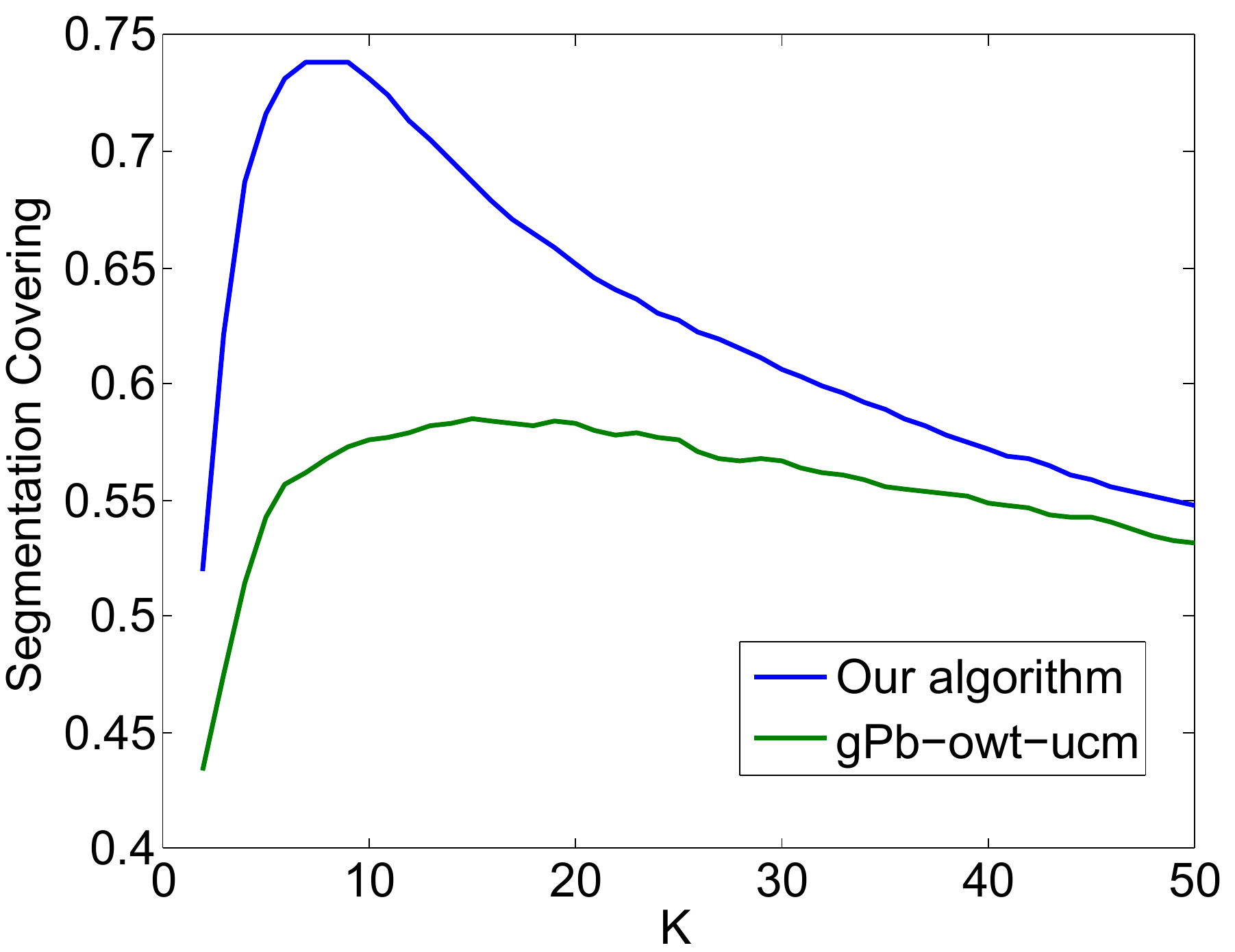}} \hspace{3mm}
\subfigure[Variation of Information]{\includegraphics[height =1.2in]{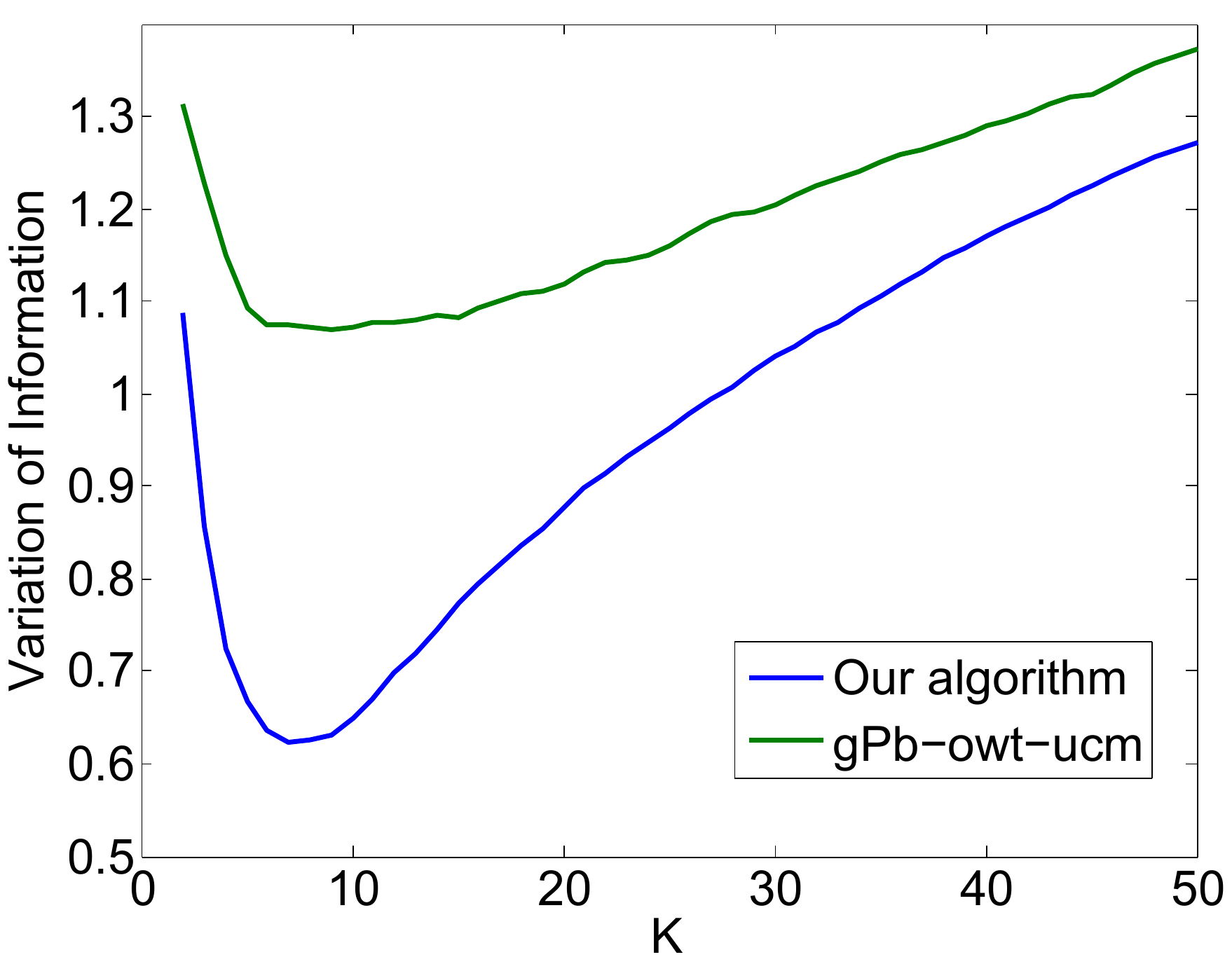}}
\caption{Segmentation benchmarks. $K$ is the number of regions.}
\label{fig:curve2}
\end{figure*}

Figure~\ref{fig:curve2} shows the benchmark results of both methods. Our method significantly outperforms $gPb$-owt-ucm on all metrics. This is consistent with the example results in Figures~\ref{fig:seg-lambda} and~\ref{fig:seg-results-1}, suggesting that our method is advantageous in segmenting the geometric structures in the scene.

\subsection{Vanishing Point Detection}
Next, we compare our vanishing point detection method with two state-of-the-art methods proposed by\break \citet{Tardif09} and~\citet{TretiakBKL12}, respectively. As discussed earlier, both methods rely on the line segments to generate vanishing point candidates. Then, a non-iterative scheme similar to the popular RANSAC technique is developed by~\citet{Tardif09} to group the line segments into several clusters, each corresponding to one vanishing point. Using the vanishing points detected by~\citet{Tardif09} as an initialization,~\citet{TretiakBKL12} further propose a non-linear optimization framework to jointly refine the extracted line segments and vanishing points.

\begin{figure}[t!]
\centering
\includegraphics[height =1.6in]{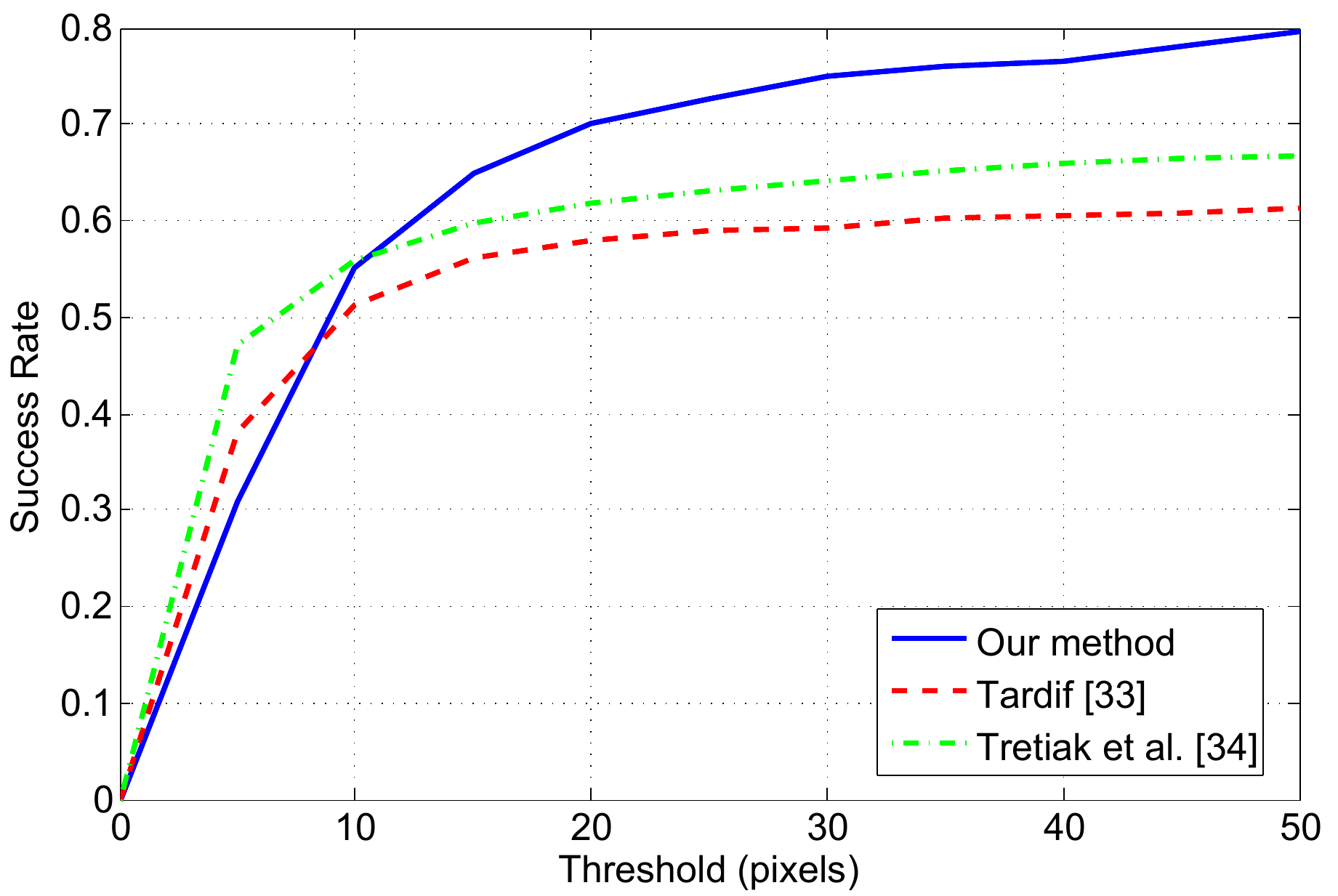}
\caption{Comparison of vanishing point detection algorithms.}
\label{fig:vp-rate}
\end{figure}

In this experiment, we use 400 images downloaded from {\tt flickr.com} whose dominant vanishing points lie within the image frame. All images are scaled to size $500\times 330$ or $330\times 500$. To make the comparison fair, for~\citet{Tardif09} and~\citet{TretiakBKL12} we only keep the vanishing point with the largest support set among all hypotheses that also lie within the image frame. We consider a detection successful if the distance between the detected vanishing point and the manually labeled ground truth is smaller than certain threshold $t$, and plot the success rates of all methods w.r.t. the threshold $t$ in Figure~\ref{fig:vp-rate}. Our method outperforms existing methods as long as the threshold is not too small ($t\geq 10$ pixels), justifying its effectiveness for detecting the dominant vanishing point in arbitrary images. When $t$ is small, our method does not perform well because its precision in locating the vanishing point is limited by the size of the grid mesh. Nevertheless, this issue can be alleviated using a denser grid mesh at the cost of more computational time. Also, we note that while the joint optimization scheme proposed by~\citet{TretiakBKL12} can recover weak line segments and vanishing points for urban scenes, its improvement over~\citet{Tardif09} is quite small in our case.

\subsection{Triangle Detection in Portrait Images}
\label{sec:exp-portrait}

In order to evaluate the performance of our triangle detection method for portrait images, we build a dataset by
collecting 4,451 professional studio portrait photos from Flickr. 

\begin{figure}[ht!]
	\centering 
	\subfigure[0.88]{
		\includegraphics[width=0.11\textwidth]{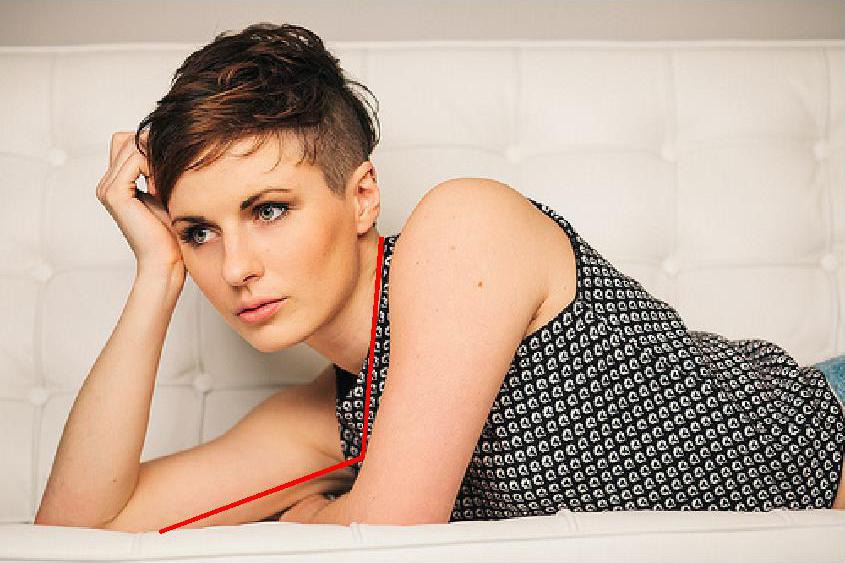}} 
	\subfigure[0.71]{
		\includegraphics[width=0.11\textwidth]{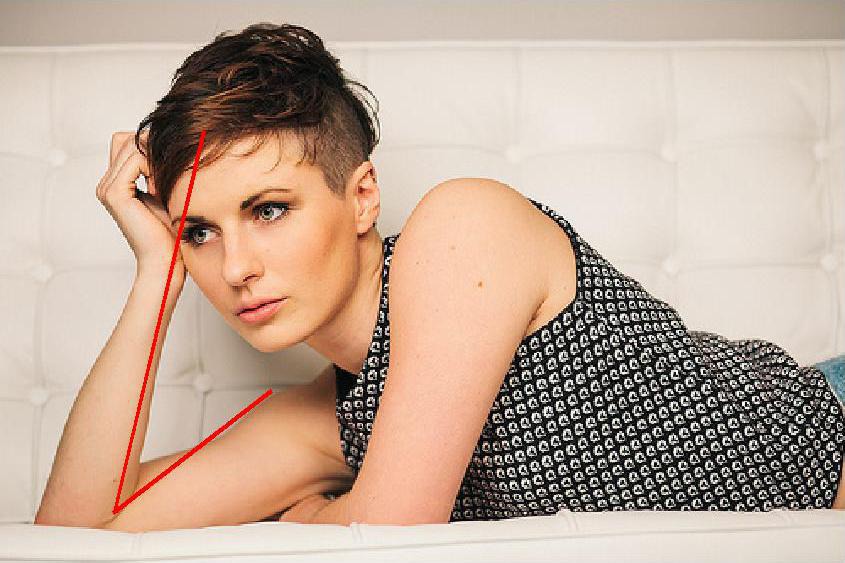}} 
	\subfigure[0.73]{
		\includegraphics[width=0.11\textwidth]{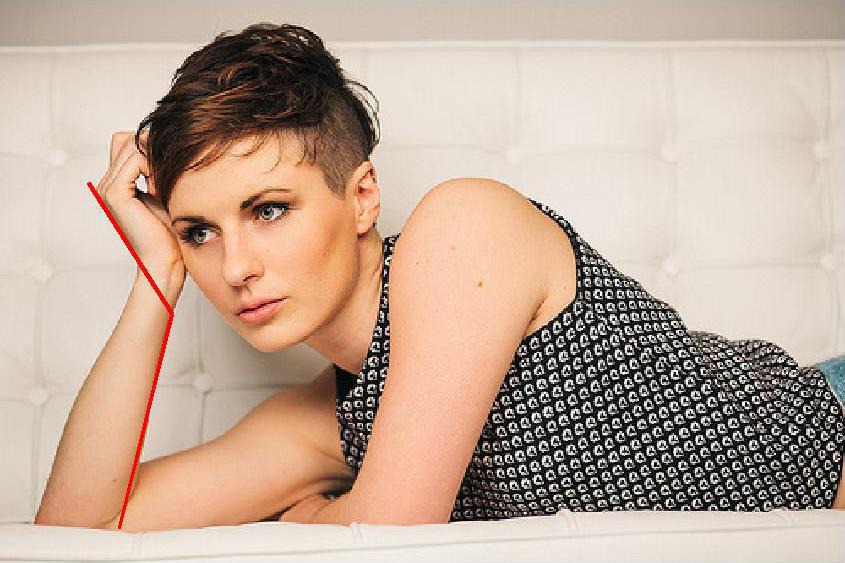}} 
	\subfigure[0.73]{
		\includegraphics[width=0.11\textwidth]{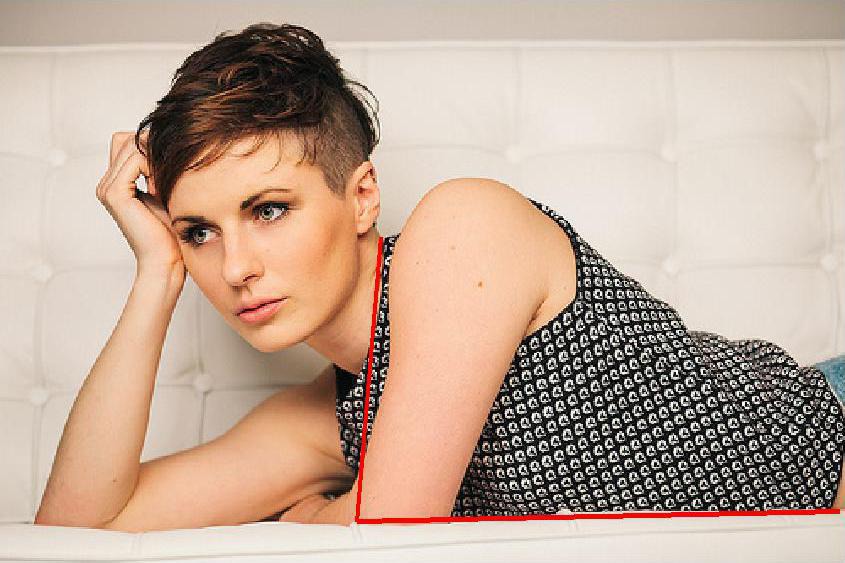}}\vspace{-2mm}

	\subfigure[ 0.59]{
		\includegraphics[width=0.11\textwidth]{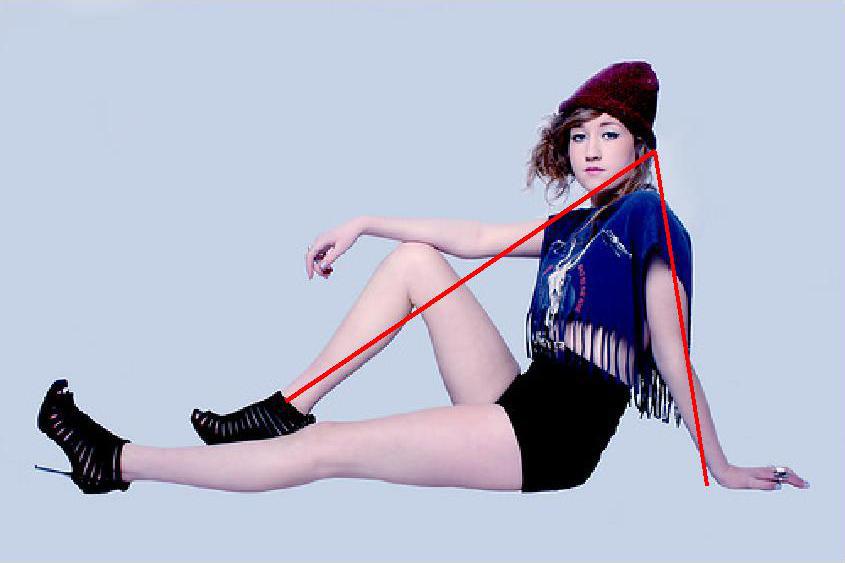}}
	\subfigure[ 0.61]{
		\includegraphics[width=0.11\textwidth]{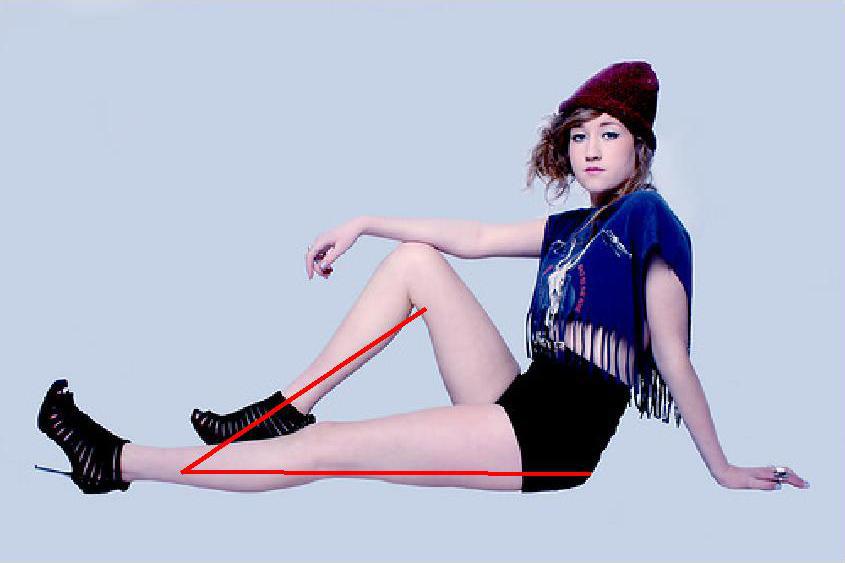}}
	\subfigure[ 0.68]{
		\includegraphics[width=0.11\textwidth]{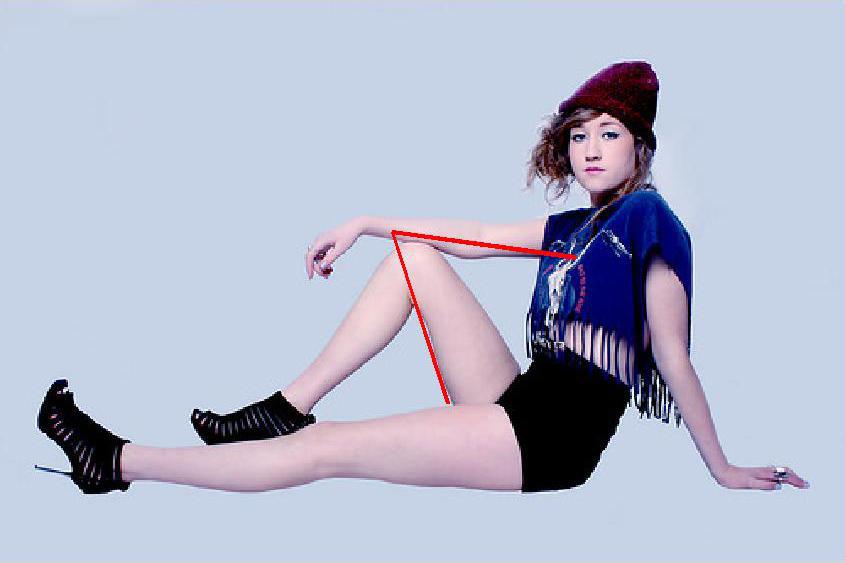}}
	\subfigure[ 0.78]{
		\includegraphics[width=0.11\textwidth]{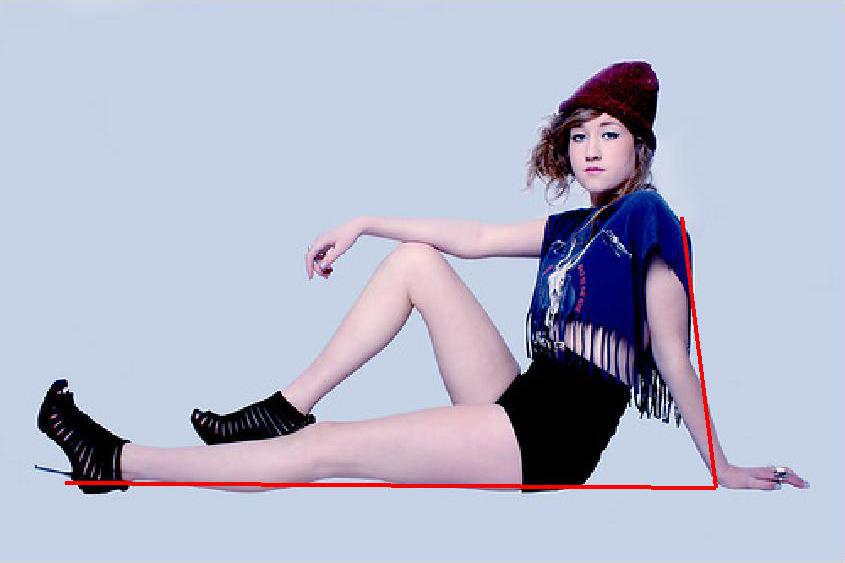}}\vspace{-2mm}	
	
	\subfigure[ 0.49]{
		\includegraphics[width=0.075\textwidth]{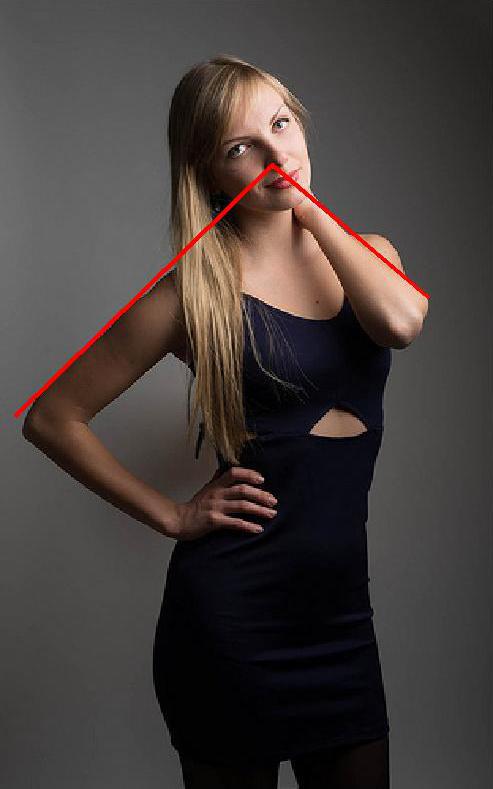}}
	\subfigure[ 0.50]{
		\label{fig:useful1}
		\includegraphics[width=0.075\textwidth]{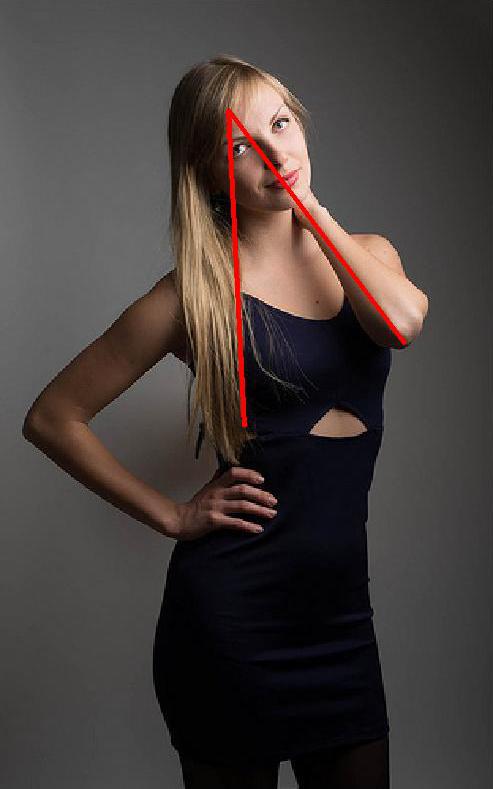}}
	\subfigure[ 0.53]{
		\includegraphics[width=0.075\textwidth]{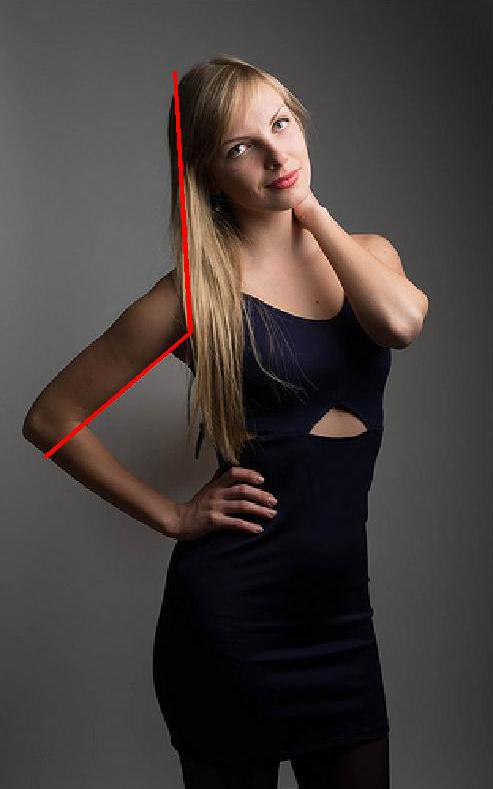}}
	\subfigure[ 0.71]{
		\includegraphics[width=0.075\textwidth]{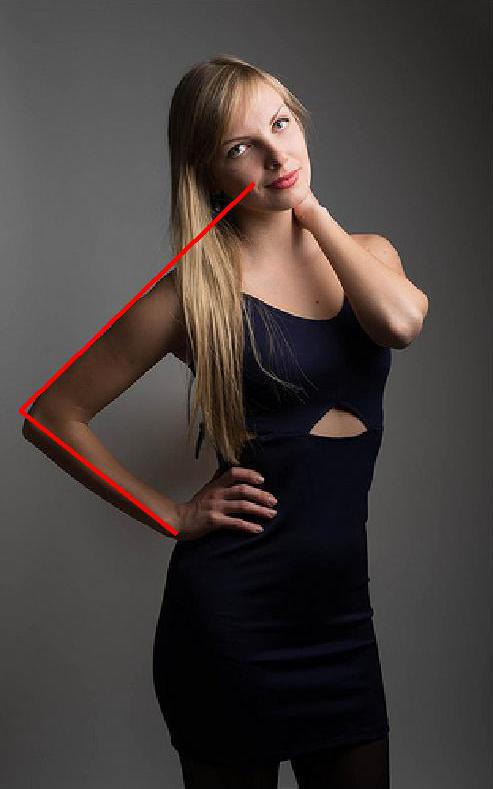}}
	\subfigure[ 0.94]{
		\label{fig:highcr1}
		\includegraphics[width=0.075\textwidth]{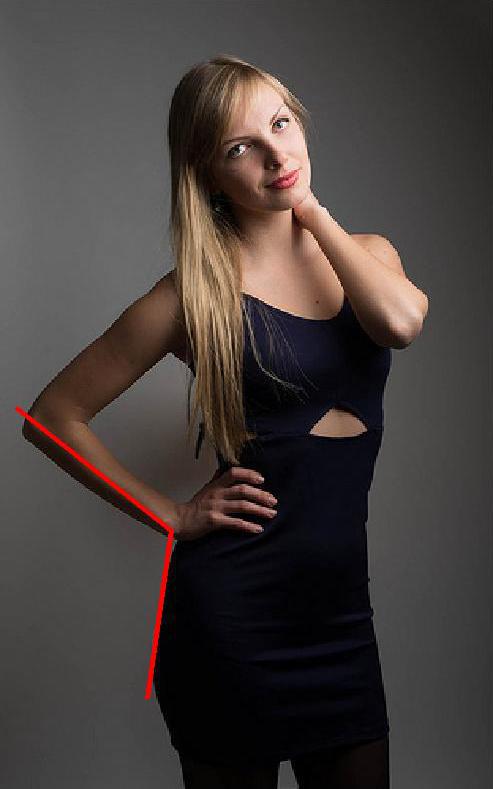}}  \vspace{-2mm}
		
	\subfigure[ 0.51]{
		\includegraphics[width=0.075\textwidth]{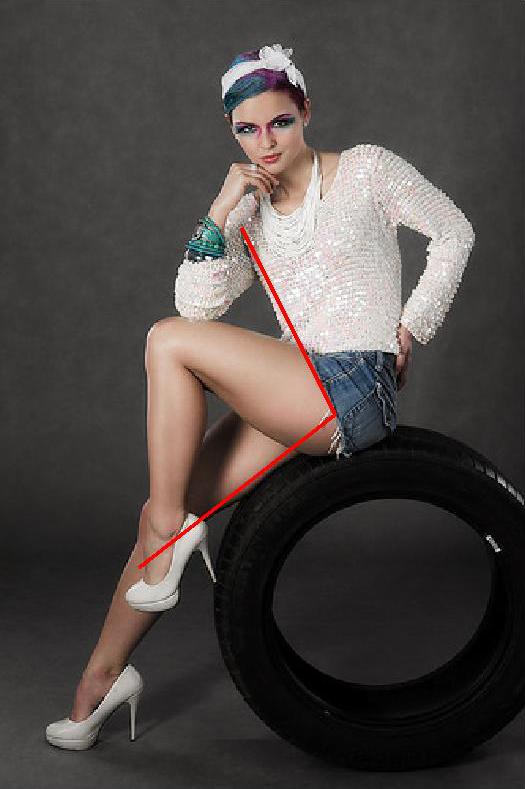}}
	\subfigure[ 0.52]{
		\includegraphics[width=0.075\textwidth]{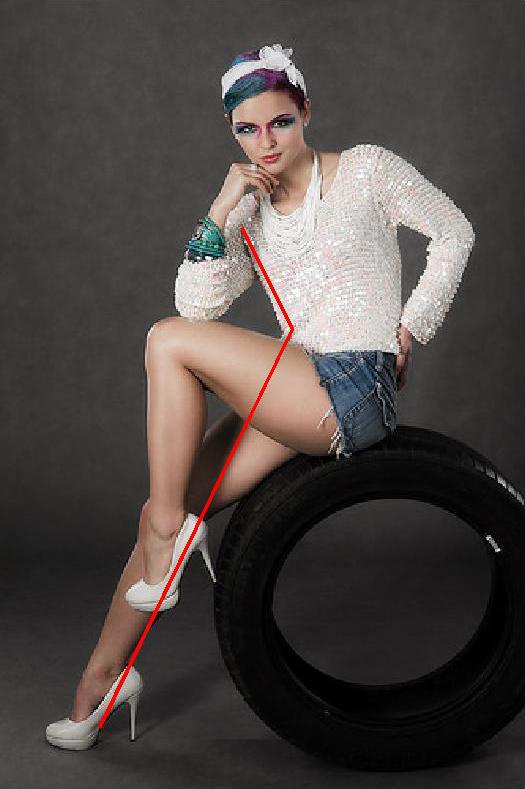}}
	\subfigure[ 0.54]{
		\includegraphics[width=0.075\textwidth]{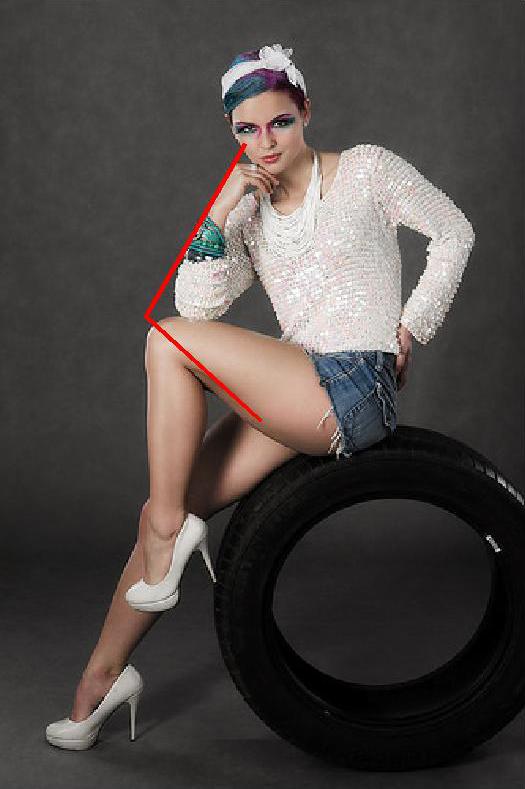}}
	\subfigure[ 0.73]{
		\includegraphics[width=0.075\textwidth]{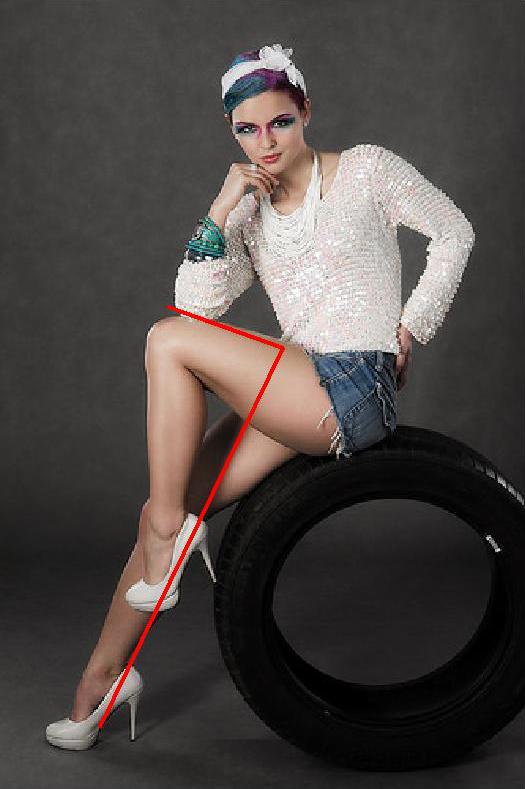}}
	\subfigure[ 0.79]{
		\includegraphics[width=0.075\textwidth]{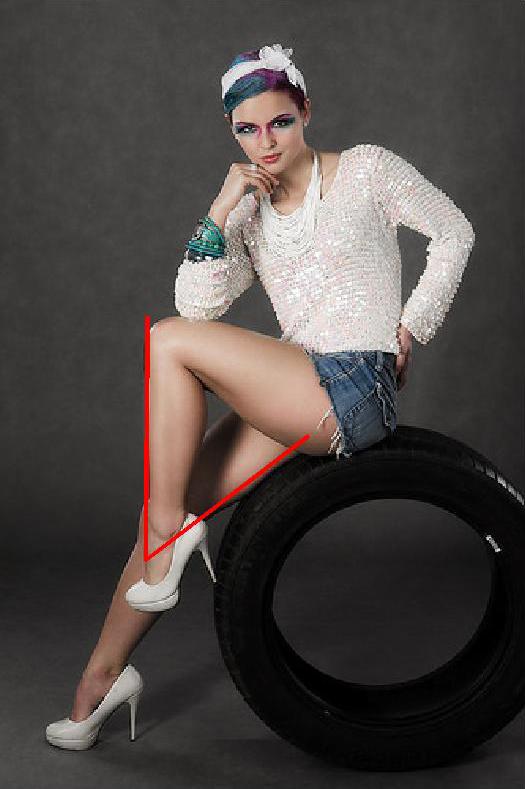}}\vspace{-2mm}	
		
	\subfigure[ 0.63]{
		\includegraphics[width=0.075\textwidth]{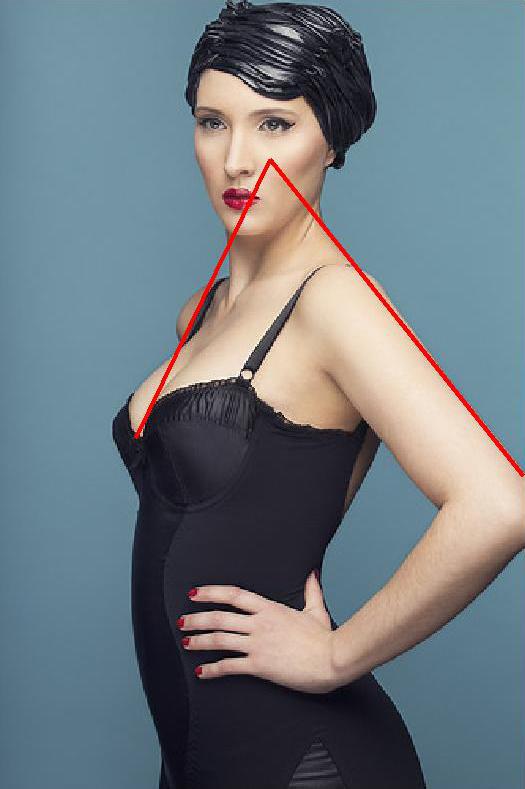}}
	\subfigure[ 0.73]{
		\includegraphics[width=0.075\textwidth]{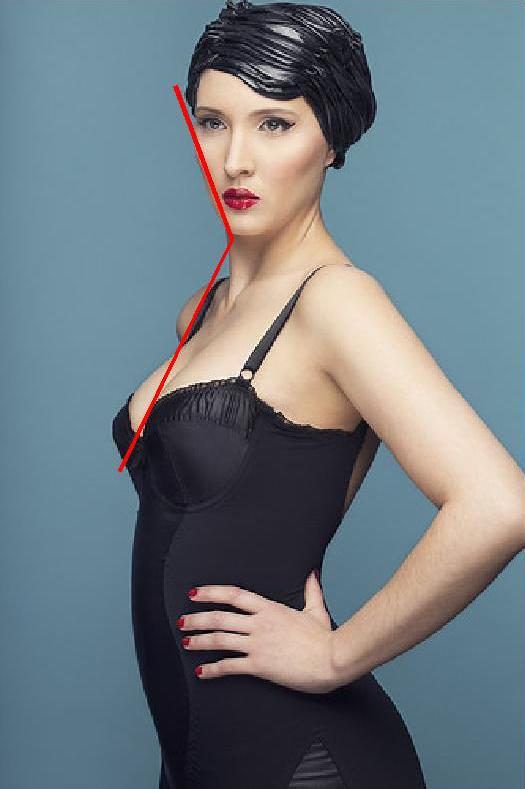}}
	\subfigure[ 0.60]{
		\includegraphics[width=0.075\textwidth]{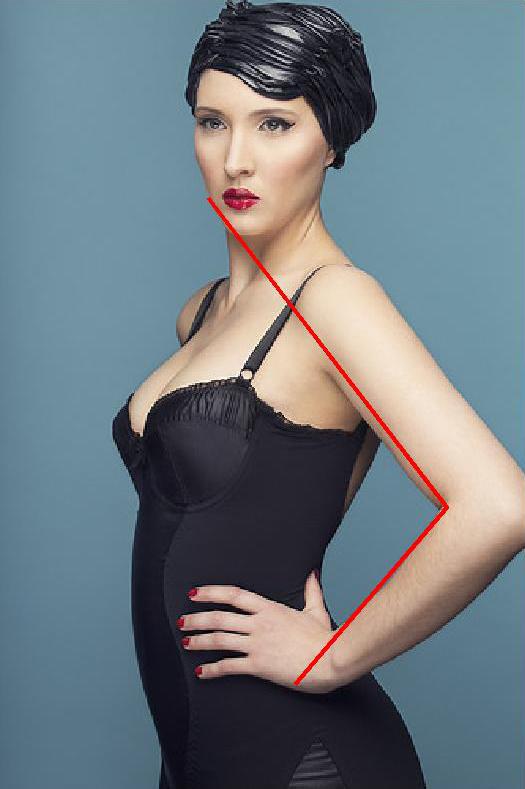}}
	\subfigure[ 0.68]{
		\includegraphics[width=0.075\textwidth]{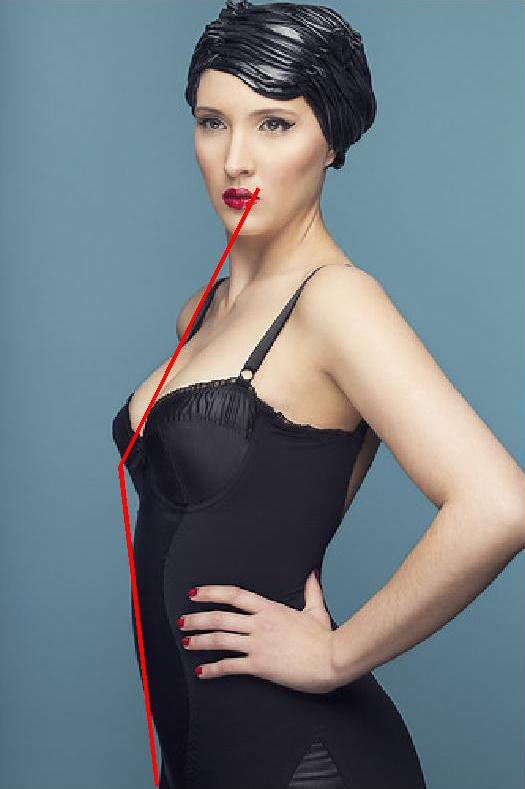}}
	\subfigure[ 0.71]{
		\includegraphics[width=0.075\textwidth]{fig-portrait/differentCR/2626/3.jpg}}
	
	\caption{Detected triangles with different \emph{Continuity Ratios}. The red lines indicate two sides of the detected triangle.}
	\label{fig:results_scs} 
\end{figure}

Figure~\ref{fig:results_scs} presents some detected triangles with different continuity ratios. As one can see, triangles with high continuity ratios are more easily recognized than those with low continuity ratios. However, they do not necessarily outperform those of low continuity ratios in conveying useful compositional information about the photo. For instance, Figure~\ref{fig:useful1} has a much lower continuity ratio than Figure~\ref{fig:highcr1} but it conveys a more interesting compositional skill. From the detected triangle, we can notice that the model slightly tilts her head to align with her left arm which constructs a beautiful triangle with her hair. It is very common that multiple different triangles are embedded in one image, and some of them can be easily overlooked by amateurs. Here, our goal is to identify all potential triangles from images, which enables amateur photographers to better learn compositional techniques.

\begin{figure}[t!]
  \centering 
    \includegraphics[width=0.48\textwidth]{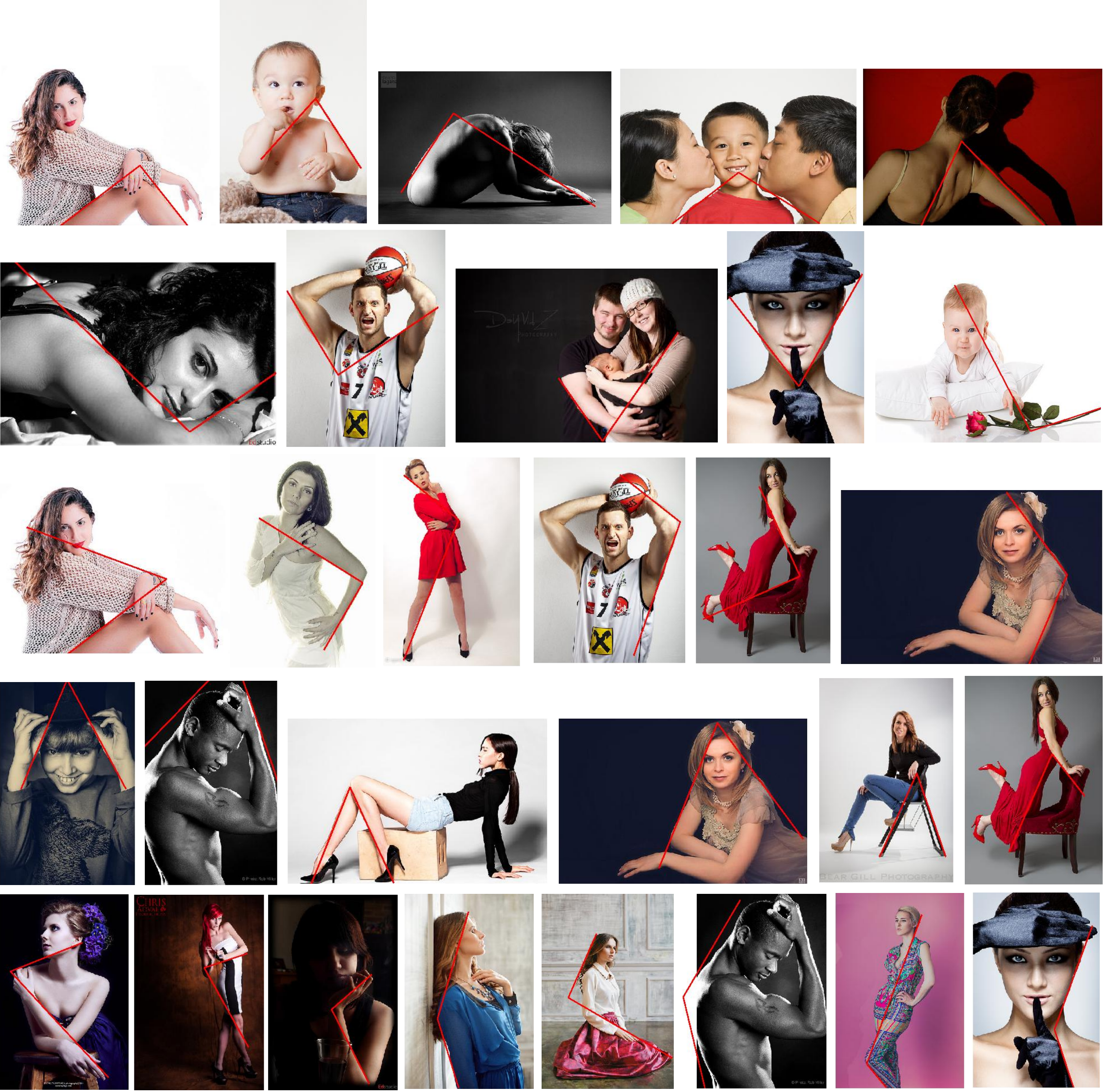}
  \caption{Examples of detected triangles in portrait photographs.} 
  \label{fig:results} 
\end{figure}

More detected results can be found in Figure~\ref{fig:results}. It demonstrates that our triangle detection system can clearly
identify triangles in portrait photographs despite the
existence of noises such as human hair, props, and shadows/patterns/folds on
shirts, etc. Moreover, triangles involving multiple
human subjects can be accurately detected as well ({\it e.g.}, the fourth picture
in the first row and the third picture in the second row). More interestingly, our system is able to locate
triangles that may not be easy to identify by human eyes. For
instance, considering the first photo in the last row, it is quite
easy to find the triangle containing the girl's two
arms. However, we often overlook another triangle which is constructed
with one arm of the girl and the edge of her lower jaw. Another example is the fifth picture in the
last row. The girl puts her arm on her dress in a deliberately
designed pose so that it extends the boundary of her dress and forms a big triangle together with her long hair. Both examples
indicate that professional photographers usually design delicate poses for
the subjects in order to achieve quality photo composition. However, choosing an interesting 
pose requires much experience and artistic inspiration. The
triangles detected by our system can help amateurs gain deeper
understanding and inspirations from high-quality photography works.

\begin{figure}[ht!]
\centering
\includegraphics[height =0.77in]{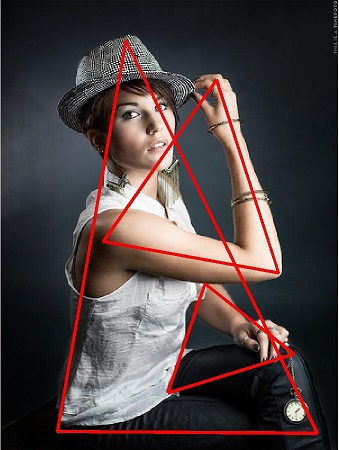} 
\includegraphics[height =0.77in]{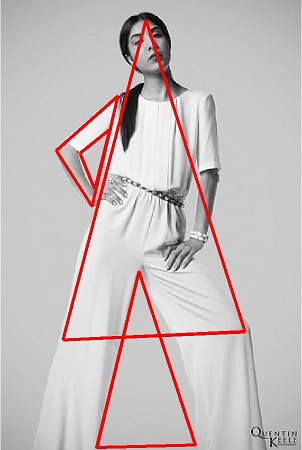} 
\includegraphics[height =0.77in]{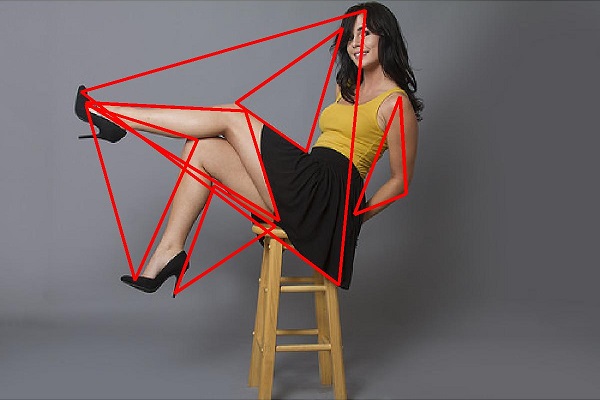} 
\includegraphics[height =0.77in]{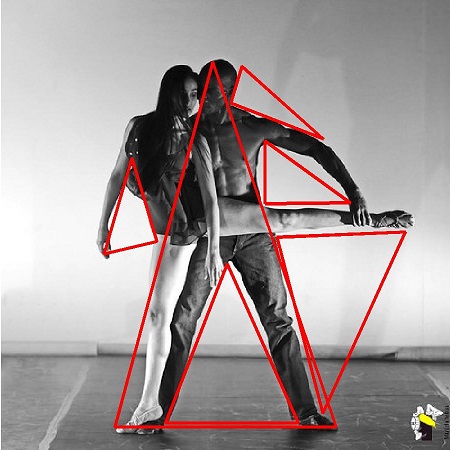}  
\\
\vskip 0.05in
\includegraphics[height =0.78in]{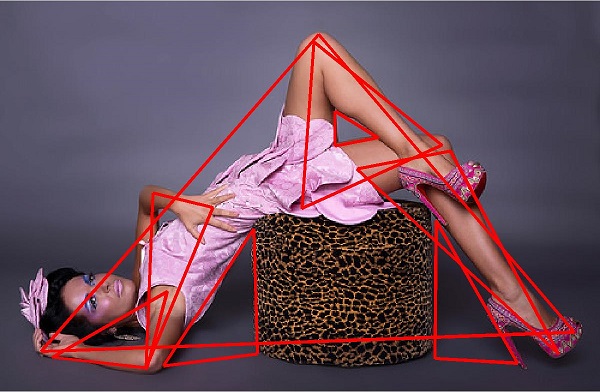} 
\includegraphics[height =0.78in]{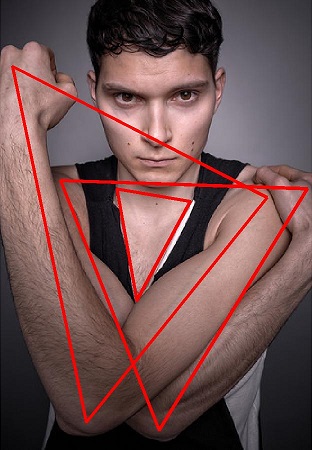} 
\includegraphics[height =0.78in]{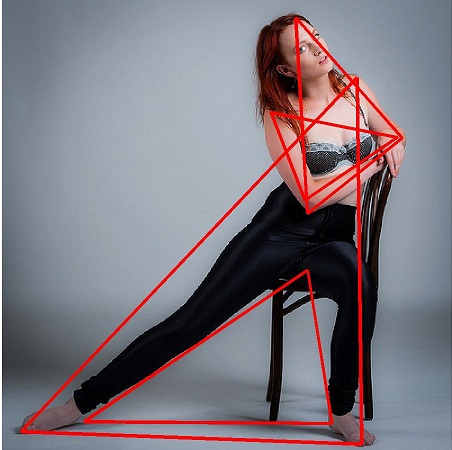} 
\includegraphics[height =0.78in]{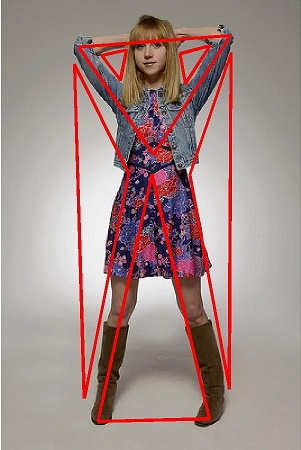} 
\\
\vskip 0.05in
\includegraphics[height =0.8in]{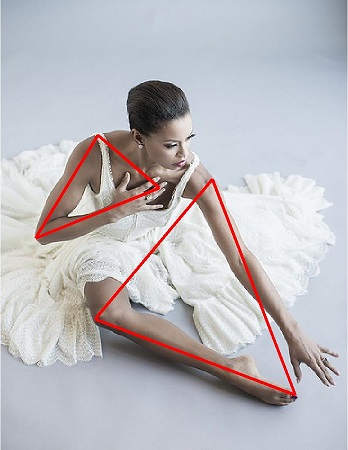} 
\includegraphics[height =0.8in]{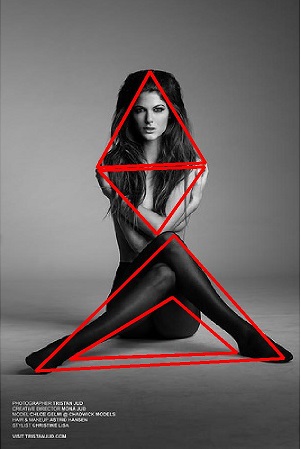} 
\includegraphics[height =0.8in]{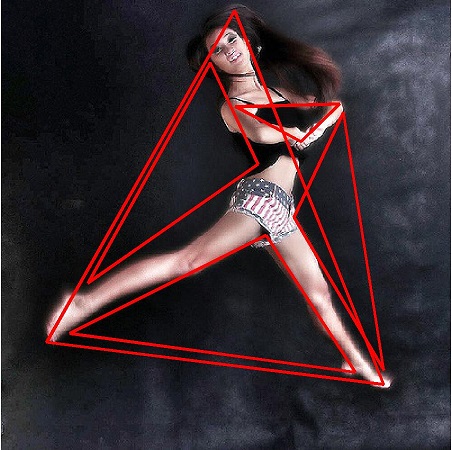} 
\includegraphics[height =0.8in]{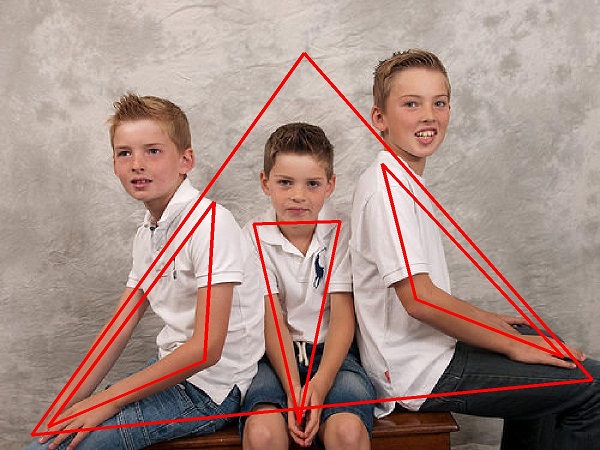} 
\\
\caption{Example portrait photos with triangles labeled by an experienced professional photographer.}
\label{fig:portrait-pro}
\end{figure}

\smallskip
\noindent{\bf Quantitative Evaluation.} To further study the effectiveness of our method, we involved an experienced professional portrait photographer, who has studied arts and architecture and has operated a professional portrait studio for over 30 years.
Using the online annotation tool LabelMe\footnote{http://labelme.csail.mit.edu/Release3.0/}, we asked the photographer to manually annotate triangles that indicate interesting pose or composition in the images we collected. A total of 173 images were annotated by the photographer. We show some example triangles in Figure~\ref{fig:portrait-pro}, which illustrate a wide variety of numbers, sizes and orientations of triangles used in the portrait photography. As the professional annotations can be valuable to computer vision community in studying portrait composition, we will make the dataset and any future extensions freely available to researchers.

We compare the triangles detected by our\break method with the manual annotations. In this experiment, we only consider triangles whose continuity ratio and total ratio are above $0.1$. Let $(A,B,C)$ denote the set of vertices of a ground truth triangle, we consider a candidate triangle $(A',B',C')$ a matching triangle if
\begin{equation}
\frac{|\overline{AA'}| + |\overline{BB'}| + |\overline{CC'}|}{|\overline{AB}| + |\overline{BC}| + |\overline{CA}|} \leq \delta\;,
\end{equation}
where $|\overline{AB}|$ is the length of the line segment connecting $A$ and $B$, and $\delta$ is a threshold. We fix $\delta=0.3$ in this paper.

\begin{figure}[t!]
\centering
\begin{tabular}{cc}
\hspace{-3mm}\includegraphics[height =1.2in]{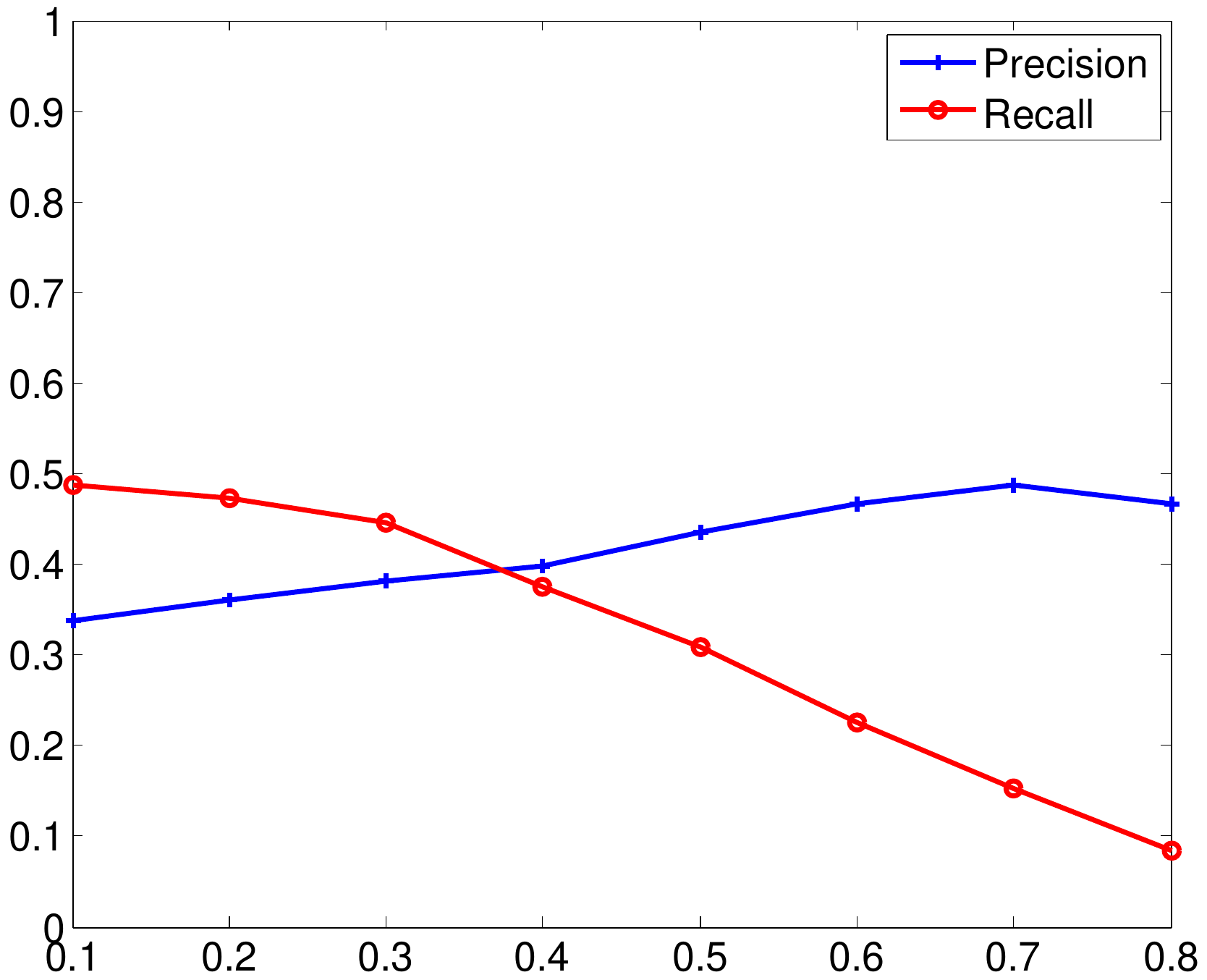} &
\hspace{-3mm}\includegraphics[height =1.2in]{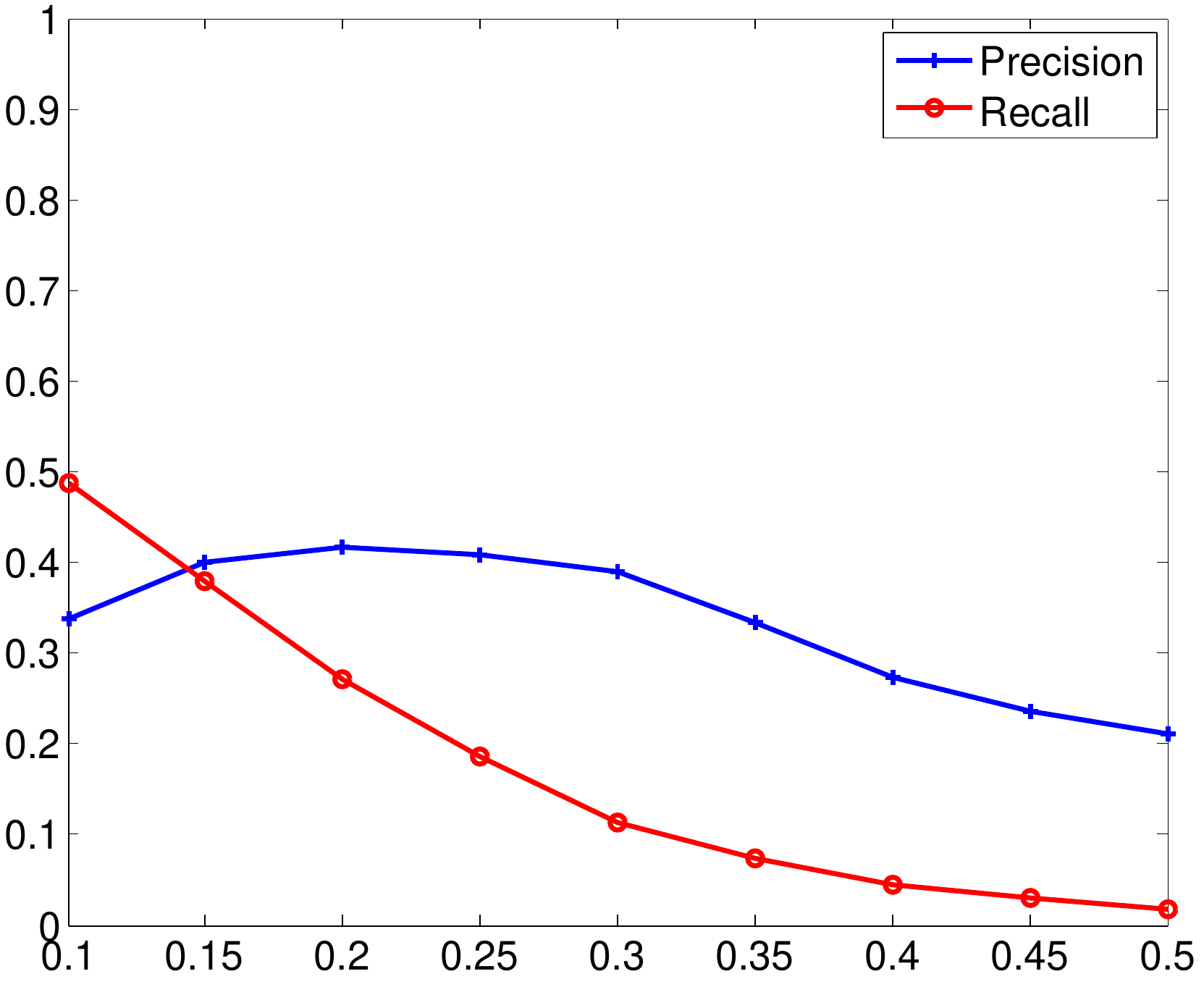} \\
(a) Continuity Ratio & (b) Total Ratio
\end{tabular}
\caption{Quantitative evaluation of triangle detection in portraits. The plots show the precision and recall of our method as a function of (a) continuity ratio and (b) total ratio. }
\label{fig:portrait-curve}
\end{figure}

In Figure~\ref{fig:portrait-curve}, we report the \emph{precision} and \emph{recall} of our method as a function of the continuity ratio and total ratio. 
Specifically, let $G$ denote the set of ground truth triangles annotated by the professional photographer, and $Q$ denote the set of triangles detected by our algorithm under a particular experiment setting, the precision and recall are defined as follows:
\begin{equation}
Precision = \frac{\left| G\bigcap Q\right|}{\left| Q \right|}\;, \quad Recall = \frac{\left| G\bigcap Q\right|}{\left| G \right|}\;.
\end{equation}

\begin{figure}[t!]
\centering
\includegraphics[height =0.88in]{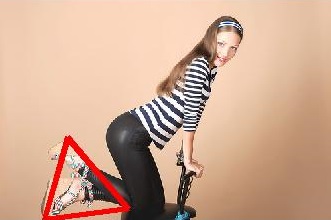} 
\includegraphics[height =0.88in]{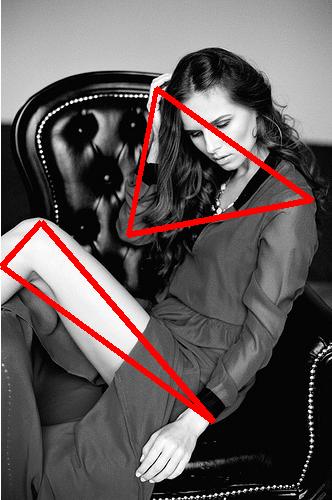} 
\includegraphics[height =0.88in]{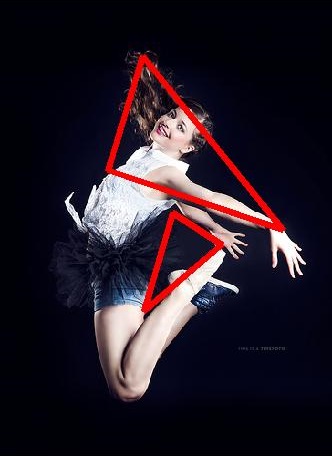} 
\includegraphics[height =0.88in]{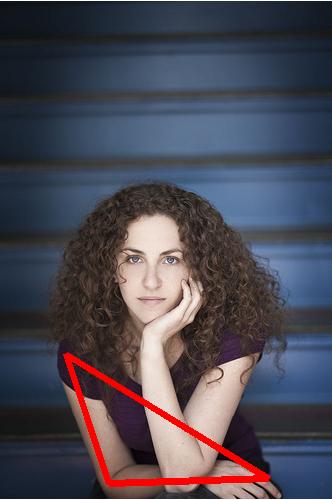} \\
\includegraphics[height =0.89in]{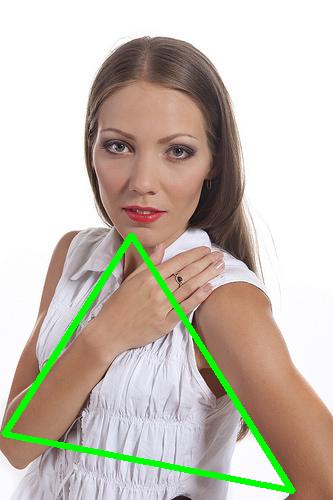} 
\includegraphics[height =0.89in]{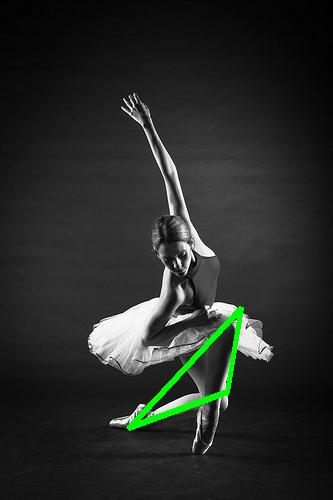} 
\includegraphics[height =0.89in]{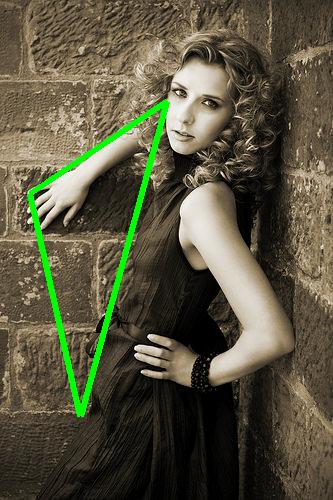} 
\includegraphics[height =0.89in]{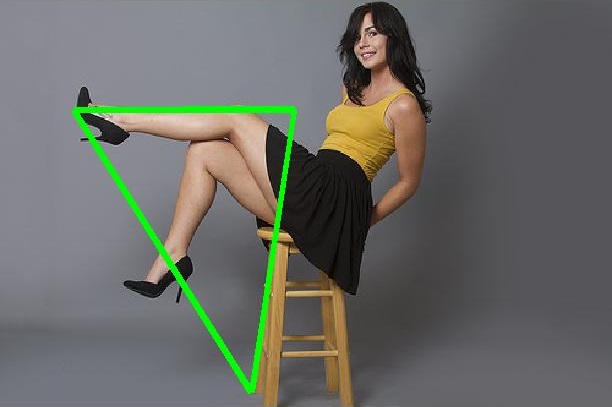} 
\caption{Comparison of triangles detected by our method with professional annotations. {\bf First row:} Triangles annotated by the professional photographer but missed by our algorithm. {\bf Second row:} Some interesting triangles detected by our algorithm, but not labeled by the photographer.}
\label{fig:portrait-compare}
\end{figure}

As shown in Figure~\ref{fig:portrait-curve}(a), the precision of our\break method increases as the continuity ratio increases. When the continuity ratio is high, about half of the triangles detected are true positives. Meanwhile, among all manually annotated triangles, up to about half of them can be detected by our method (\ie, when continuity ratio is $0.1$). In Figure~\ref{fig:portrait-compare} (first row) we show some triangles missed by our algorithm. As one can see, for many of such triangles, there is a lack of explicit edges or contours in the image. For example, in the first image of Figure~\ref{fig:portrait-compare}, the triangle is formed by the pair of shoes and the girl's knees, instead of explicit edges. Such triangles are often known as \emph{implicit triangles} in photography. It will be interesting future work to look for them using machine vision. In addition, Figure~\ref{fig:portrait-curve}(b) shows that our method achieves the best precision when the total ratio is about $0.2$. This may suggest the most commonly seen triangle sizes in portrait photography.

Finally, we examine the triangles that are detected by our algorithm but not labeled by the professional photographer. Figure~\ref{fig:portrait-compare} (second row) shows some interesting cases where we think the triangles are actually meaningful. These examples suggest that even experienced photographers may occasionally overlook certain elements, and our algorithm could potentially provide them with an alternative interpretation of the photo.  

\section{Application in On-Site Composition Feedback}
\label{sec:retrieval}

The proposed triangle technique detection methods capture rich information about the composition of photographs. They can be integrated with various\break composition-driven applications. Here, we discuss an application that aims at providing amateur users with on-site feedback about the composition of their photos. 

\subsection{Natural/Urban Scene Photography}


For natural/urban scene photography, given a photo taken by a user, we propose to find photos with similar compositions in a collection of photos taken by experienced or accomplished photographers. These photos are rendered as feedback to the user. The user can then examine these exemplar photos and consider re-composing his/her own photo accordingly, while the user remains on-site. \citet{YaoSQWL12} pioneered this direction, but the types of composition studied there are limited to a few categories which are pre-defined based on simple 2D rules.

In this paper, we take a completely different approach and develop a similarity measure to compare the composition of two images based on their geometric image segmentation maps. Our observation is that, experienced photographers often are able to achieve different compositions by first placing the dominant vanishing point at different image locations, before choosing how the main structures of the scene are related to it in the captured image. In addition, while the difference in the dominant vanishing point locations can be simply computed as the Euclidean distance between them, our geometric segmentation result offers a natural representation of the arrangement of structures with respect to the dominant vanishing point. Specifically, given two images $I_i$ and $I_j$, let $P_i$ and $P_j$ be the locations of dominant vanishing points and $S_i$ and $S_j$ be the segmentation results generated by our method for these two images, respectively, we define the similarity measure as:\footnote{Here, we assume the two images have the same size, after rescaling.} 
\begin{equation}
D(I_i, I_j) = F(S_i, S_j) + \alpha \|P_i - P_j\|\;,
\label{eq:retrieval}
\end{equation}
where $F(S_i, S_j)$ is a metric to compare two segmentation maps. We adopt the Rand index~\citep{Rand71} for its effectiveness. In addition, $\alpha$ controls the relative impact of the two terms in the equation. We empirically set $\alpha=0.5$.

To obtain a dataset of photos that make good use of the perspective effect, we collect 3,728 images from {\tt flickr.com} by querying the keyword ``vanishing point''. When collecting the photos, we use the sorting criterion of ``interestingness'' provided by {\tt flickr.com}, so that the retrieved photos are likely to be well composed and taken by experienced or accomplished photographers. Each photo is then scaled to size $500\times 330$ or $330\times 500$. To evaluate the effectiveness of our similarity measure (Eq.~\eqref{eq:retrieval}), we manually label the dominant vanishing point and then apply our geometric image segmentation algorithm (with the proposed distance measure $W(e_{ij})$ 
and the stopping criteria $\delta = 0.55$) to obtain a segmentation for each image. 

\begin{figure}[ht!]
\centering
\begin{tabular}{c|c}
\hspace{-1mm}\includegraphics[width=0.41in]{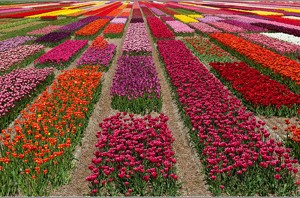}&
\hspace{-1mm}\includegraphics[width=0.41in]{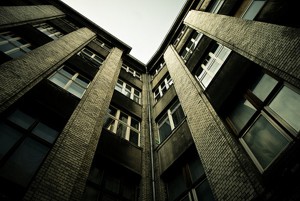}\hspace{0.1mm}
\includegraphics[width=0.41in]{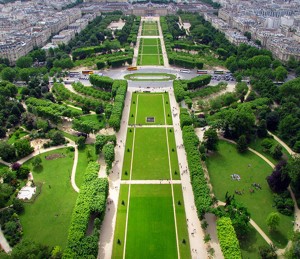}\hspace{0.1mm}
\includegraphics[width=0.41in]{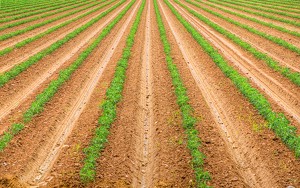}\hspace{0.1mm}
\includegraphics[width=0.41in]{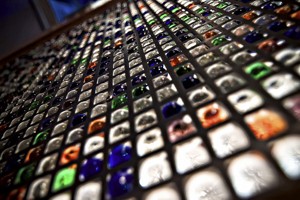}\hspace{0.1mm}
\includegraphics[width=0.41in]{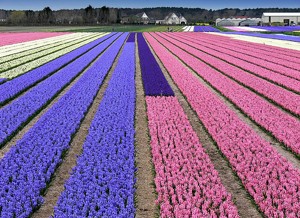}\hspace{0.1mm}
\includegraphics[width=0.41in]{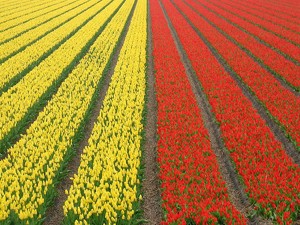}\\
\hspace{-1mm}\includegraphics[width=0.41in]{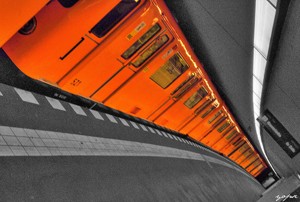}&
\hspace{-1mm}\includegraphics[width=0.41in]{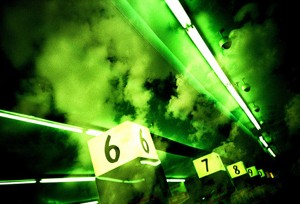}\hspace{0.1mm}
\includegraphics[width=0.41in]{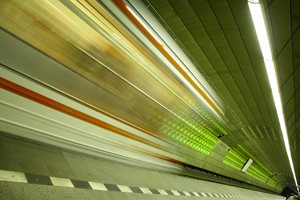}\hspace{0.1mm}
\includegraphics[width=0.41in]{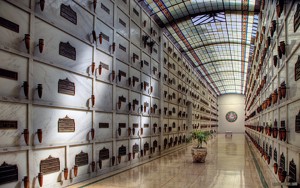}\hspace{0.1mm}
\includegraphics[width=0.41in]{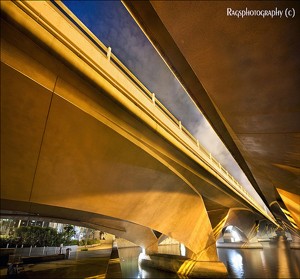}\hspace{0.1mm}
\includegraphics[width=0.41in]{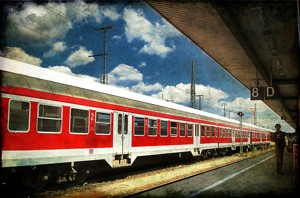}\hspace{0.1mm}
\includegraphics[width=0.41in]{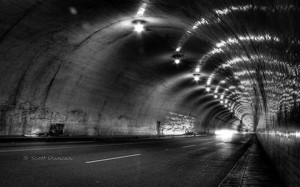}\\
\hspace{-1mm}\includegraphics[width=0.41in]{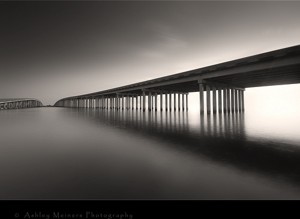}&
\hspace{-1mm}\includegraphics[width=0.41in]{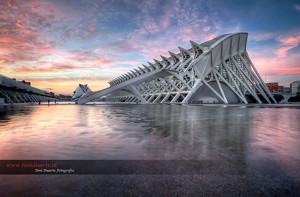}\hspace{0.1mm}
\includegraphics[width=0.41in]{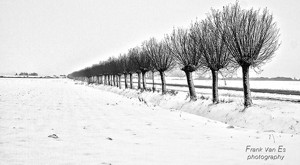}\hspace{0.1mm}
\includegraphics[width=0.41in]{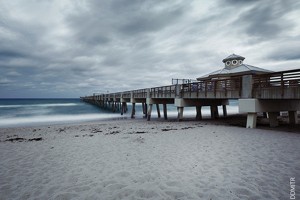}\hspace{0.1mm}
\includegraphics[width=0.41in]{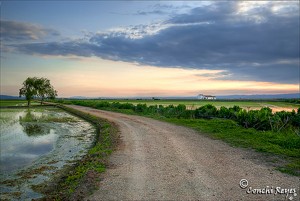}\hspace{0.1mm}
\includegraphics[width=0.41in]{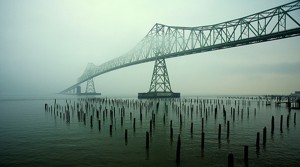}\hspace{0.1mm}
\includegraphics[width=0.41in]{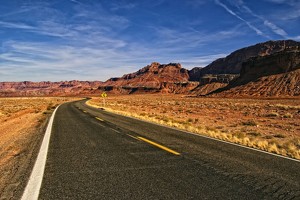}\\
\hspace{-1mm}\includegraphics[width=0.41in]{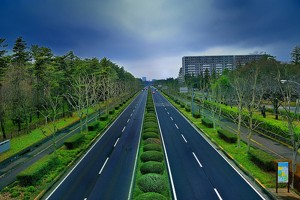}&
\hspace{-1mm}\includegraphics[width=0.41in]{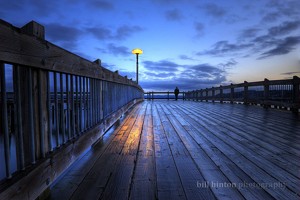}\hspace{0.1mm}
\includegraphics[width=0.41in]{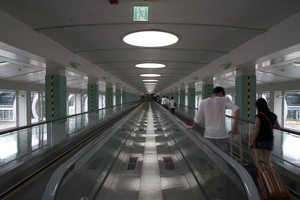}\hspace{0.1mm}
\includegraphics[width=0.41in]{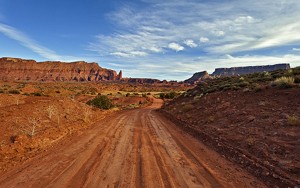}\hspace{0.1mm}
\includegraphics[width=0.41in]{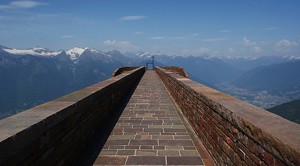}\hspace{0.1mm}
\includegraphics[width=0.41in]{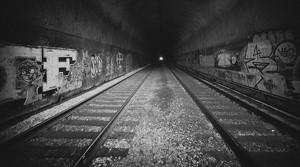}\hspace{0.1mm}
\includegraphics[width=0.41in]{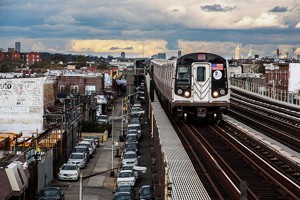}\\
\hspace{-1mm}\includegraphics[width=0.41in]{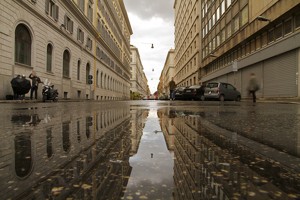}&
\hspace{-1mm}\includegraphics[width=0.41in]{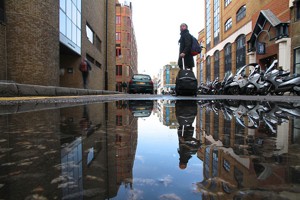}\hspace{0.1mm}
\includegraphics[width=0.41in]{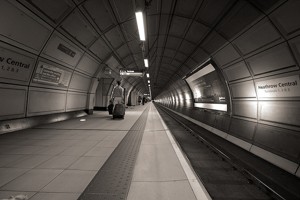}\hspace{0.1mm}
\includegraphics[width=0.41in]{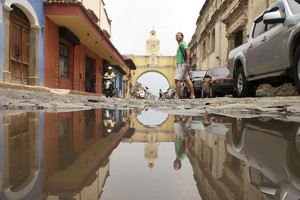}\hspace{0.1mm}
\includegraphics[width=0.41in]{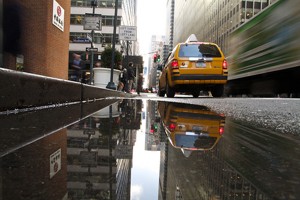}\hspace{0.1mm}
\includegraphics[width=0.41in]{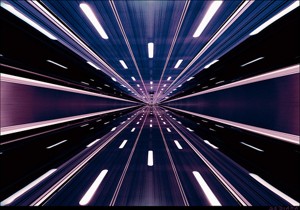}\hspace{0.1mm}
\includegraphics[width=0.41in]{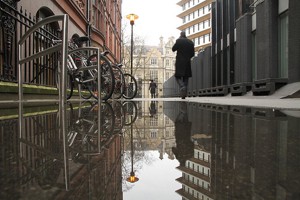}\\
\hspace{-1mm}\includegraphics[width=0.41in]{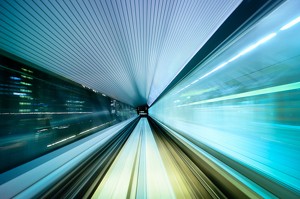}&
\hspace{-1mm}\includegraphics[width=0.41in]{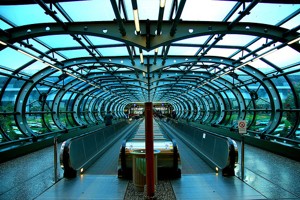}\hspace{0.1mm}
\includegraphics[width=0.41in]{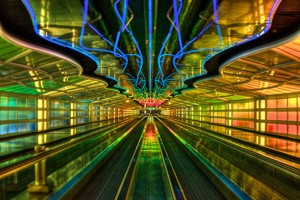}\hspace{0.1mm}
\includegraphics[width=0.41in]{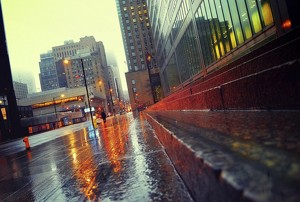}\hspace{0.1mm}
\includegraphics[width=0.41in]{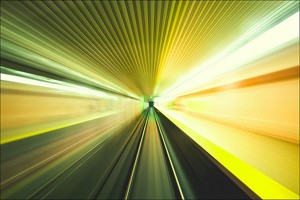}\hspace{0.1mm}
\includegraphics[width=0.41in]{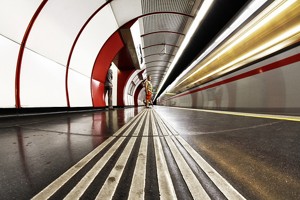}\hspace{0.1mm}
\includegraphics[width=0.41in]{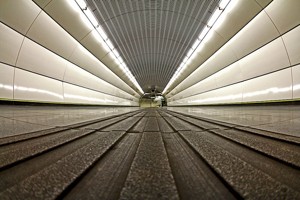}\\
\hspace{-1mm}\includegraphics[width=0.41in]{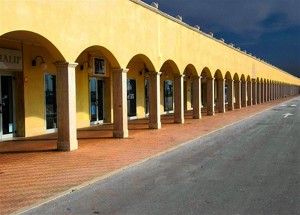}\hspace{-1mm}&
\hspace{-1mm}\includegraphics[width=0.41in]{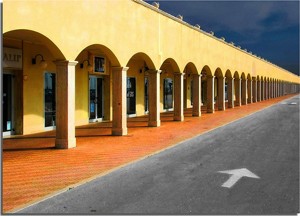}\hspace{0.1mm}
\includegraphics[width=0.41in]{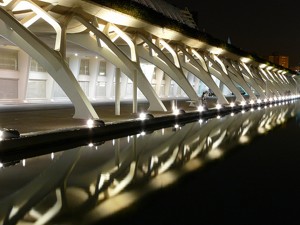}\hspace{0.1mm}
\includegraphics[width=0.41in]{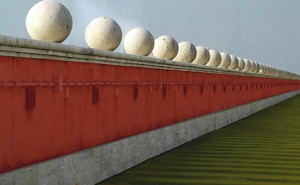}\hspace{0.1mm}
\includegraphics[width=0.41in]{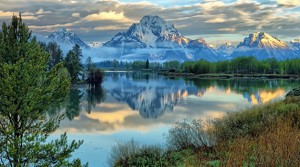}\hspace{0.1mm}
\includegraphics[width=0.41in]{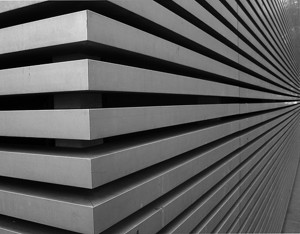}\hspace{0.1mm}
\includegraphics[width=0.41in]{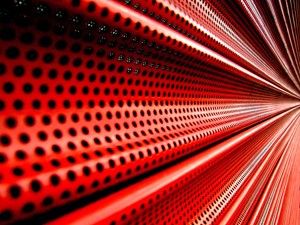}\\
\hspace{-1mm}\includegraphics[width=0.41in]{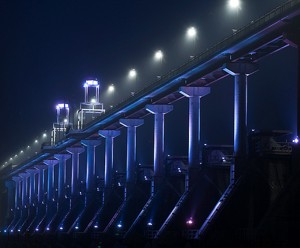}\hspace{-1mm}&
\hspace{-1mm}\includegraphics[width=0.41in]{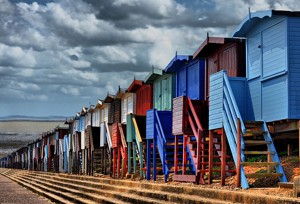}\hspace{0.1mm}
\includegraphics[width=0.41in]{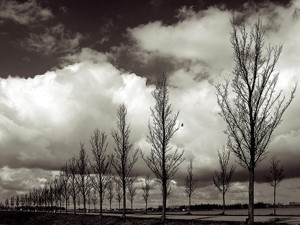}\hspace{0.1mm}
\includegraphics[width=0.41in]{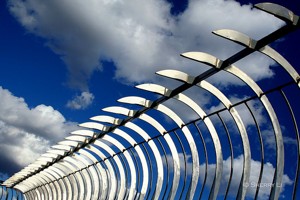}\hspace{0.1mm}
\includegraphics[width=0.41in]{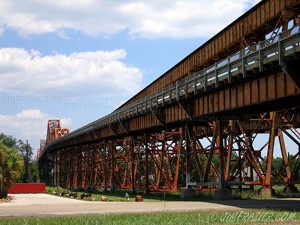}\hspace{0.1mm}
\includegraphics[width=0.41in]{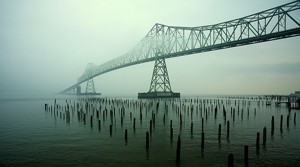}\hspace{0.1mm}
\includegraphics[width=0.41in]{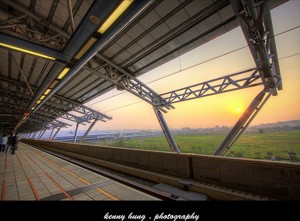}\\
\end{tabular}
\begin{tabular}{c|c}
\hspace{-1mm}\includegraphics[width =0.31in]{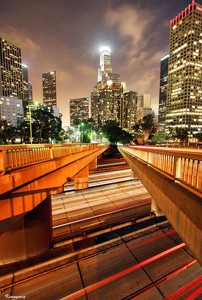}\hspace{-1mm}&
\hspace{-1mm}\includegraphics[width =0.31in]{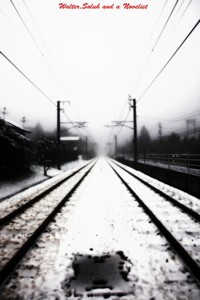}\hspace{0.1mm}
\includegraphics[width =0.31in]{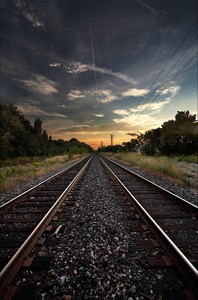}\hspace{0.1mm}
\includegraphics[width =0.31in]{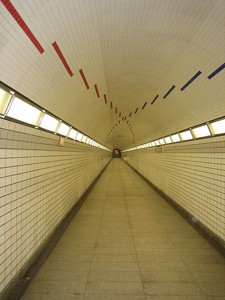}\hspace{0.1mm}
\includegraphics[width =0.31in]{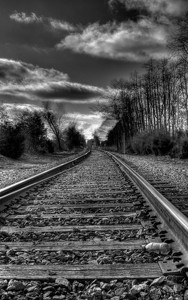}\hspace{0.1mm}
\includegraphics[width =0.31in]{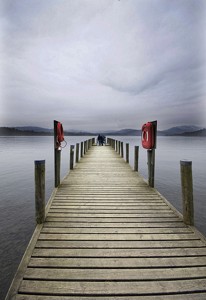}\hspace{0.1mm}
\includegraphics[width =0.31in]{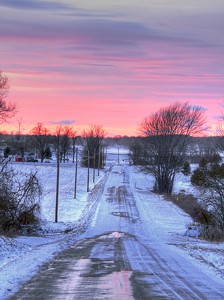}\hspace{0.1mm}
\includegraphics[width =0.31in]{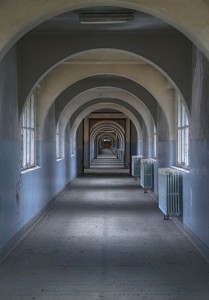}\hspace{0.1mm}
\includegraphics[width =0.31in]{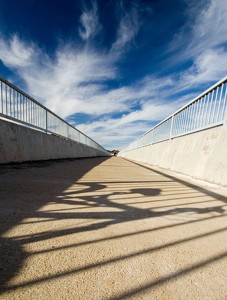}\\
\hspace{-1mm}\includegraphics[width =0.31in]{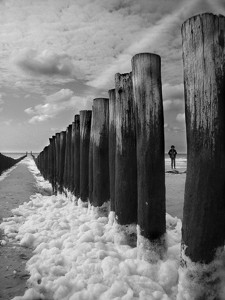}\hspace{-1mm}&
\hspace{-1mm}\includegraphics[width =0.31in]{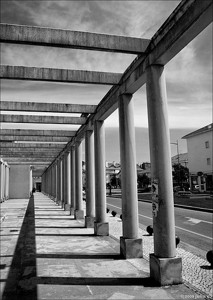}\hspace{0.1mm}
\includegraphics[width =0.31in]{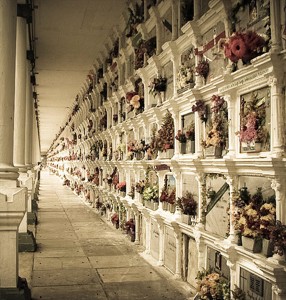}\hspace{0.1mm}
\includegraphics[width =0.31in]{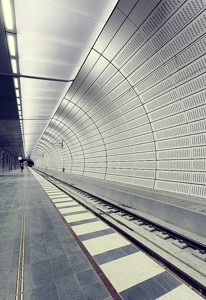}\hspace{0.1mm}
\includegraphics[width =0.31in]{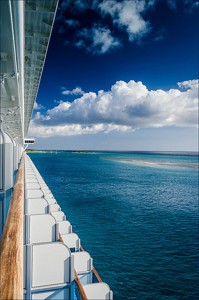}\hspace{0.1mm}
\includegraphics[width =0.31in]{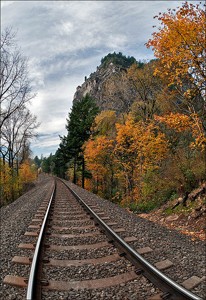}\hspace{0.1mm}
\includegraphics[width =0.31in]{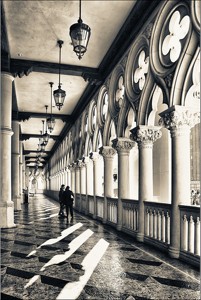}\hspace{0.1mm}
\includegraphics[width =0.31in]{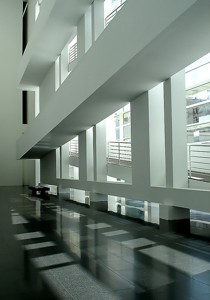}\hspace{0.1mm}
\includegraphics[width =0.31in]{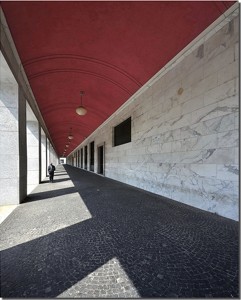}\\
\hspace{-1mm}\includegraphics[width =0.31in]{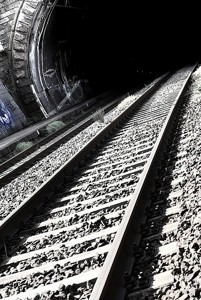}\hspace{-1mm}&
\hspace{-1mm}\includegraphics[width =0.31in]{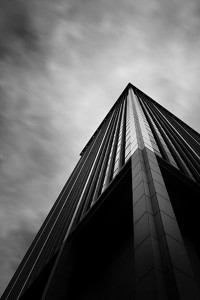}\hspace{0.1mm}
\includegraphics[width =0.31in]{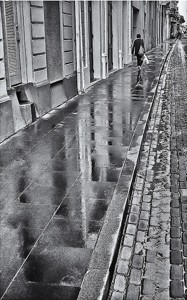}\hspace{0.1mm}
\includegraphics[width =0.31in]{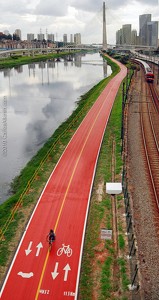}\hspace{0.1mm}
\includegraphics[width =0.31in]{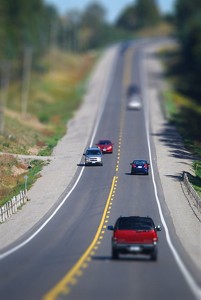}\hspace{0.1mm}
\includegraphics[width =0.31in]{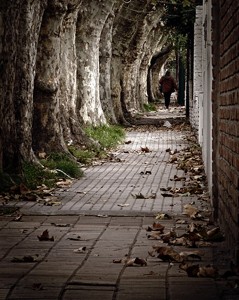}\hspace{0.1mm}
\includegraphics[width =0.31in]{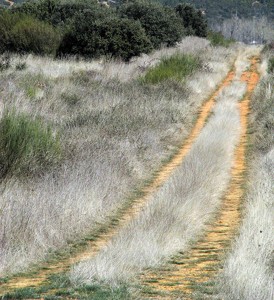}\hspace{0.1mm}
\includegraphics[width =0.31in]{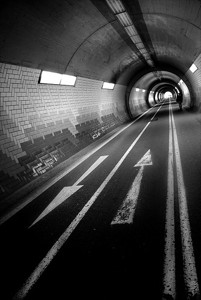}\hspace{0.1mm}
\includegraphics[width =0.31in]{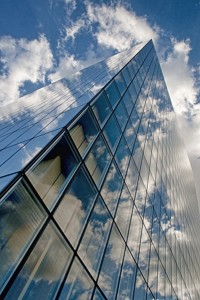}\\
\end{tabular}
\caption{Composition-sensitive image retrieval results. Each row shows a query image (first image from the left) and the top-6 or top-8 ranked images retrieved.}
\label{fig:rerank}
\end{figure}

In Figure~\ref{fig:rerank}, we show the retrieved images for various query images. The results clearly show that the proposed measure is not only able to find images with similar dominant vanishing point locations, but also effectively captures how each region in the image is related to the vanishing point. 

\subsubsection{Comparison to Existing Retrieval Systems}

We also compare our composition-sensitive image retrieval system to the following recent retrieval pipelines:  

\begin{figure*}[t!]
\centering
\begin{tabular}{cc}
\begin{tabular}{cl}
\hspace{-2mm}\cfbox{blue}{\includegraphics[height=0.41in]{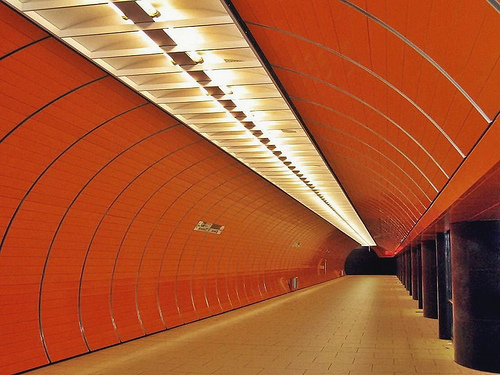}}&\hspace{-2mm}\includegraphics[height=0.41in]{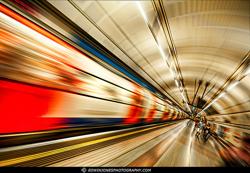}
\hspace{0.1mm}\includegraphics[height=0.41in]{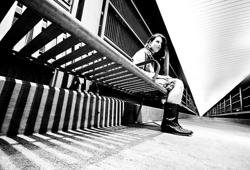}
\hspace{0.1mm}\includegraphics[height=0.41in]{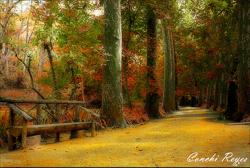}
\hspace{0.1mm}\includegraphics[height=0.41in]{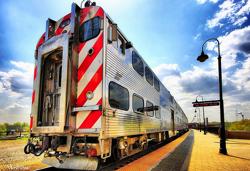}\\
&\hspace{-2mm}\includegraphics[height=0.41in]{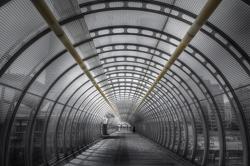}
\hspace{0.1mm}\includegraphics[height=0.41in]{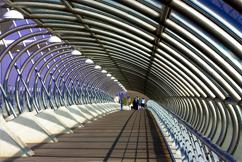}
\hspace{0.1mm}\includegraphics[height=0.41in]{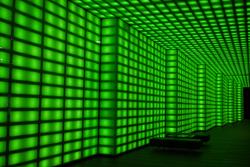}
\hspace{0.1mm}\includegraphics[height=0.41in]{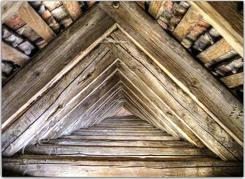}\\
&\hspace{-2mm}\includegraphics[height=0.39in]{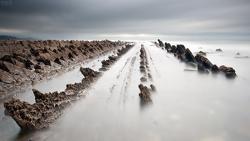}
\hspace{0.1mm}\includegraphics[height=0.39in]{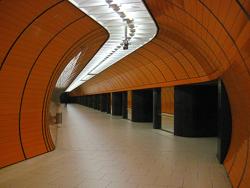}
\hspace{0.1mm}\includegraphics[height=0.39in]{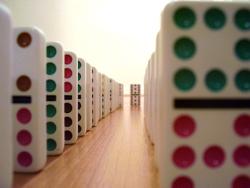}
\hspace{0.1mm}\includegraphics[height=0.39in]{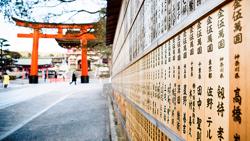}\\
&\hspace{-2mm}\includegraphics[height=0.52in]{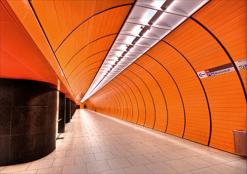}
\hspace{0.1mm}\includegraphics[height=0.52in]{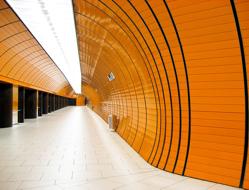}
\hspace{0.1mm}\includegraphics[height=0.52in]{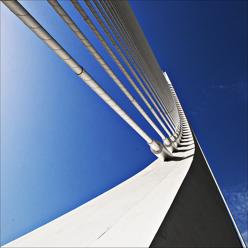}
\hspace{0.1mm}\includegraphics[height=0.52in]{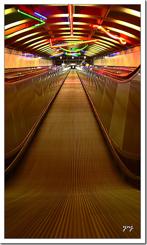}\\
\end{tabular}&\hspace{-2mm}\begin{tabular}{cl}
\hspace{-2mm}\cfbox{blue}{\includegraphics[height=0.39in]{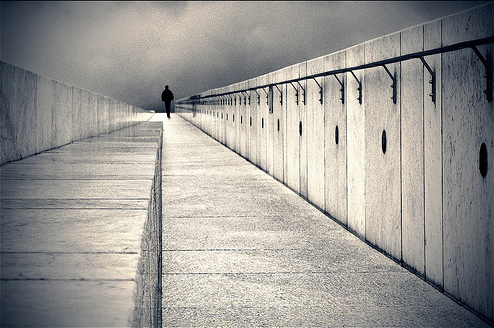}}&\hspace{-2mm}\includegraphics[height=0.39in]{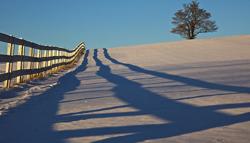}
\hspace{0.1mm}\includegraphics[height=0.39in]{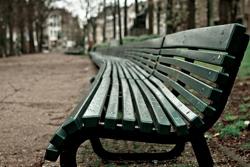}
\hspace{0.1mm}\includegraphics[height=0.39in]{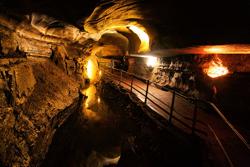}
\hspace{0.1mm}\includegraphics[height=0.39in]{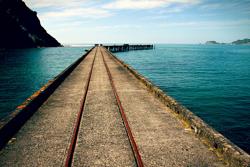}\\
&\hspace{-2mm}\includegraphics[height=0.41in]{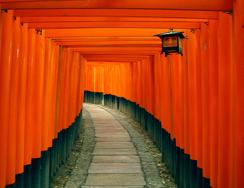}
\hspace{0.1mm}\includegraphics[height=0.41in]{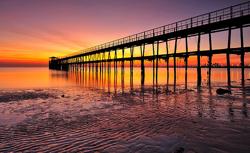}
\hspace{0.1mm}\includegraphics[height=0.41in]{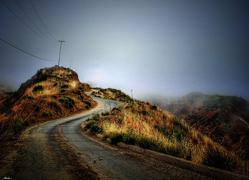}
\hspace{0.1mm}\includegraphics[height=0.41in]{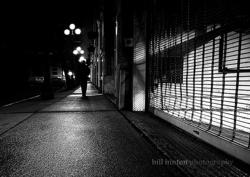}\\
&\hspace{-2mm}\includegraphics[height=0.52in]{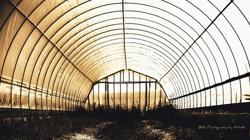}
\hspace{0.1mm}\includegraphics[height=0.52in]{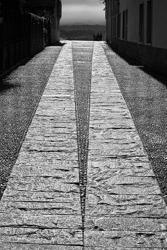}
\hspace{0.1mm}\includegraphics[height=0.52in]{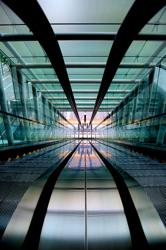}
\hspace{0.1mm}\includegraphics[height=0.52in]{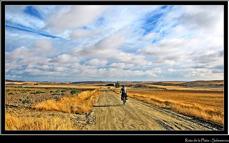}\\
&\hspace{-2mm}\includegraphics[height=0.52in]{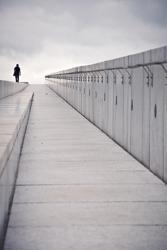}
\hspace{0.1mm}\includegraphics[height=0.52in]{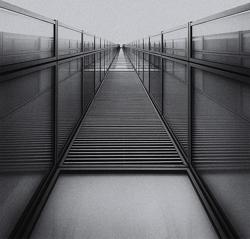}
\hspace{0.1mm}\includegraphics[height=0.52in]{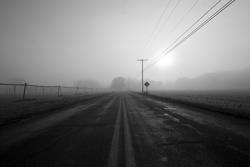}
\hspace{0.1mm}\includegraphics[height=0.52in]{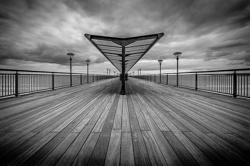}\\
\end{tabular}
\end{tabular}
\vspace{0.1in}
\begin{tabular}{cc}
\begin{tabular}{cl}
\hspace{-2mm}\cfbox{blue}{\includegraphics[height=0.48in]{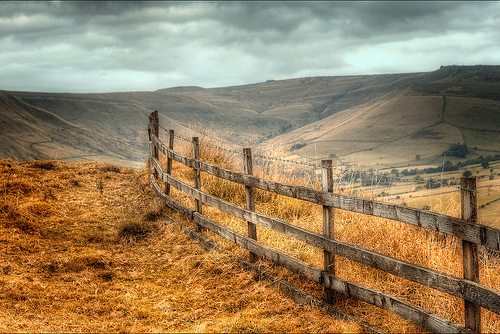}}&
\hspace{-2mm}\includegraphics[height=0.48in]{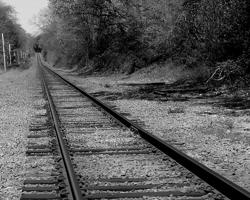}
\hspace{0.1mm}\includegraphics[height=0.48in]{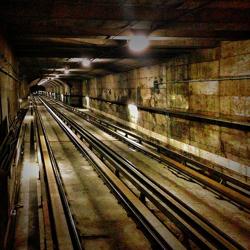}
\hspace{0.1mm}\includegraphics[height=0.48in]{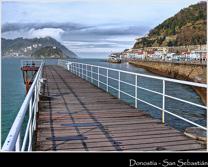}
\hspace{0.1mm}\includegraphics[height=0.48in]{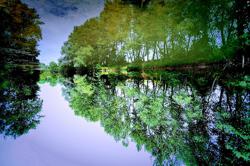}\\
&
\hspace{-2mm}\includegraphics[height=0.39in]{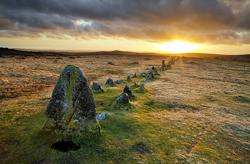}
\hspace{0.1mm}\includegraphics[height=0.39in]{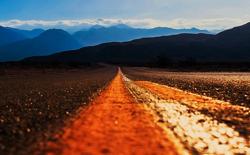}
\hspace{0.1mm}\includegraphics[height=0.39in]{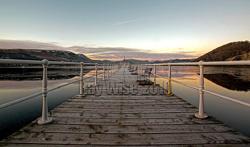}
\hspace{0.1mm}\includegraphics[height=0.39in]{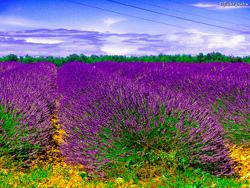}\\
&
\hspace{-2mm}\includegraphics[height=0.55in]{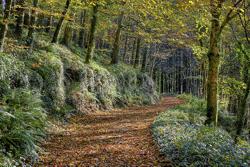}
\hspace{0.1mm}\includegraphics[height=0.55in]{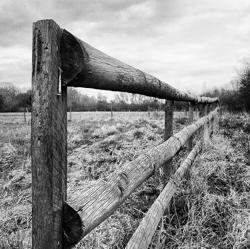}
\hspace{0.1mm}\includegraphics[height=0.55in]{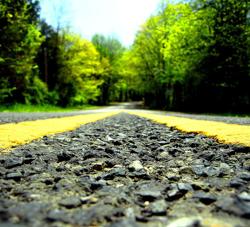}
\hspace{0.1mm}\includegraphics[height=0.55in]{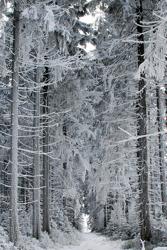}\\
&
\hspace{-2mm}\includegraphics[height=0.44in]{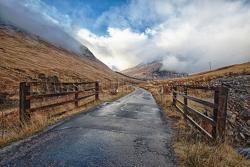}
\hspace{0.1mm}\includegraphics[height=0.44in]{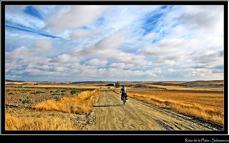}
\hspace{0.1mm}\includegraphics[height=0.44in]{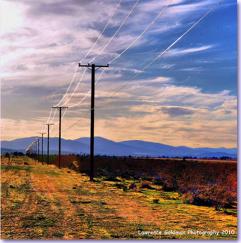}
\hspace{0.1mm}\includegraphics[height=0.44in]{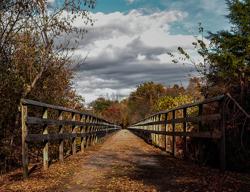}\\
\end{tabular}&\hspace{-2mm}\begin{tabular}{cl}
\hspace{-2mm}\cfbox{blue}{\includegraphics[height=0.56in]{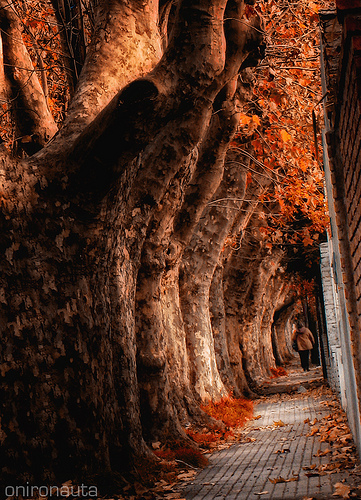}}&
\hspace{-2mm}\includegraphics[height=0.56in]{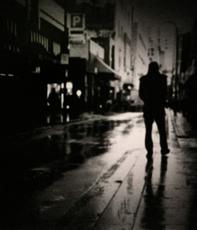}
\hspace{0.1mm}\includegraphics[height=0.56in]{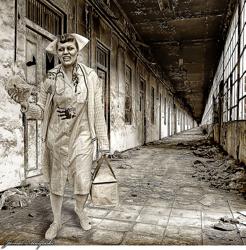}
\hspace{0.1mm}\includegraphics[height=0.56in]{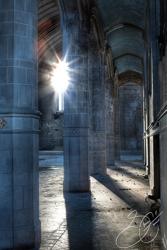}
\hspace{0.1mm}\includegraphics[height=0.56in]{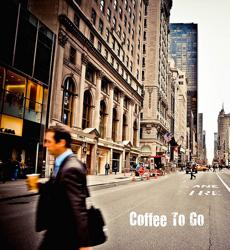}
\hspace{0.1mm}\includegraphics[height=0.56in]{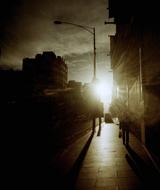}\\
&
\hspace{-2mm}\includegraphics[height=0.61in]{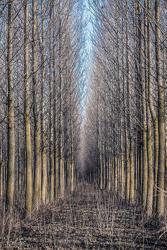}
\hspace{0.1mm}\includegraphics[height=0.61in]{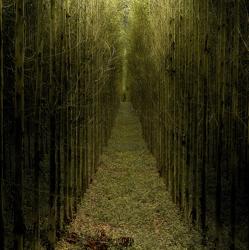}
\hspace{0.1mm}\includegraphics[height=0.61in]{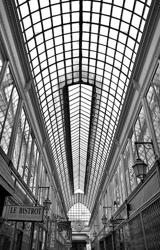}
\hspace{0.1mm}\includegraphics[height=0.61in]{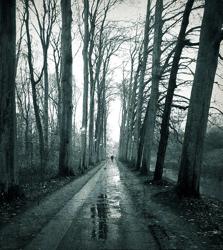}
\hspace{0.1mm}\includegraphics[height=0.61in]{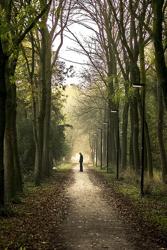}\\
&
\hspace{-2mm}\includegraphics[height=0.435in]{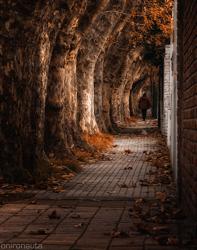}
\hspace{0.1mm}\includegraphics[height=0.435in]{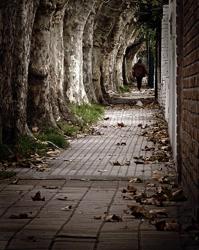}
\hspace{0.1mm}\includegraphics[height=0.435in]{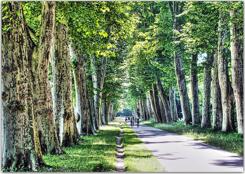}
\hspace{0.1mm}\includegraphics[height=0.435in]{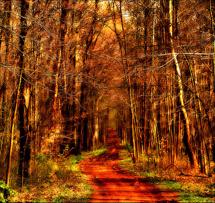}
\hspace{0.1mm}\includegraphics[height=0.435in]{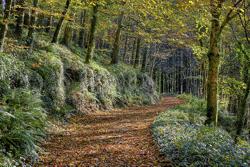}\\
&
\hspace{-2mm}\includegraphics[height=0.39in]{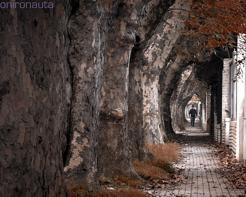}
\hspace{0.1mm}\includegraphics[height=0.39in]{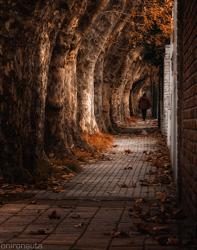}
\hspace{0.1mm}\includegraphics[height=0.39in]{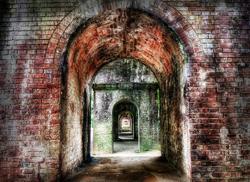}
\hspace{0.1mm}\includegraphics[height=0.39in]{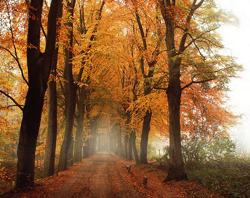}
\hspace{0.1mm}\includegraphics[height=0.39in]{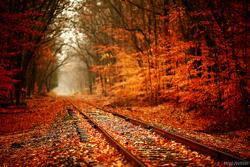}\\
\end{tabular}
\end{tabular}
\caption{Comparison of four retrieval systems for images of similar composition. For a query image, we show the top-4 or top-5 images retrieved by four different systems, where each row corresponds to one system. The four rows are ordered as follows: {\bf First row:} our system. {\bf Second row:} HOG. {\bf Third row:} VLAD. {\bf Fourth row:} CNN.}
\label{fig:retrieval1}
\end{figure*}

\smallskip
\noindent{\bf HOG:} We represent each image with a rigid grid-like HOG feature $\x_i$~\citep{DalalT05, FelzenszwalbGMR10}. In order to limit the dimensionality of HOG features to roughly $5K$, we resize the images to $150\times 100$ or $100\times 150$, and use a cell size of 8 pixels. As suggested by~\citet{ShrivastavaMGE11}, we further normalize the feature vector by subtracting its mean: $\x_i = \x_i - mean(\x_i)$, and use the cosine distance to measure the similarity of two vectors.

\smallskip
\noindent{\bf VLAD:} The vector of locally aggregated descriptors (VLAD) is a feature coding and pooling method~\citep{JegouDSP10, ArandjelovicZ13}. It encodes a set of local feature descriptors (\eg, SIFT features) extracted from an image using a dictionary built using a clustering method such as GMM or K-means. In this paper, we use the code provided on the authors' website\footnote{http://people.rennes.inria.fr/Herve.Jegou/software.html} with pre-trained dictionary to extract the VLAD descriptor for each image, and compare different descriptors using the $\ell_2$ distance.

\smallskip
\noindent{\bf CNN:} Generic descriptors extracted from the convolutional neural networks (CNNs) have been shown to be very powerful in tackling a diverse range of computer vision problems, including image retrieval~\citep{RazavianASC14}. In this paper, we use the publicly available code and model\footnote{http://www.vlfeat.org/matconvnet/pretrained/} by~\citet{ChatfieldSVZ14}, which were developed to perform classification in the ImageNet ILSVRC challenge data, and represent each image using the $\ell_2$-normalized output of the second fully connected layer (full7 of~\citep{ChatfieldSVZ14}). The feature similarity is measured by the cosine distance.

\smallskip
Figure~\ref{fig:retrieval1} shows the top-ranked images retrieved by all four systems for some example query images. As can be seen, the images retrieved by our system is more compositionally relevant in terms of the use of perspective effect than other systems. Among the three existing systems, HOG is shown to be more sensitive to the image composition, as it is based on the local image gradients. Meanwhile, VLAD and CNN features are known to preform well in capturing the semantics of a scene (\ie, scene types and objects). While both methods indeed retrieved more relevant images semantically, they are not sensitive to the image composition. 

\smallskip
\noindent{\bf Quantitative Evaluation.} Unlike traditional image retrieval tasks, currently there is no dataset with ground truth composition labels available. In order to quantitatively evaluate the performance of our system, we have conducted a user study that allows participants to manually rank the performance of the four systems based on \emph{their ability to retrieve compositionally similar images}. 

In this study, a collection of 200 query images were randomly selected to form the dataset for the comparison study. At an online website, each participant is provided with a subset of 15 randomly selected query images. For each query, we show the participant the top-8 images retrieved by all four systems. Then, we ask the users to rank the performance of the four systems without providing them with any information about the four systems. To avoid any biases, we also randomly shuffled the order in which the results of the four systems are presented on each page.

\begin{figure}[t!]
\centering
\includegraphics[height =1.8in]{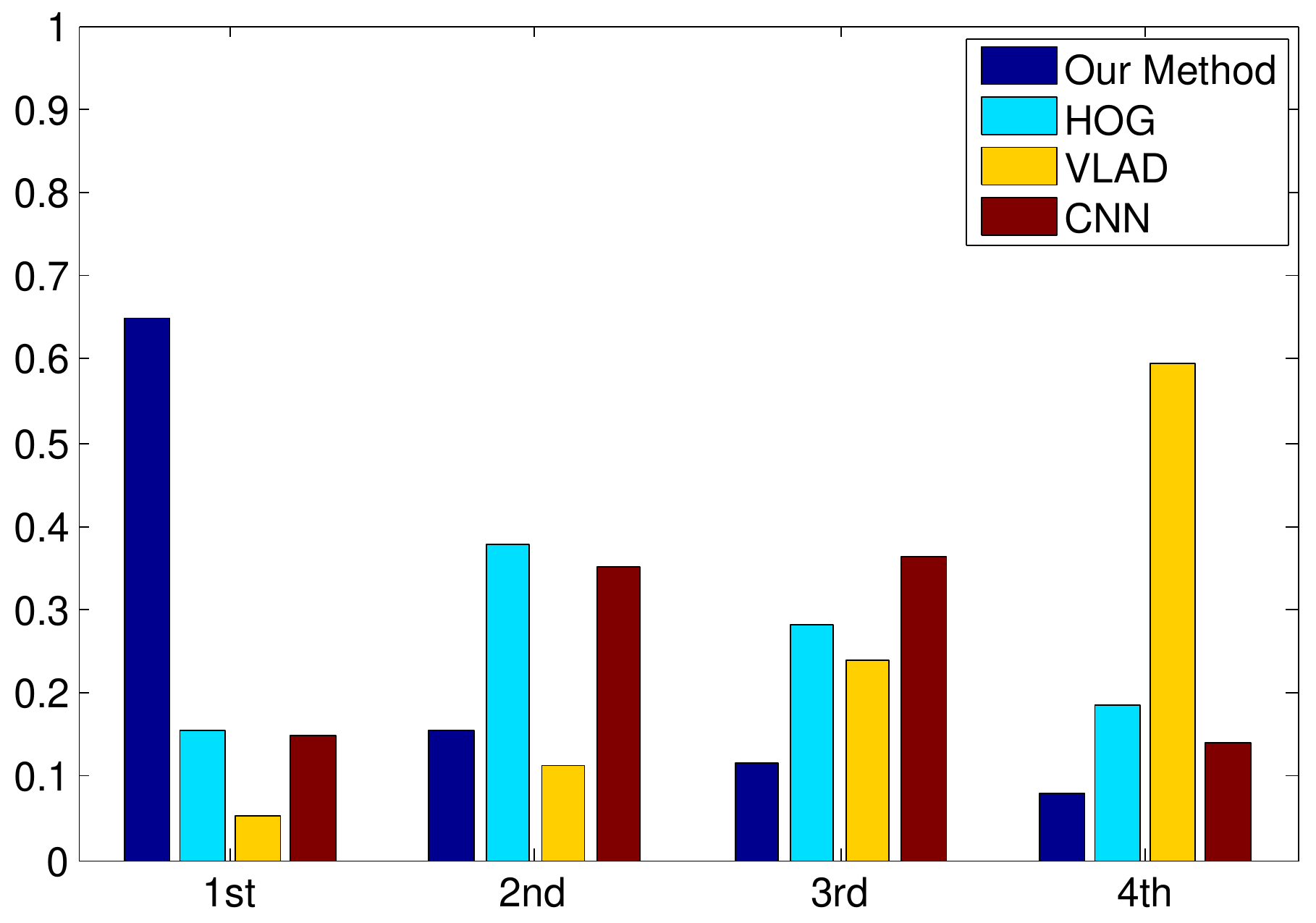}
\caption{Quantitative evaluation of retrieval systems. Percentages of user
selections are shown. Our system is ranked the first by the participants 
$65.6\%$ of the time.}
\label{fig:vote}
\end{figure}

To complete this study, we recruited 20 participants, mostly graduate students at Penn State with some basic 
photography knowledge. Figure~\ref{fig:vote} shows the overall percentage of times that each system is ranked the first, the second, the third, and the fourth, respectively. Our system is ranked the first $65.6\%$ of the time, which is significantly higher than the other systems. Among the other three systems, HOG outperforms VLAD and CNN, thanks to its sensitivity to local image gradients, which is correlated with the vanishing direction of an image. Interestingly, while CNN is trained to capture scene semantics, it is voted first for $15\%$ of the time. This suggests a link between semantics (\ie, what is in the image) and image composition (\ie, how an image is taken).

\subsection{Portrait Photography}
\label{sec:retrieval-portrait}

Our triangle technique detection method can potentially help amateur photographers design interesting poses when taking portraits. It is often 
difficult for untrained photographers to naturally embed triangles in compositions in order to form striking portraits.  
Certain professional users, {\it e.g.}, magazine editors, can also benefit from the triangle detection method. When they are selecting portraits to fill a specific page layout, the shapes and orientations of embedded triangles within these photos can be critical to the overall page composition. Therefore, we develop a portrait retrieval system that can take a \emph{triangle sketch} as a query to help users find images containing a targeted triangular configuration. Such tools can be especially useful when a large collection of portraits are available to choose from.

First, users need to provide a sketch indicating the shape and the orientation of the triangular configuration that they desire to have in the photos. In portrait photos, the third side of a triangle is often missing in most photos. Thus, the sketch query provided by users is basically an angle with two sides. Given the sketch angle, we compute the \emph{orientations} of the two sides as well as the \emph{opening direction} of the angle. Specifically, the orientation of a side is defined to be the angle between its extended straight line and the positive $x$-axis. An angle has four possible opening directions: upward, downward, leftward, or rightward. The two properties narrow down the searching space to a specific type of triangles.

Recall that our triangle detection algorithm has two steps: line segment detection and fitting triangles. All line segments detected from the first step are taken as candidate triangle sides during the fitting stage. 
In the retrieval system, we construct two candidate sets containing line segments with similar orientations as the two sides of the sketched angle. Two sides can then be randomly selected from the two candidate sets and the combination of them generates four angles with four different opening directions (Figure~\ref{fig:fittedpix}). Only the angle that has the same opening direction as the sketched angle will be taken into consideration during the fitting process. Such a sketch-based triangle retrieval system not only assists users in searching for desired types of triangles but also reduces the searching time significantly for large photo collections.

\subsubsection{User Study}

We conducted a human subject study to verify the effectiveness of the retrieved triangles in conveying valuable information about composition to amateurs. In this study, we selected 20 groups of representative queries which
covered a wide range of angles in terms of magnitude and orientation. We only consider angles in the range of $\left [45^{\circ}, 135^{\circ}\right ]$ because angles that are either too large or too small are often not perceived as interesting ones by humans. Each of the 20 groups of queries takes a distinct combination of orientations for two straight lines. Twenty line combinations $\left \langle l_1, l_2 \right \rangle$
are selected in our experiment such that the angle between $l_1$ and $l_2$ falls in the closed range of $\left [45^{\circ}, 135^{\circ}\right ]$ and the angle between $l_1/l_2$ and positive $x$-axis falls in \{0$^\circ$, 22.5$^\circ$, 45$^\circ$, 67.5$^\circ$, 90$^\circ$, 112.5$^\circ$, 135$^\circ$\}. Moreover, one combination of two straight lines generates four possible angles which differ in terms of their opening directions. 
Therefore, we use a total of 80 queries to retrieve triangles from 4,451 photos where each photo may contain many distinct triangles. For each query, we rank the results based on their continuity ratios. Higher continuity ratios represent higher quality of fitting and thus imply more accurate retrieved results. 

We recruited 20 participants to this study, mostly graduate students at Penn State with some basic 
photography knowledge. At an online website, each participant is provided with a subset of 15 randomly selected queries. For each query, we show the participant the top-20 triangles retrieved by our system. For each retrieved triangle, we ask the participant to assess whether it is ``useful'', {\it i.e.}, whether it indicates interesting pose or composition in the image, and help the participant understand the use of triangles in the photo. 


\begin{figure}[t!]
\centering
\includegraphics[height=1.8in]{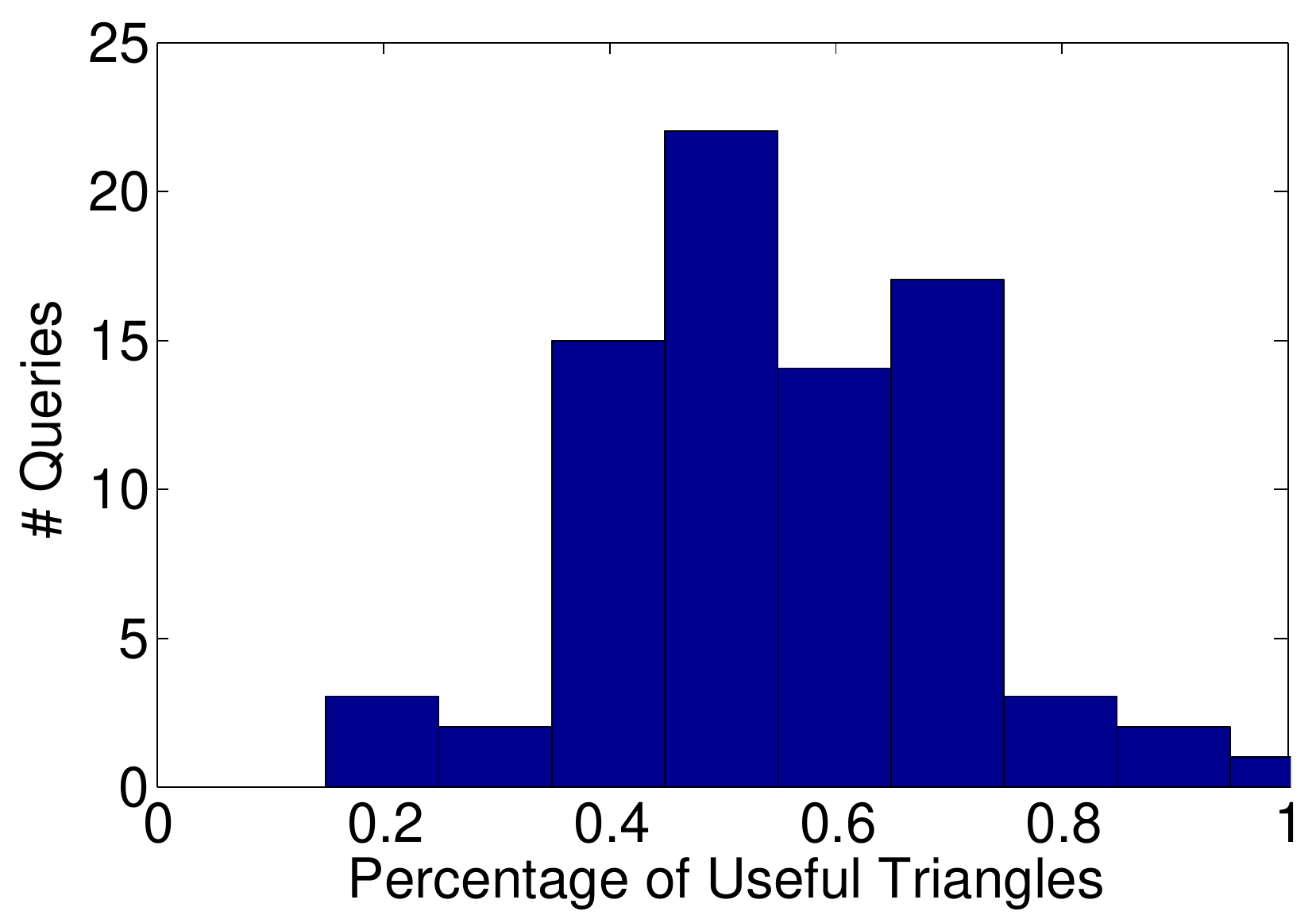}
\caption{The performance of the retrieval system for portraits.}
\label{fig:perf}
\end{figure}

Figure~\ref{fig:perf} shows the histogram of overall percentages of ``useful'' triangles for all 80 queries. For most queries, between $40\%$ and $80\%$ of the retrieved results are considered by the participants as providing useful information and guidance on the portrait composition. Overall, $53.8\%$ of the retrieved triangles are regarded as ``useful'' by the users.

\begin{figure}[ht!]
	\begin{center}
		\begin{tabular}{c|l}
			\includegraphics[height=0.05\textwidth]{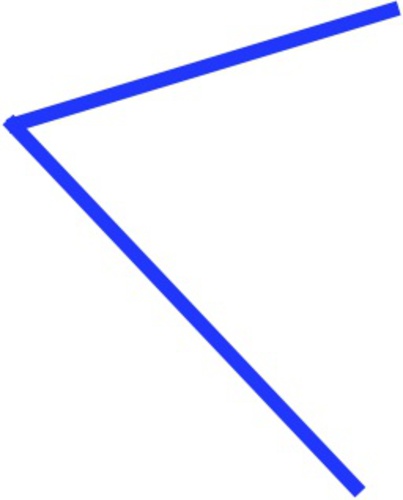} &
			\includegraphics[height=0.065\textwidth,keepaspectratio]{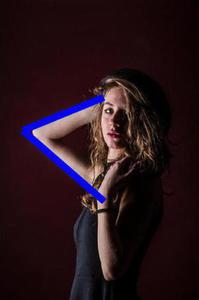} 
			\includegraphics[height=0.065\textwidth,keepaspectratio]{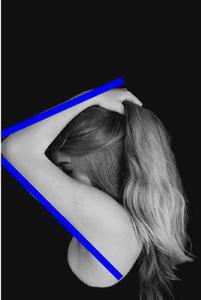} 
			\includegraphics[height=0.065\textwidth,keepaspectratio]{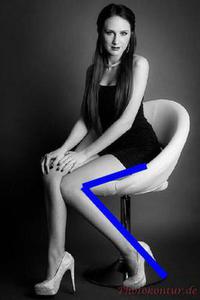} 
			\includegraphics[height=0.065\textwidth,keepaspectratio]{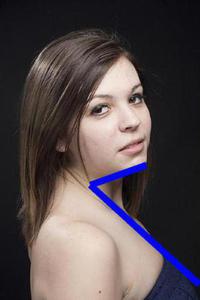} 
			\includegraphics[height=0.065\textwidth,keepaspectratio]{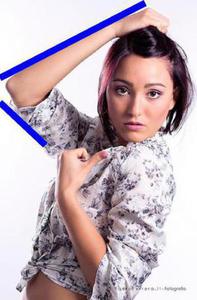} 
			\includegraphics[height=0.065\textwidth,keepaspectratio]{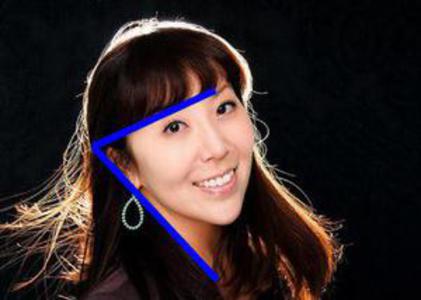} \\
			
			\includegraphics[height=0.05\textwidth]{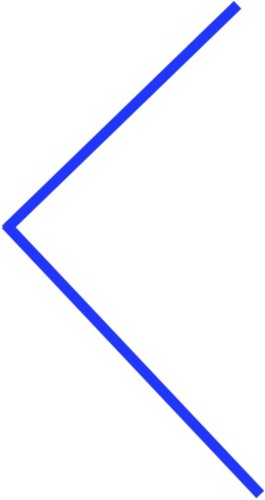} &
			\includegraphics[height=0.065\textwidth,keepaspectratio]{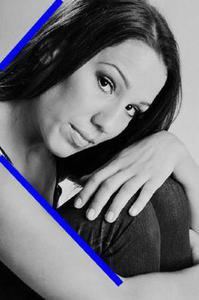} 
			\includegraphics[height=0.065\textwidth,keepaspectratio]{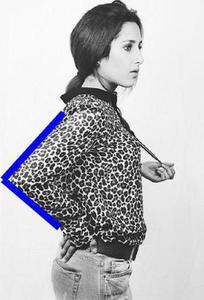}
			\includegraphics[height=0.057\textwidth,keepaspectratio]{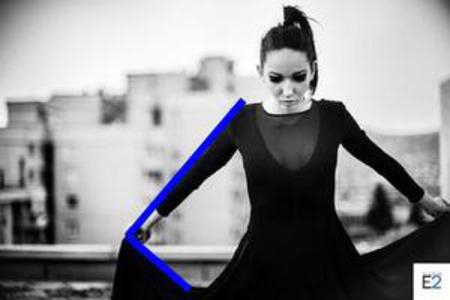}
			\includegraphics[height=0.065\textwidth,keepaspectratio]{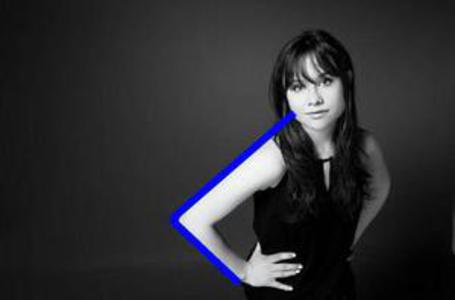} 
			\includegraphics[height=0.065\textwidth,keepaspectratio]{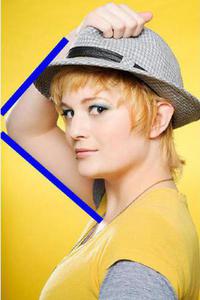}
			\includegraphics[height=0.065\textwidth,keepaspectratio]{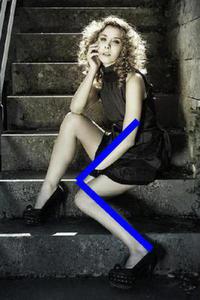} \\
			
			\includegraphics[height=0.05\textwidth]{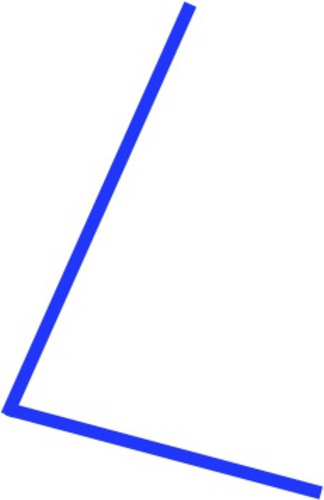} &
			\includegraphics[height=0.057\textwidth,keepaspectratio]{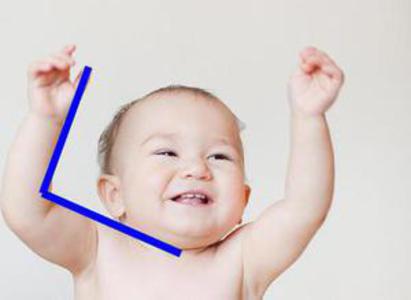} 
			\includegraphics[height=0.065\textwidth,keepaspectratio]{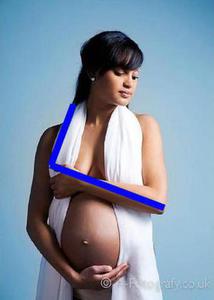}
			\includegraphics[height=0.065\textwidth,keepaspectratio]{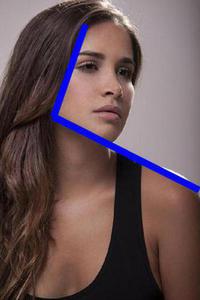}
			\includegraphics[height=0.065\textwidth,keepaspectratio]{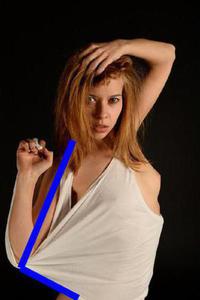}
			\includegraphics[height=0.065\textwidth,keepaspectratio]{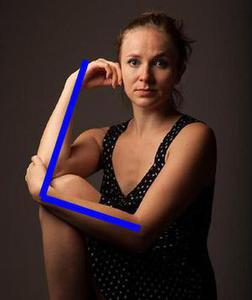}
			\includegraphics[height=0.065\textwidth,keepaspectratio]{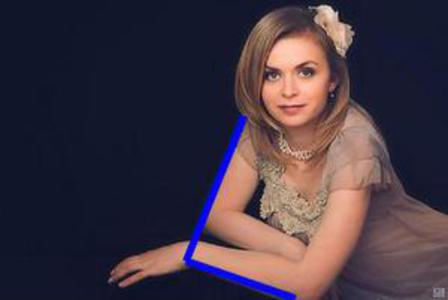} 
\\
			
			\includegraphics[width=0.05\textwidth,keepaspectratio]{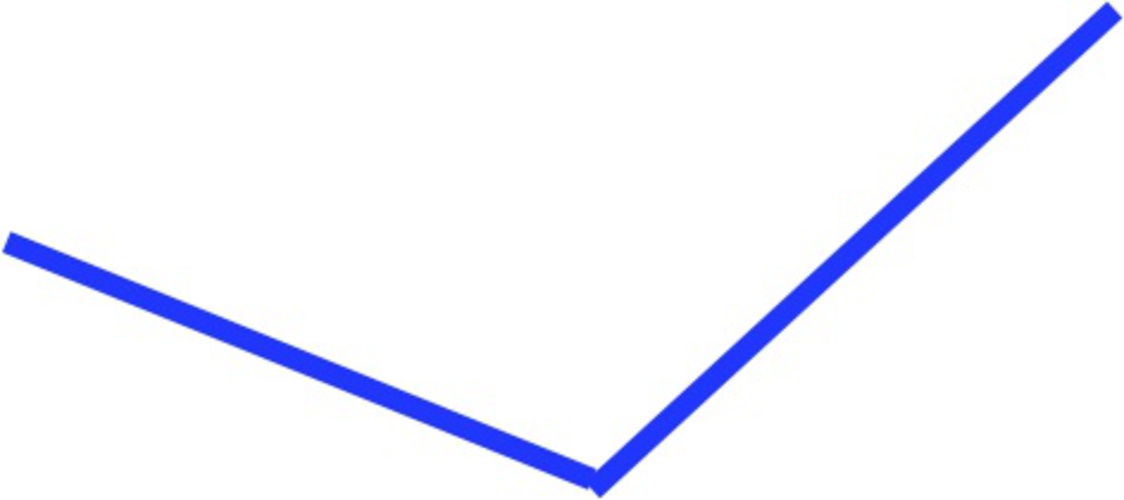} &
			\includegraphics[height=0.065\textwidth,keepaspectratio]{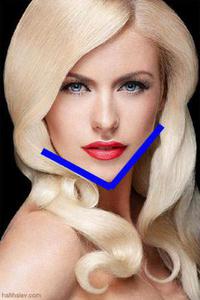} 
			\includegraphics[height=0.05\textwidth,keepaspectratio]{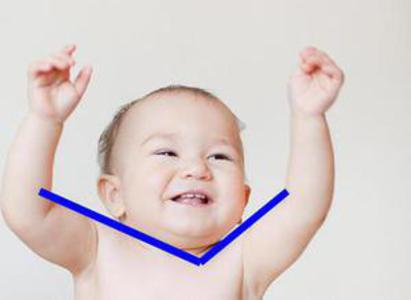} 
			\includegraphics[height=0.05\textwidth,keepaspectratio]{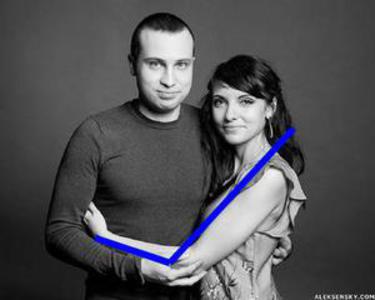}
			\includegraphics[height=0.065\textwidth,keepaspectratio]{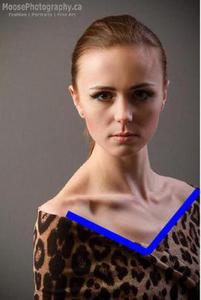} 
			\includegraphics[height=0.065\textwidth,keepaspectratio]{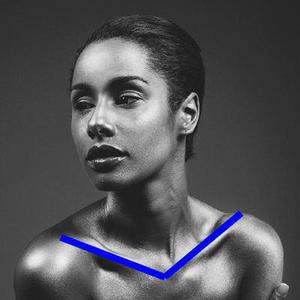}
			\includegraphics[height=0.057\textwidth,keepaspectratio]{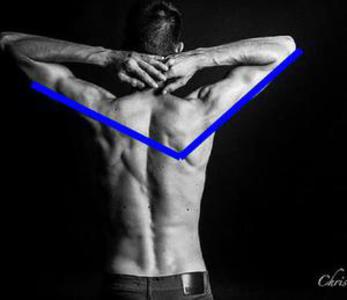} \\
			
			\includegraphics[width=0.05\textwidth]{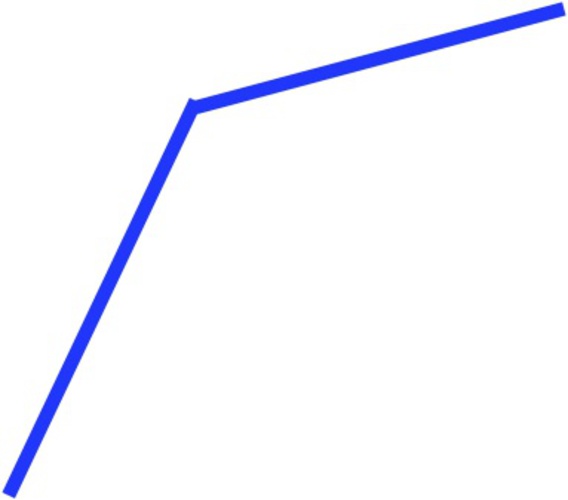} &
			\includegraphics[height=0.065\textwidth,keepaspectratio]{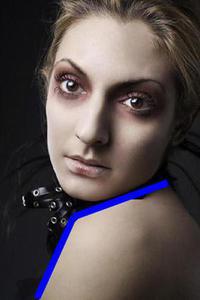} 
			\includegraphics[height=0.065\textwidth,keepaspectratio]{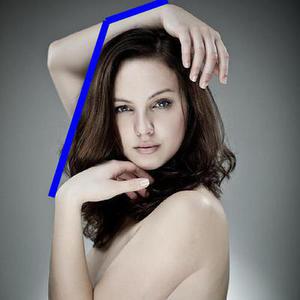} 
			\includegraphics[height=0.065\textwidth,keepaspectratio]{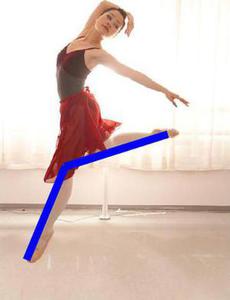} 
			\includegraphics[height=0.065\textwidth,keepaspectratio]{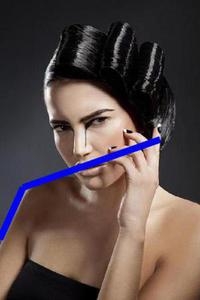} 
		\includegraphics[height=0.054\textwidth,keepaspectratio]{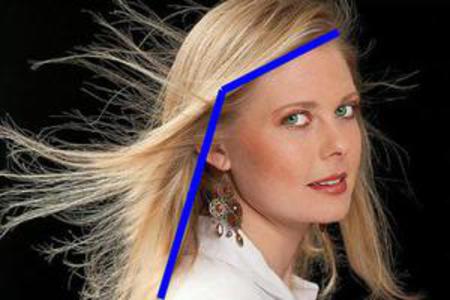} 
		\includegraphics[height=0.054\textwidth,keepaspectratio]{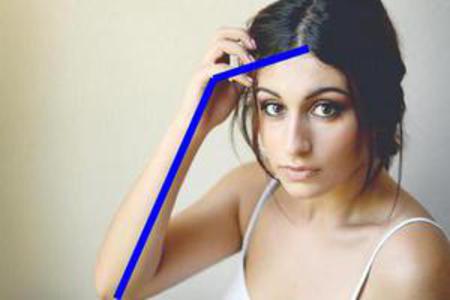} \\
			
			\includegraphics[height=0.05\textwidth]{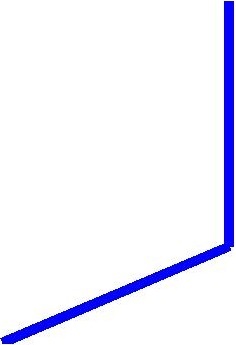} &
			\includegraphics[height=0.065\textwidth,keepaspectratio]{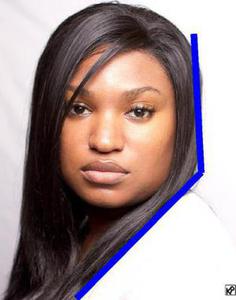} 
			\includegraphics[height=0.065\textwidth,keepaspectratio]{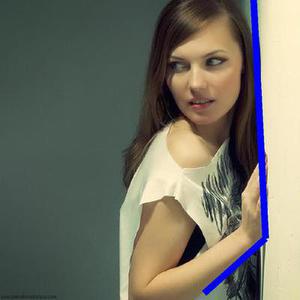} 
			\includegraphics[height=0.065\textwidth,keepaspectratio]{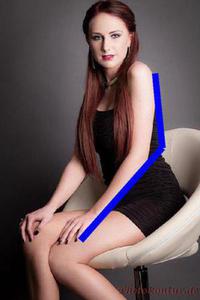} 
			\includegraphics[height=0.065\textwidth,keepaspectratio]{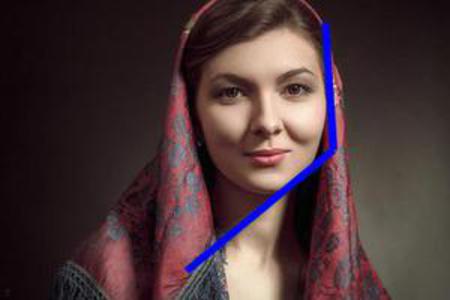} 
			\includegraphics[height=0.065\textwidth,keepaspectratio]{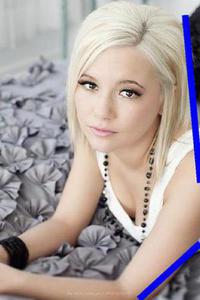} 
			\includegraphics[height=0.065\textwidth,keepaspectratio]{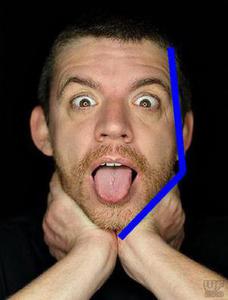} \\
			
			\includegraphics[height=0.05\textwidth]{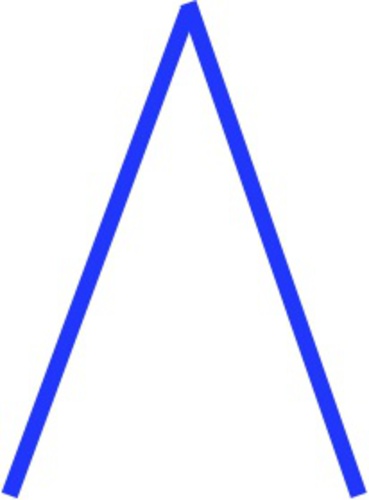} &
			\includegraphics[height=0.065\textwidth,keepaspectratio]{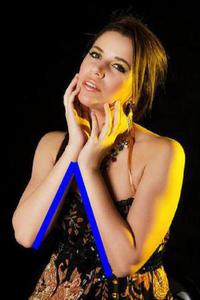} 
			\includegraphics[height=0.065\textwidth,keepaspectratio]{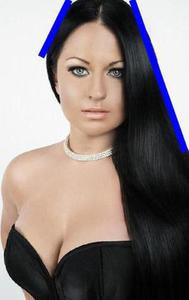}
			\includegraphics[height=0.065\textwidth,keepaspectratio]{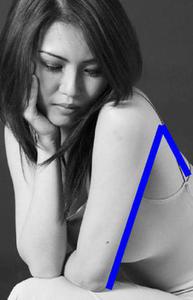}
			\includegraphics[height=0.065\textwidth,keepaspectratio]{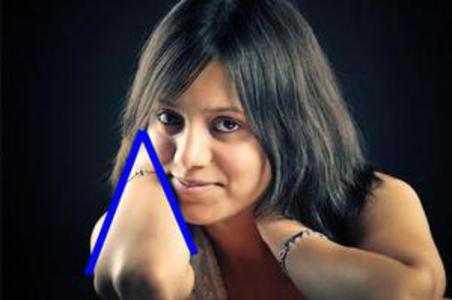}
			\includegraphics[height=0.065\textwidth,keepaspectratio]{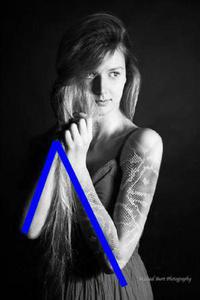} 
			\includegraphics[height=0.065\textwidth,keepaspectratio]{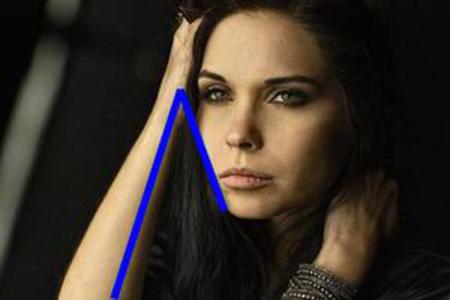} \\
			
			\includegraphics[height=0.05\textwidth]{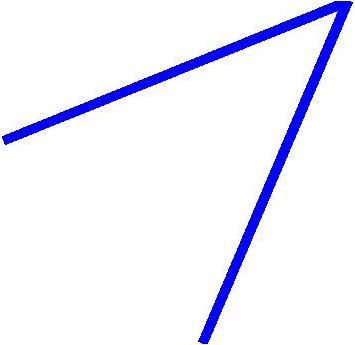} &
			\includegraphics[height=0.065\textwidth,keepaspectratio]{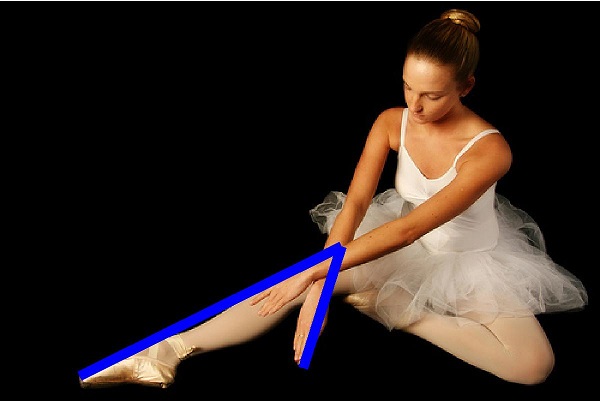}
			\includegraphics[height=0.065\textwidth,keepaspectratio]{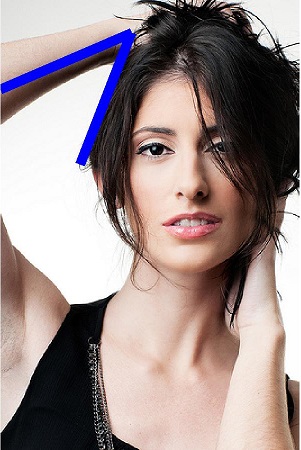}
			\includegraphics[height=0.065\textwidth,keepaspectratio]{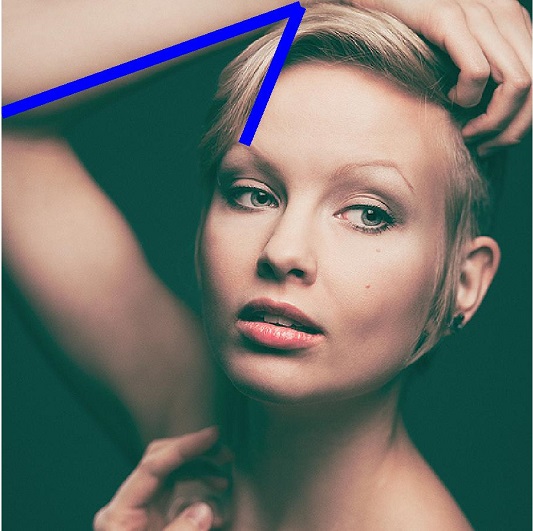}
			\includegraphics[height=0.065\textwidth,keepaspectratio]{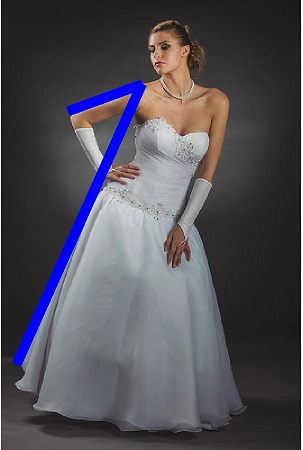}
			\includegraphics[height=0.065\textwidth,keepaspectratio]{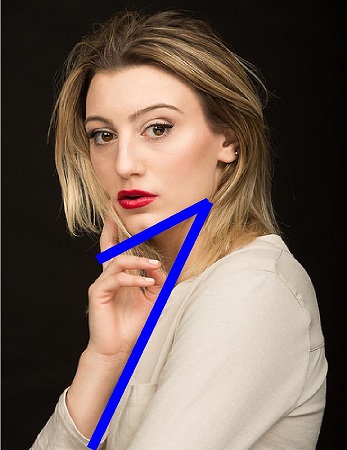} 
			\includegraphics[height=0.065\textwidth,keepaspectratio]{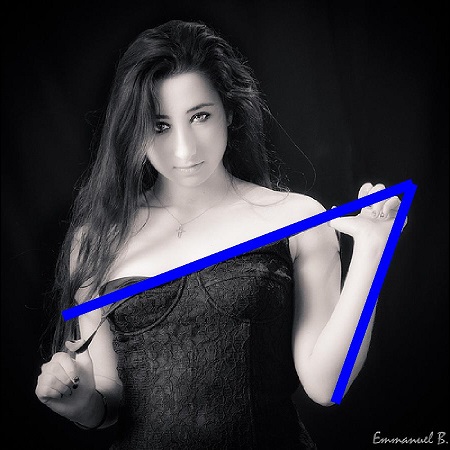} \\ \hline 
			
			\includegraphics[height=0.04\textwidth]{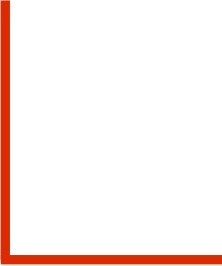} &			
			\includegraphics[height=0.047\textwidth,keepaspectratio]{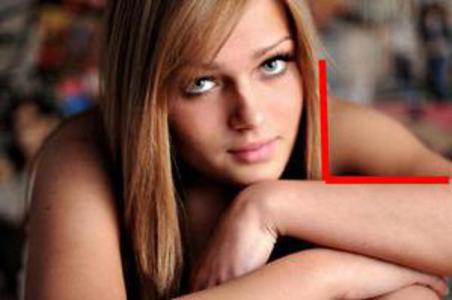} 
			\includegraphics[height=0.065\textwidth,keepaspectratio]{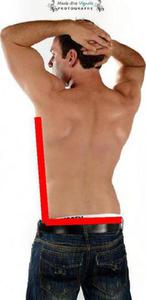} 
			\includegraphics[height=0.047\textwidth,keepaspectratio]{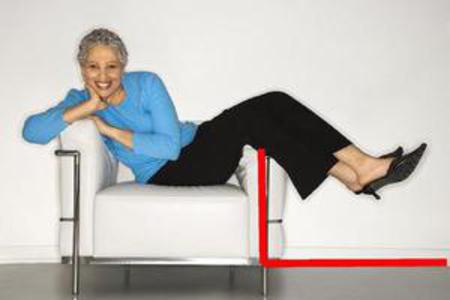} 
			\includegraphics[height=0.047\textwidth,keepaspectratio]{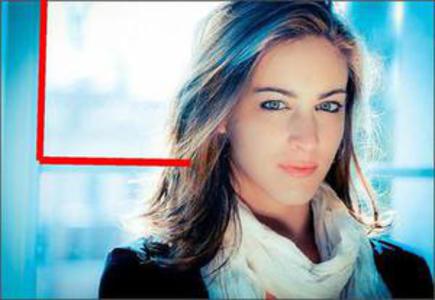}
			\includegraphics[height=0.065\textwidth,keepaspectratio]{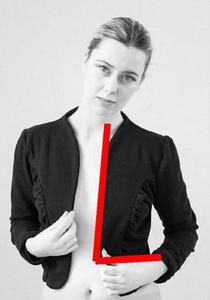} 
			\includegraphics[height=0.047\textwidth,keepaspectratio]{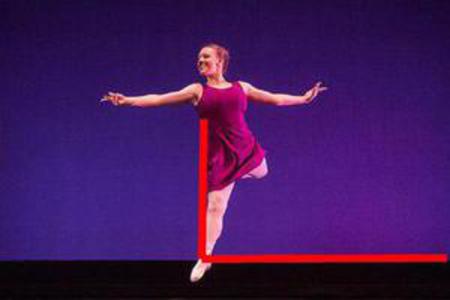} \\	
			
			\includegraphics[height=0.04\textwidth]{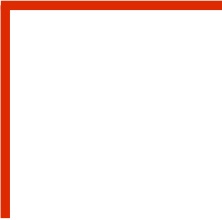} &
			\includegraphics[height=0.047\textwidth,keepaspectratio]{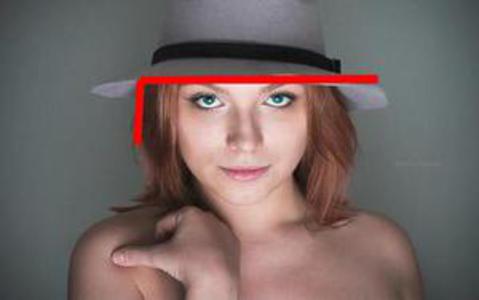}
			\includegraphics[height=0.065\textwidth,keepaspectratio]{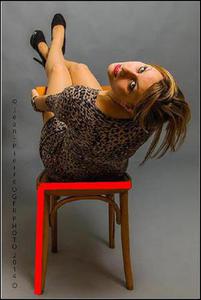} 
			\includegraphics[height=0.065\textwidth,keepaspectratio]{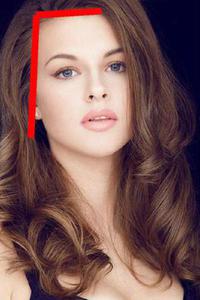} 
			\includegraphics[height=0.047\textwidth,keepaspectratio]{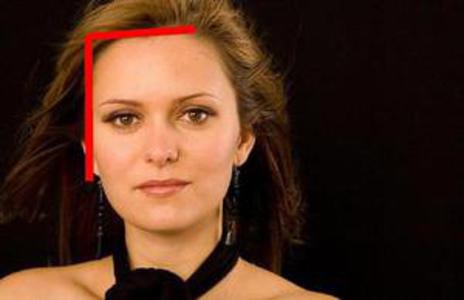} 
			\includegraphics[height=0.065\textwidth,keepaspectratio]{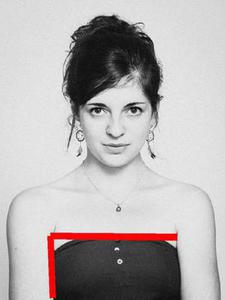} 
			\includegraphics[height=0.065\textwidth,keepaspectratio]{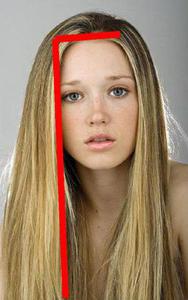} \\
			
			\includegraphics[height=0.04\textwidth]{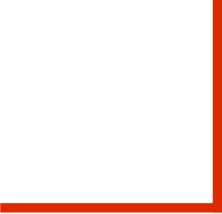} &
			\includegraphics[height=0.065\textwidth,keepaspectratio]{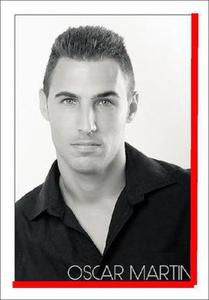}  \hskip 0.03in
			\includegraphics[height=0.065\textwidth,keepaspectratio]{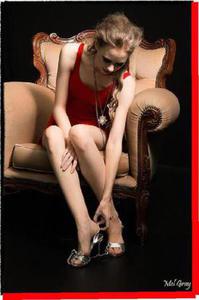}  \hskip 0.03in
			\includegraphics[height=0.065\textwidth,keepaspectratio]{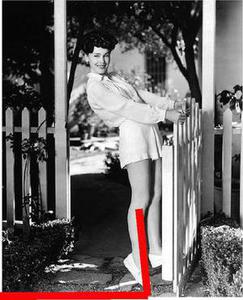}  \hskip 0.03in
			\includegraphics[height=0.065\textwidth,keepaspectratio]{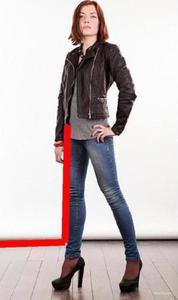}  \hskip 0.03in
			\includegraphics[height=0.065\textwidth,keepaspectratio]{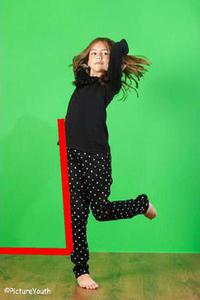}  \hskip 0.03in
			\includegraphics[height=0.065\textwidth,keepaspectratio]{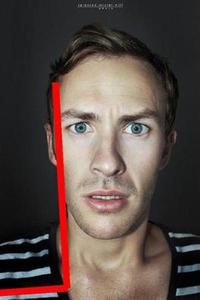} \\
			
			\includegraphics[height=0.04\textwidth]{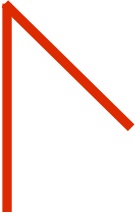} &
			\includegraphics[height=0.065\textwidth,keepaspectratio]{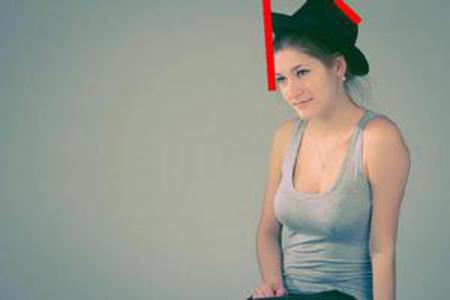}
			\includegraphics[height=0.065\textwidth,keepaspectratio]{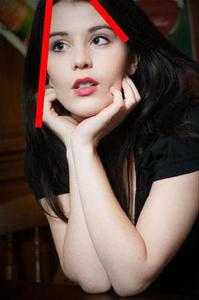}
			\includegraphics[height=0.065\textwidth,keepaspectratio]{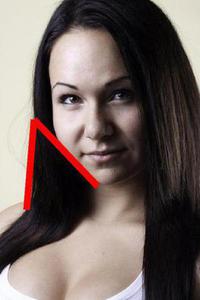}
			\includegraphics[height=0.065\textwidth,keepaspectratio]{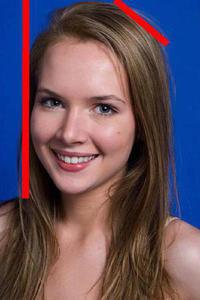}
			\includegraphics[height=0.065\textwidth,keepaspectratio]{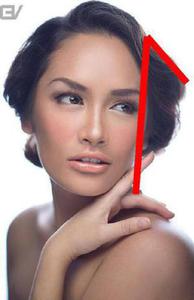}
			\includegraphics[height=0.065\textwidth,keepaspectratio]{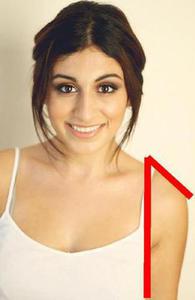} \\
		\end{tabular}
	\end{center}
	\caption{Examples of retrieved portraits. Each row shows results from a query. Red examples are considered ``unuseful'' in our evaluation.}
	\label{fig:retrievedres}
\end{figure}

In Figure~\ref{fig:retrievedres}, we provide examples of both ``useful'' and ``unuseful'' triangles retrieved by our system. The first column contains twelve queries among which eight return high percentages of ``useful'' triangles and four return low percentage of ``useful'' triangles. 
From the retrieved examples, it can be seen that professional photographers are skillful at using all kinds of objects, such as arm, leg, shoulder, chair, wall, ground, hair, apparel, or even shadow, to construct triangles. Interestingly, very often a slight adjustment of a pose can form beautiful triangles which make the entire composition aesthetically appealing. For instance, the girl in row 3, column 3 slightly turns her head towards left to perfectly align with her shoulder. 
For the same reason, the girl in row 1, column 4 lifts up her head a little bit. 

In addition, our system also retrieves many ``unuseful'' triangles for some queries. As shown in the last four rows of Figure~\ref{fig:retrievedres}, many of these triangles are right triangles. In fact, it is known that professional photographers often avoid $90^\circ$ body angles because they
often look unnatural and strained~\citep{valenzuela2012picture}. This may partly explain why we are unable to retrieve more ``useful'' triangles in these cases.

\section{Conclusions and Future Work}\label{sec:con}

This paper proposes a system that detects the usage of triangle photo
composition techniques in both natural/urban scene and portrait/people
photography. We show preliminary evidence through human subject
studies that such systems can potentially help consumer photographers
learn about professional composition of photographs. The two broad
categories that we have chosen cover most major consumer photography
genres. Additionally, we show that line-based methods are effective
for both categories, despite the fact that many photos have no
clearly-defined straight lines. For photographs of natural/urban scene
with strong perspective effect, we have demonstrated a new method for
modeling visual composition through analyzing the perspective effects
and segmenting the image based on photometric and geometric cues. The
method can effectively detect the dominant vanishing point from an
arbitrary scene.  For portrait photographs, we extract a set of
candidate line segments from a photo and then successfully fit a
triangle to these segments despite a large proportion of outliers. The
fitted result accurately identifies the presence of triangles in
photographs.  Among a variety of potential applications, we have
illustrated how our techniques can provide on-site feedback to
photographers.

Our work opens up several future directions. First, we plan to investigate the relationships between triangles and other visual elements and design principles. For example, one challenge in composition recognition for real-world photos is the presence of large foreground objects. They typically correspond to regions
which are not associated with any vanishing point in the image. In addition, some photos may solely focus on the objects (\eg, a flower) and do not
possess a well-defined perspective geometry. We will analyze the composition of these images by first separating the foreground objects from the background.
We note that, while our analysis of the perspective geometry provides valuable information about the 3D space, many popular composition rules studied in early work, such as the simplicity of the scene, golden ratio, rule of thirds, and visual balance have focused on the arrangement of objects in the 2D image plane. We believe that combining the strength of both approaches will enable us to obtain a deeper understanding of the composition of these images.

Beyond image composition, the relationship
between triangles and the aesthetic quality of compositions can be
further studied. For example, photographers may use different composition techniques in different situations. How to assess the relevance of perspective in a natural/urban scene photo? Also, for portraits, how do the number, sizes, shapes, and orientations of
triangles influence the aesthetics of photo composition? Answering
such questions can help amateur photographers learn more specific
photography techniques. 

Finally, apart from providing on-site feedback to
photographers, our method can also be implemented as a component in
large-scale image retrieval engines in the cloud. When a query results in a large number of images that have similar levels of visual similarity or aesthetic quality, the query results can be structured as a tree with levels of refinement in terms of composition by grouping the images using a hierarchical clustering scheme.

\begin{acknowledgements}

Chuck S. Fong of the Studio 2 Photography provided ground truth
annotations for the professional portrait dataset used in our
evaluation. Edward Chen and Sahil Mishra assisted in developing the
data-collection system for the human subject studies. We thank the
participants in the studies for their assistance. The authors would
also like to acknowledge the comments and suggestions from the
reviewers and the guest editors. 

\end{acknowledgements}

\small
\bibliographystyle{spbasic}      
\bibliography{triangle}   

\end{document}